%% file: main.tex
\documentclass[sigconf, nonacm, pbalance, pdfa]{acmart}
\usepackage[a-2b]{pdfx}
\newcommand\vldbdoi{10.14778/3705829.3705844}
\newcommand\vldbpages{1 - 16}
\newcommand\vldbvolume{18}
\newcommand\vldbissue{2}
\newcommand\vldbyear{2024}

\newcommand\vldbtitle{\shorttitle} 
\newcommand\vldbavailabilityurl{https://github.com/kvmduc/TEAM-topo-evo-traffic-forecasting}
\newcommand\vldbpagestyle{plain}

\usepackage{graphicx} 
\usepackage{latexsym}
\usepackage{amsmath,amsfonts}
\usepackage{pifont}
\usepackage{mathtools}
\usepackage{footmisc}
\usepackage[linesnumbered,ruled,vlined]{algorithm2e}
\usepackage{empheq}
\usepackage{pgfplots,tikz}
\pgfplotsset{compat=1.18}
\usepackage{cases} 
\usepackage{placeins}
\usepackage{microtype}
\usepackage[font=small,labelfont=bf]{caption}
\usepackage{subcaption}
\usepackage{float}
\usepackage{booktabs}
\usepackage{multicol}
\usepackage{multirow}
\usepackage{hhline}
\usepackage{tabu}
\usepackage{xcolor}
\usepackage{pbox}
\usepackage{cuted}
\usepackage[switch]{lineno}

\usepackage{array}
\usepackage{etoolbox}
\usepackage[normalem]{ulem}
\newcolumntype{P}[1]{>{\centering\arraybackslash}p{#1}}%
\SetKwInput{KwInput}{Input}                
\SetKwInput{KwOutput}{Output}              
\SetKw{Break}{break}

\SetCommentSty{mycommfont}

\SetAlgoSkip{}
\makeatletter
\newcommand{\xMapsto}[2][]{\ext@arrow 0599{\Mapstofill@}{#1}{#2}}
\def\Mapstofill@{\arrowfill@{\Mapstochar\Relbar}\Relbar\Rightarrow}
\makeatother

\title{TEAM: Topological Evolution-aware Framework \\for Traffic Forecasting--Extended Version}

\author{\texorpdfstring{Duc Kieu$^{1,2,\ast}$, Tung Kieu$^{3,\ast}$, Peng Han$^4$, Bin Yang$^{5,\#}$, Christian S. Jensen$^3$, and Bac Le$^{1,2}$}{Duc Kieu, Tung Kieu, Peng Han, Bin Yang, Christian S. Jensen, and Bac Le}}

\affiliation{%
  \institution{\texorpdfstring{$^1$University of Science, Ho Chi Minh City, Vietnam
  $^2$Vietnam National University, Ho Chi Minh City, Vietnam\\
  $^3$Aalborg University, Denmark
  $^4$University of Electronic Science and Technology of China, China\\
  $^5$East China Normal University, China}}\\
  \country {\texorpdfstring{$^{1,2}$\{18127080,lhbac\}@hcmus.edu.vn $^3$\{tungkvt,csj\}@cs.aau.dk $^4$penghan@uestc.edu.cn $^5$byang@dase.ecnu.edu.cn}} \\
}

\begin{document}


\clearpage
\pagenumbering{arabic}

\input{sections_arXiv/abstract}
\thispagestyle{plain}
\pagestyle{plain}
\maketitle

\pagestyle{\vldbpagestyle}
\begingroup\small\noindent\raggedright\textbf{PVLDB Reference Format:}\\
\texorpdfstring{Duc Kieu$^{1,2}$, Tung Kieu$^{3}$, Peng Han$^4$, Bin Yang$^{5}$, Christian S. Jensen$^3$, and Bac Le$^{1,2}$}{Duc Kieu, Tung Kieu, Peng Han, Bin Yang, Christian S. Jensen, and Bac Le}
. \vldbtitle. PVLDB, \vldbvolume(\vldbissue): \vldbpages, \vldbyear.\\
\href{https://doi.org/\vldbdoi}{doi:\vldbdoi}
\endgroup
\begingroup
\renewcommand\thefootnote{}\footnote{\noindent
{\large $\ast$}: Equal contributions,  \#: Corresponding author\\ 
This work is licensed under the Creative Commons BY-NC-ND 4.0 International License. Visit \url{https://creativecommons.org/licenses/by-nc-nd/4.0/} to view a copy of this license. For any use beyond those covered by this license, obtain permission by emailing \href{mailto:info@vldb.org}{info@vldb.org}. Copyright is held by the owner/author(s). Publication rights licensed to the VLDB Endowment. \\
\raggedright Proceedings of the VLDB Endowment, Vol. \vldbvolume, No. \vldbissue\ %
ISSN 2150-8097. \\
\href{https://doi.org/\vldbdoi}{doi:\vldbdoi} \\
}\addtocounter{footnote}{-1}\endgroup

\ifdefempty{\vldbavailabilityurl}{}{
\vspace{.3cm}
\begingroup\small\noindent\raggedright\textbf{PVLDB Artifact Availability:}\\
The source code, data, and/or other artifacts have been made available at \url{\vldbavailabilityurl}.
\endgroup
}
\thispagestyle{plain}
\input{sections_arXiv/introduction}
\input{sections_arXiv/preliminaries}

\input{sections_arXiv/methodology}
\input{sections_arXiv/experiments}
\input{sections_arXiv/relatedwork}
\input{sections_arXiv/conclusion}

\section*{Acknowledgments}
We thank Khanh-Toan Nguyen and Thin Nguyen from A\textsuperscript{2}I\textsuperscript{2}, Deakin University, Australia for fruitful discussions and technical help.

\bibliographystyle{ACM-Reference-Format}
\bibliography{references/reference.bib}

\end{document}

%% file: sections_arXiv/abstract.tex
\begin{abstract}
Due to the global trend towards urbanization, people increasingly move to and live in cities that then continue to grow.
Traffic forecasting plays an important role in the intelligent transportation systems of cities as well as in spatio-temporal data mining. State-of-the-art forecasting is achieved by deep-learning approaches due to their ability to contend with complex spatio-temporal dynamics. However, existing methods assume the input is fixed-topology road networks and static traffic time series. These assumptions fail to align with urbanization, where time series are collected continuously and road networks evolve over time. In such settings, deep-learning models require frequent re-initialization and re-training, imposing high computational costs. To enable much more efficient training without jeopardizing model accuracy, we propose the \underline{\texttt{T}}opological \underline{\texttt{E}}volution-\underline{\texttt{a}}ware Fra\underline{\texttt{m}}ework (\texttt{TEAM}) for traffic forecasting that incorporates convolution and attention. This combination of mechanisms enables better adaptation to newly collected time series, while being able to maintain learned knowledge from old time series. \texttt{TEAM} features a continual learning module based on the Wasserstein metric that acts as a buffer that can identify the most stable and the most changing network nodes. Then, only data related to stable nodes is employed for re-training when consolidating a model. Further, only data of new nodes and their adjacent nodes as well as data pertaining to changing nodes are used to re-train the model. Empirical studies with two real-world traffic datasets offer evidence that \texttt{TEAM} is capable of much lower re-training costs than existing methods are, without jeopardizing forecasting accuracy.
\end{abstract}

%% file: sections_arXiv/introduction.tex
\section{Introduction}
Transportation has recently experienced substantial development, owing to technological advancements. 
One of the main developments is the widespread diffusion of sensor-equipped devices, resulting in the massive production of transportation data. 
This information further supports inexpensive and effective transportation management solutions~\cite{DBLP:conf/cikm/KieuYGJ18}. 
For example, proximity sensors or speeding cameras are installed on different roads and intersections to continuously collect traffic information such as traffic flow and speed. 
The result is the availability of large amounts of time-ordered traffic observations, known as traffic time series.

Traffic forecasting fueled by traffic data, is a core element of many applications in intelligent transportation and plays an important role in spatio-temporal data mining. 
Forecasted information (e.g., traffic flow and traffic speed), which offers insight into the dynamic characteristics of the underlying traffic network, can contribute to, e.g., alarms for hotspots~\cite{DBLP:conf/mdm/KieuYJ18}, route planning~\cite{guo2024efficient}, and route recommendation~\cite{DBLP:conf/kdd/YangHGYJ23}. 
To capture complex spatio-temporal dynamics, many deep learning based methods have recently been proposed and show promising results on challenge datasets due to their ability to learn non-linear dynamics~\cite{DBLP:conf/ijcai/YuYZ18,DBLP:conf/iclr/LiYS018}. 
These methods are often built on Graph Neural Networks (\texttt{GNN}s)~\cite{DBLP:conf/nips/DefferrardBV16,DBLP:conf/iclr/KipfW17} and Temporal Neural Networks (\texttt{TCN}s)~\cite{DBLP:journals/neco/HochreiterS97,DBLP:jour/corr/OordDZSVGKSK16} to capture spatial and the temporal information, respectively.
Although achieving competitive performance, these methods face two challenges.

\begin{figure}[t]
    \centering
        \begin{subfigure}[t]{0.63\linewidth}
            \centering
            \includegraphics[width=0.9\linewidth]{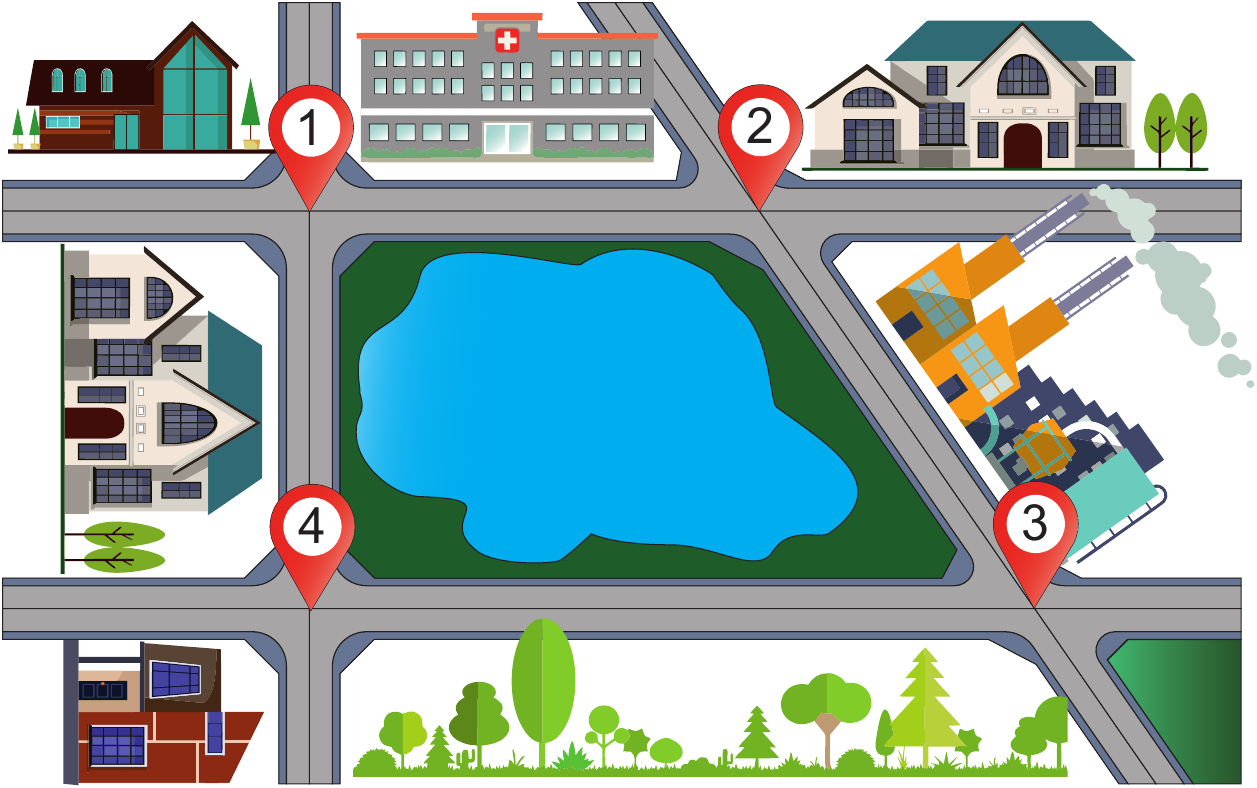}
            \caption{Historical city.}
            \label{subfig:city_history}
        \end{subfigure}
        \begin{subfigure}[t]{0.355\linewidth}
            \centering
            \includegraphics[width=0.9\linewidth]{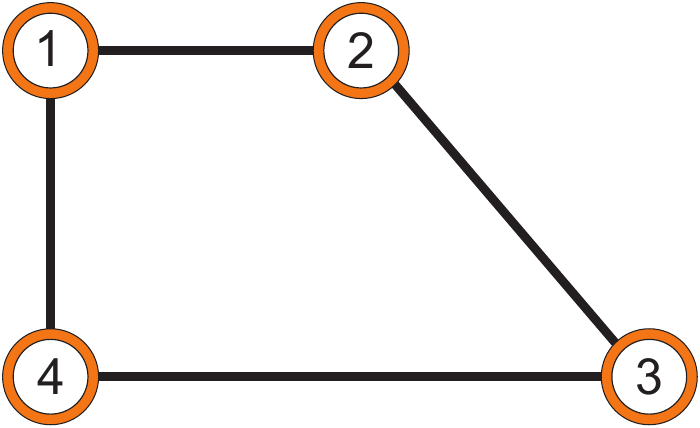}
            \caption{Historical RN.}
            \label{subfig:road_network_history}
        \end{subfigure}
        \begin{subfigure}[t]{0.63\linewidth}
            \centering
            \includegraphics[width=0.9\linewidth]{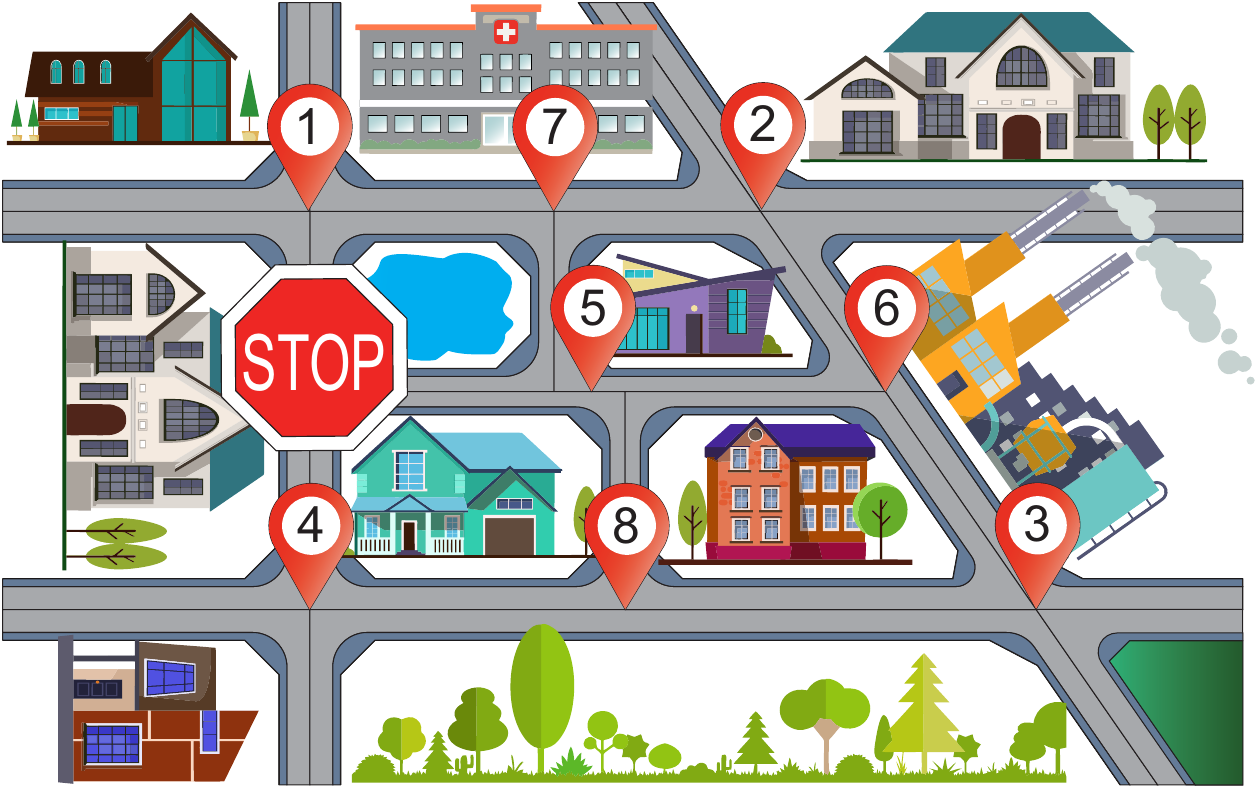}
            \caption{Current city.}
            \label{subfig:city_current}
        \end{subfigure}
        \begin{subfigure}[t]{0.355\linewidth}
            \centering
            \includegraphics[width=0.9\linewidth]{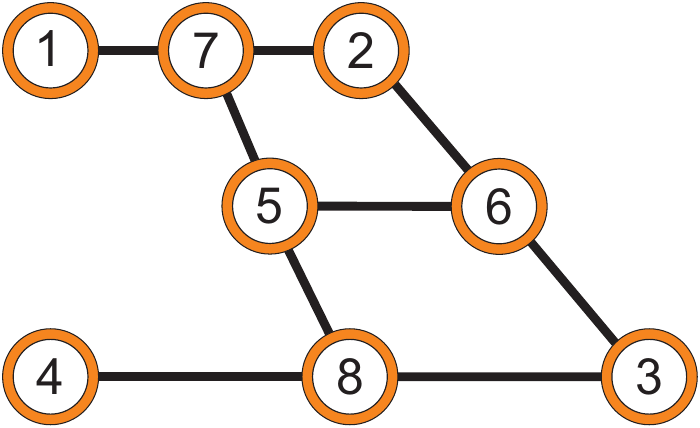}
            \caption{Current RN.}
            \label{subfig:network_current}
        \end{subfigure}
        \caption{Example of a city and its corresponding evolving RN.}
        \label{fig:evolving_road_network_example}
        \vspace{0.5em}
\end{figure}

First, traffic forecasting is a long-term task, where traffic behaviors can change over time due to changes in the graph structure of road networks (RNs). 
For example, RNs that represent cities often expand to support growing populations caused by urbanization. This results in new nodes and edges being added to existing RNs.
Meanwhile, old roads and regions can become obsolete, leading to the elimination of nodes and edges from existing RNs. Also, in some countries, roads are frequently blocked, and directions are frequently changed depending on the day of the week. Observations such as these motivate solutions to the problem of traffic forecasting for RNs with evolving topology.

Fig.~\ref{fig:evolving_road_network_example} shows an example of a city and its evolving RN. 
However, existing forecasting proposals only work on static RNs~\cite{DBLP:conf/ijcai/YuYZ18,DBLP:conf/iclr/LiYS018,DBLP:conf/aaai/ZhengFWQ20} and thus do not accommodate real-world scenarios.
A trivial solution is to re-initialize a new model and re-train it on the new RN data whenever the RN has evolved.
However, this imposes challenges to both storage and computation because of high data processing and model training overheads, particularly when RNs grow.
Another way of accommodating evolving RNs is to train a model on historical regions and then transfer the learned knowledge to a new model, which is further trained on data from updated RN parts.
However, the direct use of transferred models faces two limitations.  
(i) The historical and new temporal data of a node may not exhibit similar patterns because traffic data evolves and exhibits the data shift characteristics~\cite{DBLP:journals/tnn/YangCDT22,DBLP:journals/pvldb/CamposYKZGJ24}. 
Transferred models cannot learn inconsistent patterns between the historical and the new data, thus exhibiting substandard performance.
Also, when the topology of an RN changes, traffic behaviors corresponding to the new topology deviate from the previous behaviors. Here, spatial dependencies captured by transferred models may no longer be appropriate.
(ii) Useful information, such as stable patterns that are captured by the transferred models, may be forgotten after being transferred, rather than being consolidated~\cite{DBLP:conf/nips/ChenWFLW19}. 
Models that forget stable patterns may experience substandard performance.

Second, we observe that traffic forecasting models can achieve impressive accuracy due to the practice of fitting on massive datasets. However, in practical scenarios, the cost of constructing and maintaining a forecasting model is high. 
As mentioned above, a natural consideration is to transfer the knowledge and only require model training on updated regions.
However, the data for the evolution of RNs, which is used to transfer the previous model, is substantially smaller than that used for a re-initialized model. 
An effective model is then required to incrementally capture dynamic and complex spatio-temporal correlations given a small-scale dataset. 
However, recent studies \cite{DBLP:conf/aaai/GuoLFSW19, DBLP:conf/ijcai/WuPLJZ19, DBLP:conf/aaai/SongLGW20} focus on building forecasting models with high accuracy, overlooking the ability to learn effectively to model complex and non-linear spatio-temporal dependencies with small-scale data.

\textcolor{black}{We propose a \texttt{T}opological \texttt{E}volution}-\texttt{a}ware Fra\texttt{m}ework (\texttt{TEAM}) for traffic forecasting to solve the above two problems.
To tackle the first problem, we propose a continual learning module that adopts the rehearsal-based continual learning mechanism. 
This module works as a buffer and stores a limited number of historical data samples. 
The module is then integrated into the traffic forecasting framework, where it provides stored samples to enforce forecasting model rehearsal when new data is similar to historical data. 
By rehearsing, the historical knowledge is consolidated, and the model can mitigate forgetting.
In case the new data is different from the historical data (i.e., exhibits distribution shift~\cite{DBLP:conf/kdd/GuoZL20}), the historical knowledge is no longer useful, and thus rehearsal is unnecessary. 

To measure the difference between historical and new data, we transform the historical and new data into two histograms and use the Wasserstein metric to compute their similarity.
A limited number of historical data samples of several historical graph nodes that have high similarity (i.e., the most stable nodes) are selected and stored in the buffer.
In addition, those that have the lowest similarity (i.e., the most changing nodes) are selected for update with the new data.
A new adjacency matrix is constructed from the most stable nodes, the most changing nodes, and the newly added nodes. 
Finally, data in the buffer and evolved data of the new adjacency matrix serve as the input for training the transferred model.
The constructed adjacency matrix is significantly smaller than the adjacency matrix of the entire RN. Thus, the complexity of the model training is reduced substantially.

To overcome the second issue, we propose a model, called \textbf{\texttt{C}}onvo-\\lution \textbf{\texttt{A}}ttention for \textbf{\texttt{S}}patio-\textbf{\texttt{T}}emporal (\texttt{CAST}), that uses a hybrid architecture that combines convolution and attention modules for both spatial and temporal computations.
Existing studies~\cite{DBLP:conf/nips/DaiLLT21,DBLP:conf/iccv/WuXCLDY021} show that combining convolution and attention allows a model to learn faster and to converge more easily so that the model can work well on small-scale datasets.
Hybridizing convolution and attention also enables a model to better model    dynamic and non-linear spatial-temporal correlations.
Further, convolution focuses on local patterns such as seasonalities, variations w.r.t. the temporal aspect, and closed-neighbor-nodes w.r.t. the spatial aspect~\cite{DBLP:conf/nips/OnoTFY18}. In contrast,  attention focuses on global patterns such as trends w.r.t. the temporal aspect and far-neighbor-nodes w.r.t. the spatial aspect~\cite{DBLP:conf/ijcai/CirsteaG0KDP22}. 
By combining convolution and attention, we aim to obtain a model that exploits both local and global patterns to yield better accuracy.

To the best of our knowledge, this is the first study to enable traffic forecasting for evolving RNs.
We make four contributions.
First, we formulate the problem of traffic forecasting in evolving RNs.
Second, we propose \texttt{CAST}, a framework for traffic forecasting, where the main model can learn effectively on small-scale data using a hybrid architecture.
Third, we propose a continual learning module that enables the forecasting model to effectively learn on evolving RNs. Then, we integrate the module into \texttt{CAST} to form \texttt{TEAM}.
Fourth, we report on extensive empirical studies that offer insight into pertinent design properties of the proposed framework and offer evidence that the proposed framework is able to outperform the state-of-the-art approaches w.r.t. accuracy and runtime.

The rest of the paper is organized as follows. 
Section~\ref{sec:preliminaries} describes the preliminaries and formalizes the problem. 
Section~\ref{sec:methodology} presents the methodology.
Section~\ref{sec:experiments} describes the experimental study and the results. 
Section~\ref{sec:relatedwork} discusses related work, and Section~\ref{sec:conclusion} concludes the paper.

%% file: sections_arXiv/preliminaries.tex
\section{Preliminaries}
\label{sec:preliminaries}

\subsection{Traffic Forecasting}
Consider an RN modeled as a graph $G = (V, E)$, where $V$ is a finite set of $|V| = N$ vertices and $E$ is a set of edges, and let $G$ be represented by an adjacency matrix $\mathbf{A}$. 
Assume corresponding data for all vertices over $P$ historical time steps $\mathcal{X} = \langle \mathbf{X}_{T-P+1}, \dots, \mathbf{X}_{T} \rangle$. 
The data at time step $T$ is denoted as $\mathbf{X}_{T} \in \mathbb{R}^{N \times F}$, where $F$ is the number of trafﬁc signal features. 
The goal of traffic forecasting is to build a model that works as a function $f_{\theta}(.)$ that learns from $P$ previous steps of data $\mathcal{X}$ to forecast data for $H$ future steps of data $\mathcal{\hat{X}} = \langle \mathbf{\hat{X}}_{T+1}, \dots, \mathbf{\hat{X}}_{T+H} \rangle$ according to Eq.~\ref{eq:rescaled}.

\vspace{-1.0em}
\begin{equation}
    \langle \textbf{X}_{T-P+1}, \dots, \textbf{X}_{T} \rangle \xmapsto{f_\theta(.)} \langle \hat{\textbf{X}}_{T+1}, \dots, \hat{\textbf{X}}_{T+H}\rangle
    \label{eq:rescaled}
\end{equation}

\subsection{Graph Evolution}

\begin{figure}[H]
    \centering
    \includegraphics[width=0.55\linewidth]{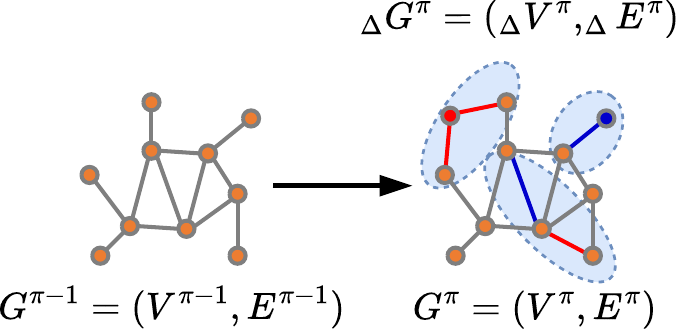}
    \caption{Evolution from $G^{\pi-1}$ to $G^{\pi}$. The blue oval regions highlight the evolved parts. Red nodes and edges denote added nodes and edges, respectively. Blue nodes and edges denote removed nodes and edges, respectively.
    }
    \label{fig:evolved_graph_network}
\end{figure}

The evolution of graph over time is conceptually represented by a series of graphs $\langle G^{1}, G^{2}, \ldots, G^{Q}\rangle$, so that $G^{\pi} = (V^{\pi}, E^{\pi})$ represents the snapshot of the graph at \textit{period} $\pi$. A period is a generic term. We can flexibly define the length of a period to a \textit{day}, a \textit{week}, a \textit{month}, a \textit{quarter}, a \textit{year}, or any other length. Since $\langle G^{1}, G^{2}, \ldots, G^{Q} \rangle$ represent snapshots of a specific graph, we have that $G^{\pi} = G^{\pi - 1} + _\Delta G^{\pi}$. Here, $_\Delta G^{\pi} = (_{\Delta} V^{\pi}, _{\Delta} E^{\pi})$ is the incremental data that captures the change (i.e., the difference) w.r.t. edges and nodes (either addition or removal) between the graph $G^{\pi}$ and the previous graph $G^{\pi-1}$. More specifically, $_\Delta G^{\pi}$ encompasses added and removed nodes, added and removed edges, and all existing nodes incident on the added and removed edges. Fig.~\ref{fig:evolved_graph_network} exemplifies $G^{\pi - 1}$, $G^{\pi}$, and the evolved part $_\Delta G^{\pi}$.

\subsection{Problem Statement}

Assume a long-term RN modeled as a sequence of graph snapshots $\langle G^{1}, G^{2}, \dots, G^{Q} \rangle$, where each $G^{\pi} = G^{\pi - 1} + _\Delta G^{\pi}$ represents the evolution of the RN from period $\pi-1$ to period $\pi$ due to many reasons such as the expansion or shrinkage of a region, newly-built roads, and discontinued roads.

For each $_\Delta G^{\pi}$, $P$ corresponding historical time steps $_\Delta \mathcal{X}^{\pi} = \langle _\Delta \mathbf{X}^{\pi}_{T-P+1}, \dots, _\Delta \mathbf{X}^{\pi}_{T} \rangle$ are given. 
We aim to build a framework that works as a function series $\langle f^{1}_{\theta}(.), f^{2}_{\theta}(.), \dots, f^{Q}_{\theta}(.) \rangle$, where each $f^{\pi}_{\theta}(\cdot)$ is transferred from $f^{\pi - 1}_{\theta}(\cdot)$ and learns from the data from the $P$ previous steps of the evolved parts in $_\Delta \mathcal{X}^{\pi}$ to forecast data for $H$ steps into the future for the entire RN in $\mathcal{\hat{X}}^{\pi} = \langle \mathbf{\hat{X}}_{T+1}, \dots, \mathbf{\hat{X}}_{T+H} \rangle$ as show in Eq.~\ref{eq:problem_definition}.

\vspace{-1.5em}
{
    \begin{align}
        \begin{cases}
            \langle \textbf{X}^{\pi}_{T-P+1}, \dots, \textbf{X}^{\pi}_{T} \rangle \xmapsto{f^{\pi}_\theta(.)} \langle \hat{\textbf{X}}^{\pi}_{T+1}, \dots, \hat{\textbf{X}}^{\pi}_{T+H} \rangle, \text{if } \pi = 1 \\
            \langle _\Delta \textbf{X}^{\pi}_{t-P+1}, \dots, _\Delta \textbf{X}^{\pi}_{T} \rangle \xmapsto{f^{\pi}_\theta(.)} \langle \hat{\textbf{X}}^{\pi}_{T+1}, \dots, \hat{\textbf{X}}^{\pi}_{T+H} \rangle , \text{if } \pi > 1
        \end{cases}
        \label{eq:problem_definition}
    \end{align}
}
\vspace{-0.5em}

%% file: sections_arXiv/methodology.tex
\section{Methodology}
\label{sec:methodology}
\subsection{Framework Overview}

\begin{figure}[h]
    \centering
    \includegraphics[width=1.0\linewidth]{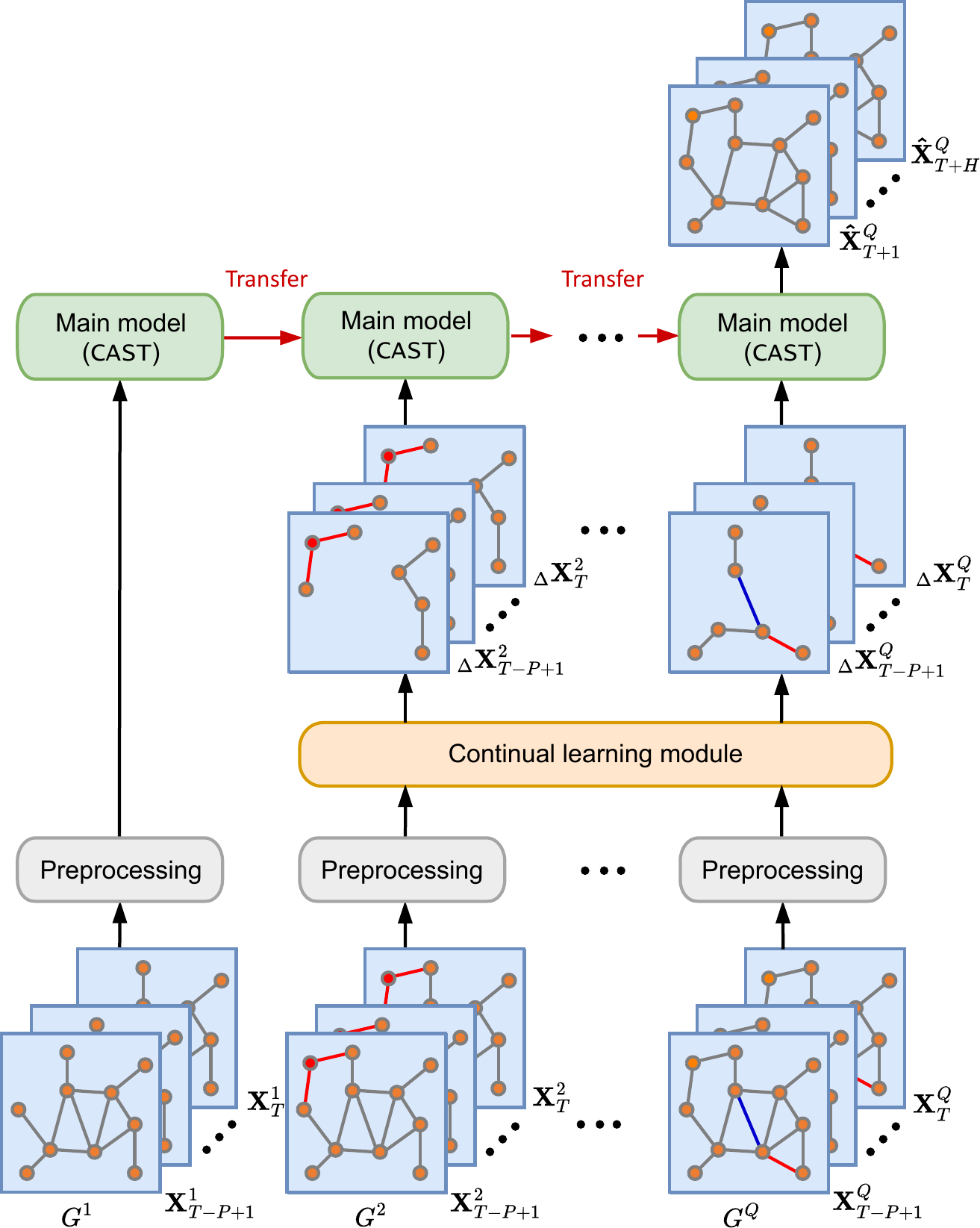}
    \caption{\texttt{TEAM} framework overview.}
    \label{fig:framework_overview}
\end{figure}

We show an overview of the proposed framework in Fig.~\ref{fig:framework_overview}.
First, the data is fed into the \textit{preprocessing} component that uses a common preprocessing technique to reduce the difference in terms of the magnitude between observations.
An observation $\mathbf{X}^{\pi}_{t} \in \mathcal{X}^{\pi}$ at each $G^{\pi}$ is re-scaled by using the mean $\mu^{\pi}$ and standard deviation $\sigma^{\pi}$ of $\mathcal{X}^{\pi}$ as follows $\displaystyle\mathbf{X}^{\pi}_{t} = \frac{(\mathbf{X}^{\pi}_{t} - \mu^{\pi})}{\sigma^{\pi}}$. 
Next, the preprocessed data $\langle \textbf{X}^{1}_{T-P+1}, \dots, \textbf{X}^{1}_{T} \rangle$ from the first period is fed into the \textit{main model}, which is covered in Section~\ref{subsec:main_model}. 
The main model is fully trained by the data from the first period.
Next, the preprocessed data from the following periods are fed into the \textit{continual learning module}, to be presented in Section~\ref{subsec:continual_module}. 
This module identifies the most stable and most changing nodes as well as their incidental edges and combines these with the evolved parts to produce the incremental data $\langle _{\Delta} \textbf{X}^{\pi}_{T-P+1}, \dots, _{\Delta} \textbf{X}^{\pi}_{T} \rangle, \pi > 1$ in the $\pi$-th period for partially training the main model. 
To do that, the trained model from the previous period is transferred and takes the incremental data as input.
The incremental data is significantly smaller than the full data.
Thus, training the model with incremental data is much faster. 
In other words, the cost of training the models in the following periods is substantially reduced.
This procedure continues until the last period $Q$. 

\subsection{Main Model}
\label{subsec:main_model}

\begin{figure}[t]
    \centering
        \begin{subfigure}[t]{0.394\linewidth}
        \centering
            \includegraphics[width=0.9\linewidth]{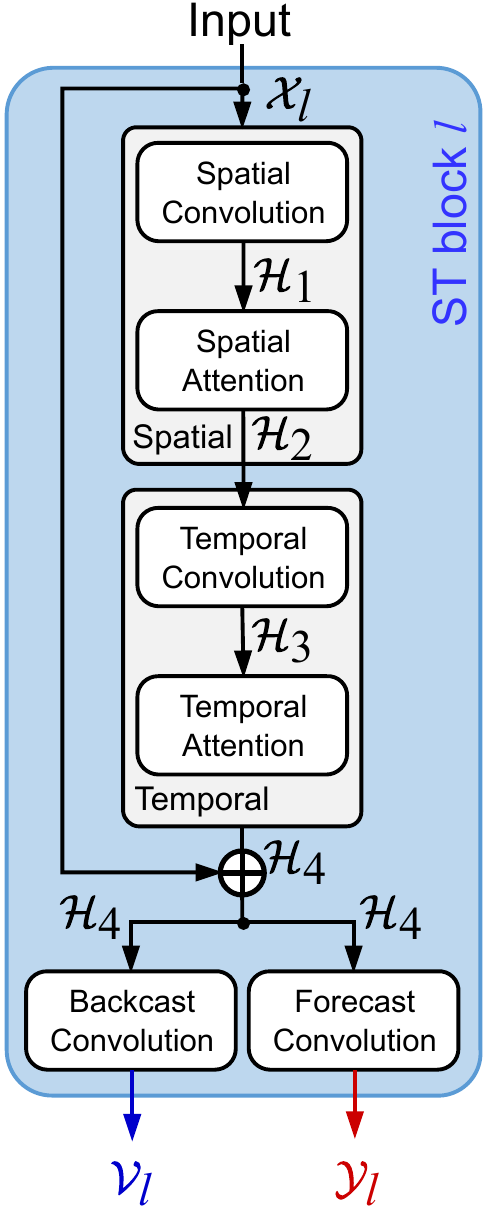}
            \caption{Block structure.}
            \label{subfig:cast_block}
        \end{subfigure}
        \begin{subfigure}[t]{0.300\linewidth}
        \centering
            \includegraphics[width=0.9\linewidth]{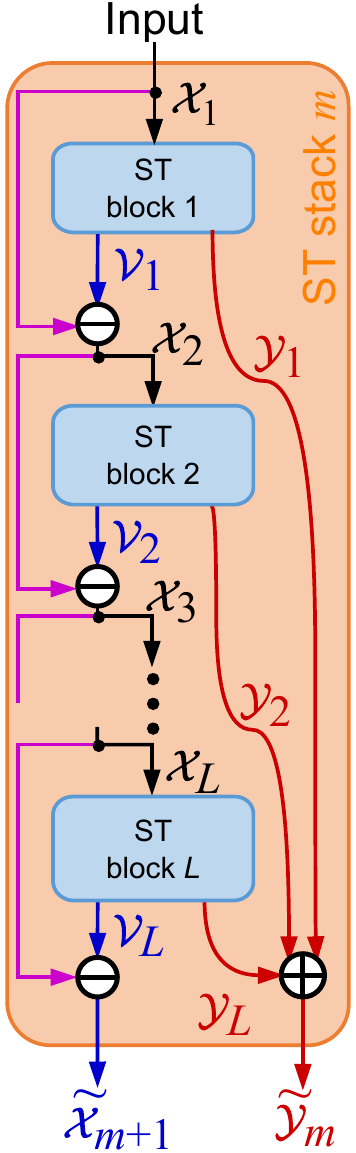}
            \caption{Stack structure.}
            \label{subfig:cast_stack}
        \end{subfigure}
        \begin{subfigure}[t]{0.289\linewidth}
        \centering
            \includegraphics[width=0.9\linewidth]{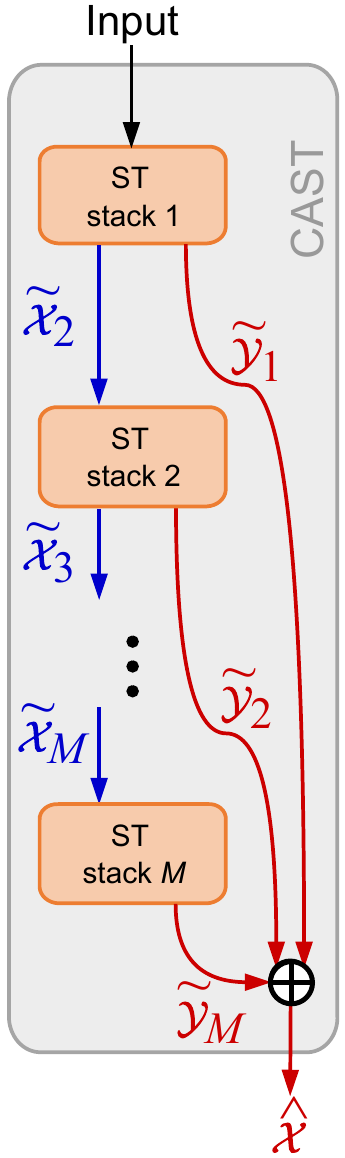}
            \caption{\texttt{CAST} overview.}
            \label{subfig:cast_overview}
        \end{subfigure}
        \caption{Main model (\texttt{CAST}).}
    \label{fig:cast}
\end{figure}

\textcolor{black}{We propose a hybrid architecture~\cite{DBLP:conf/ijcai/WuPLJZ19} that combines attention and convolution \cite{DBLP:conf/nips/DaiLLT21}, namely \textbf{\texttt{C}}onvolution \textbf{\texttt{A}}ttention for \textbf{\texttt{S}}patio-\textbf{\texttt{T}}emporal traffic forecasting (\texttt{CAST}) as shown in Fig.~\ref{fig:cast}.
Intuitively, the convolution layers can be viewed as filters that produce high-level features to be fed into the attention module, which is treated as an implicit memory, capable of storing a representation of complex knowledge. By doing so, the model can swiftly adapt to new patterns using limited data as the RN evolves, while still preserving intricate historical knowledge.
Further, convolution focuses on local patterns such as seasonalities, variations w.r.t. the temporal aspect, and close-neighbor-nodes w.r.t. the spatial aspect. In contrast, attention focuses on global patterns. By combining the two, we exploit both local and global information to improve forecasting accuracy.}
\texttt{CAST} consists of spatio-temporal \textit{stack}s (Fig.~\ref{subfig:cast_overview}). 
Each such stack consists of spatio-temporal \textit{block}s (Fig.~\ref{subfig:cast_stack}). 
Each block is a basic component that performs spatial and temporal computation (Fig.~\ref{subfig:cast_block}). 

\vspace{-0.5em}
\subsubsection{Spatio-temporal Block}
A spatio-temporal (\texttt{ST}) block consists of three components, including a spatial component to learn spatial information, a temporal component to learn temporal information, and a component to compute forecasting and backcasting values.
The spatial and temporal components combine convolution and attention layers with the constraint that the attention layers must follow the convolution layers.
This ordering allows the model to converge faster 
(i.e., capture more high-level features and then memorize these by only training on a small amount of data)~\cite{DBLP:conf/nips/DaiLLT21,DBLP:conf/iccv/WuXCLDY021}.
Next, we describe the operation of the $l$-th block in detail.
The first \texttt{ST} block is a special case and receives the graph signal $\mathcal{X}^{\pi}_{l} \equiv \mathcal{X}^{\pi}, \pi=1$ or $\mathcal{X}^{\pi}_{l} \equiv _{\Delta}\mathcal{X}^{\pi}, \pi>1$ as input.
The inputs $\mathcal{X}_{l}$ to the remaining \texttt{ST} blocks are the residual outputs from the previous blocks.
The input is first fed into the spatial component, which performs spatial convolution and attention.
For the spatial convolution, we employ \texttt{ChebnetII}~\cite{DBLP:journals/corr/HeWW22}, which is an improved version of \texttt{ChebNet}~\cite{DBLP:conf/nips/DefferrardBV16} that
\textcolor{black}{has more expressive capabilities than \texttt{GCN}~\cite{DBLP:conf/iclr/KipfW17}.} 
More specifically, the computation is defined as follows.
First, Laplacian matrices $\mathbf{L}^{\pi}$ and $_{\Delta}\mathbf{L}^{\pi}$ are computed as shown in Eq.~\ref{eq:laplacian}.

\vspace{-1.5em}
\begin{equation}
    \begin{cases}
        \displaystyle
        \mathbf{L}^{\pi} = \mathbf{I}^{\pi} - \mathbf{D}^{{\pi}^{-\frac{1}{2}}} \mathbf{A}^{\pi} \mathbf{D}^{{\pi}^{-\frac{1}{2}}}, \text{if } \pi = 1\\
        _{\Delta}\mathbf{L}^{\pi} = _{\Delta}\mathbf{I}^{\pi} - _{\Delta} \mathbf{D}^{{\pi}^{-\frac{1}{2}}} \;_{\Delta} \mathbf{A}^{\pi} \;_{\Delta}\mathbf{D}^{{\pi}^{-\frac{1}{2}}}, \text{if } \pi > 1
    \end{cases}
    \label{eq:laplacian}
\end{equation}
\vspace{-0.5em}

\noindent Here, $\mathbf{D}^{\pi} \in \mathbb{R}^{N^{\pi} \times N^{\pi}}$ and $_{\Delta}\mathbf{D}^{\pi} \in \mathbb{R}^{_{\Delta}N^{\pi} \times _{\Delta}N^{\pi}}$ are degree matrices, $\mathbf{A}^{\pi} \in \mathbb{R}^{N^{\pi} \times N^{\pi}}$ and $_{\Delta}\mathbf{A}^{\pi} \in \mathbb{R}^{_{\Delta}N^{\pi} \times _{\Delta}N^{\pi}}$ are adjacency matrices, and $\mathbf{I}^{\pi} \in \mathbb{R}^{N^{\pi} \times N^{\pi}}$ and $_{\Delta}\mathbf{I}^{\pi} \in \mathbb{R}^{_{\Delta}N^{\pi} \times _{\Delta}N^{\pi}}$ are identity matrices.
We follow existing studies to construct the adjacency matrices, i.e., generating $\mathbf{A}^{\pi}$ and $_{\Delta}\mathbf{A}^{\pi}$, using distances between entities~\cite{DBLP:conf/iclr/LiYS018}. 
We thus compute the RN distances to construct $\mathbf{A}^{\pi}$ and $_{\Delta}\mathbf{A}^{\pi}$ by using a Gaussian kernel. 
The weight of the edge between sensors $i$ and $j$ is defined as follows.

\vspace{-1.5em}
{
\begin{align}
    \begin{cases}
        \displaystyle \mathbf{A}^{\pi}_{ij} = \exp{-\frac{\mathit{dist}(V_{i}, V_{j})^{2}}{\sigma^2}}, \text{if } \pi = 1\\
        \displaystyle _{\Delta}\mathbf{A}^{\pi}_{ij} = \exp{-\frac{\mathit{dist}(_{\Delta}V_{i}, _{\Delta}V_{j})^{2}}{\sigma^2}}, \text{if } \pi > 1
    \end{cases}
    \label{eq:adjacency}
\end{align}
}

\noindent Here, $\mathit{dist}(V_{i}, V_{j})$ and $\mathit{dist}(_{\Delta}V_{i}, _{\Delta}V_{j})$ are the distances between entities $i$ and $j$, and $\sigma$ is the standard deviation of the distances. 
We set $\mathbf{A}^{\pi}_{ij}=0$ and $_{\Delta}\mathbf{A}^{\pi}_{ij}=0$ if they are below a threshold $\epsilon$.
Next, a rescaled Laplacian matrix $\mathbf{\hat{L}}^{\pi} \in \mathbb{R}^{N^{\pi} \times N^{\pi}}$ and $_{\Delta}\mathbf{\hat{L}}^{\pi} \in \mathbb{R}^{_{\Delta}N^{\pi} \times _{\Delta}N^{\pi}}$ are computed as shown in Eq.~\ref{eq:simple_laplacian}.

\vspace{-0.5em}
\begin{equation}
    \begin{cases}
        \displaystyle \hat{\mathbf{L}}^{\pi} = \frac{2 \mathbf{L}^{\pi}}{\lambda^{\pi}_{max}} - \mathbf{I}, \text{if } \pi = 1\\
        \displaystyle _{\Delta} \hat{\mathbf{L}}^{\pi} = \frac{2 _{\Delta} \mathbf{L}^{\pi}}{_{\Delta}\lambda^{\pi}_{max}} - _{\Delta} \mathbf{I}, \text{if } \pi > 1
    \end{cases}
    \label{eq:simple_laplacian}
\end{equation}
\vspace{-0.5em}

\noindent Here, $\lambda^{\pi}_{max}$ and $_{\Delta}\lambda^{\pi}_{max}$ are the maximum of eigenvalues of $\mathbf{L}^{\pi}$ and $_{\Delta}\mathbf{L}^{\pi}$, respectively. 
For a graph signal with $F_{\text{in}}$ channels, $\mathcal{X}^{\pi}_{l} \in \mathbb{R}^{N^{\pi} \times F_{\text{in}} \times P}$ or $_{\Delta}\mathcal{X}^{\pi}_{l} \in \mathbb{R}^{_{\Delta}N^{\pi} \times F_{\text{in}} \times P}$, the output of the spatial convolution is computed as shown in Eq.~\ref{eq:chebnet_2}.

\vspace{-1.5em}
\begin{equation}
    \mathcal{H}_{1} = 
    \begin{cases}
        \displaystyle
        \frac{2}{O+1} \sum^{O}_{o=0}  \sum^{O}_{q=0} \gamma_{q} \mathrm{P}_{o}\left(\Gamma\left(q,O\right)\right)\mathrm{P}_{o}\left(\hat{\mathbf{L}}\right)\mathcal{X}^{\pi}_{l}, \text{if } \pi = 1&\\
        \displaystyle
        \frac{2}{O+1} \sum^{O}_{o=0}  \sum^{O}_{q=0} \gamma_{q} \mathrm{P}_{o}\left(\Gamma\left(q,O\right)\right)\mathrm{P}_{o}\left(\hat{\mathbf{L}}\right)_{\Delta}\mathcal{X}^{\pi}_{l}, \text{if } \pi > 1&
    \end{cases}
    \label{eq:chebnet_2}
\end{equation}
\vspace{-0.5em}

\noindent Here, $\gamma_{q} \in \mathbb{R}^{F_{\text{in}} \times F_{1}}$ is a learnable weight matrix, $\displaystyle\Gamma\left(q,O\right)=\mathrm{cosin}\left(\frac{\pi(q+\frac{1}{2})}{O+1}\right)$, and $\mathrm{P}_{o}(\cdot)$ is a Chebyshev function, which is defined as a recursive function as follows.

\vspace{-0.5em}
\begin{equation}
    \begin{cases}
        \mathrm{P}_{o}(x) = 2\mathrm{P}_{o-1}(x) + \mathrm{P}_{o-2}(x) \\
        \mathrm{P}_{1}(x) = x \\
        \mathrm{P}_{0}(x) = 1 \\
    \end{cases}
    \label{eq:chebyshev}
\end{equation}
\vspace{-0.5em}

The output of the spatial convolution $\mathcal{H}_{1} \in \mathbb{R}^{N^\pi \times F_{1} \times P}, \pi = 1$ or $\in \mathbb{R}^{{_{\Delta}N}^\pi \times F_{1} \times P}, \pi > 1$  is then fed into the spatial attention.
For simplicity, we denote $\mathcal{H}_{1} \in \mathbb{R}^{N^\pi \times F_{1} \times P}, \pi = 1$ or $\in \mathbb{R}^{{_{\Delta}N}^\pi \times F_{1} \times P}, \pi > 1$ using the unified notation $\mathcal{H}_{1} \in \mathbb{R}^{(N^\pi || _{\Delta}N^\pi) \times F_{1} \times P}$ from now.
We employ Graph Attention Network (\texttt{GAT})~\cite{DBLP:conf/iclr/VelickovicCCRLB18} for the spatial attention.
First, two matrices $\mathbf{W}_{s_{\text{key}}}, \mathbf{W}_{s_{\text{query}}} \in \mathbb{R}^{(N^\pi || _\Delta N^\pi) \times 1 \times P}$ are computed from $\mathbf{W}_{h} \in \mathbb{R}^{(N^\pi || _\Delta N^\pi) \times F_{2} \times P}$.

\vspace{-0.5em}
\begin{equation}
    \mathbf{W}_{h} = \mathcal{H}_{1} \mathbf{W}_{s}; \quad
    \mathbf{W}_{s_{\text{key}}} = \mathbf{W}_{h} \mathbf{a}_{s_{\text{key}}}; \quad
    \mathbf{W}_{s_{\text{query}}} = \mathbf{W}_{h} \mathbf{a}_{s_{\text{query}}}
    \label{eq:spatial_key_query}
\end{equation}
\vspace{-0.5em}

\noindent Here, $\mathbf{W}_{s} \in \mathbb{R}^{F_1 \times F_2}$, $\mathbf{a}_{s_{\text{key}}} \in \mathbb{R} ^{F_2 \times 1}$, and $\mathbf{a}_{s_{\text{query}}} \in \mathbb{R}^{F_2 \times 1}$ are learnable parameters.

Then, a spatial attention matrix $\mathbf{E}_{s} \in \mathbb{R}^{(N^{\pi} || _{\Delta} N^{\pi}) \times (N^{\pi} || _{\Delta} N^{\pi})}$ is constructed as shown in Eq.~\ref{eq:spatial_attention_matrix}. The spatial attention matrix plays the role of specifying the importance between all the nodes in the RN.

\vspace{-0.5em}
\begin{equation}
    \mathbf{E}_{s} = \mathbf{W}_{s_{\text{key}}} \mathbf{W}_{s_{\text{query}}}^\top
    \label{eq:spatial_attention_matrix}
\end{equation}
\vspace{-0.5em}

A normalized spatial attention matrix $\mathbf{E}'_{s} \in \mathbb{R}^{(N^{\pi} || _{\Delta} N^{\pi}) \times (N^{\pi} || _{\Delta} N^{\pi})}$ is computed from each element $\mathbf{E}_{s_{i,j}}$.

\vspace{-0.5em}
\begin{equation}
    \mathbf{E}'_{s} = \frac{\mathrm{exp}(\mathbf{E}_{s_{i,j}})}{\sum_{j}^{N}\mathrm{exp}(\mathbf{E}_{s_{i,j}})}
    \label{eq:normalized_spatial_attention_matrix}
\end{equation}
\vspace{-0.5em}

The output of spatial attention $\mathcal{H}_{2} \in \mathbb{R}^{(N^\pi || _{\Delta}N^\pi) \times F_{2} \times P}$ is computed as follows.

\vspace{-0.5em}
\begin{equation}
    \mathcal{H}_{2} = \mathrm{ReLU}(\mathbf{E}'_{s} \mathbf{W}_{h})
    \label{eq:spatial_attention_output}
\end{equation}
\vspace{-0.5em}

\noindent Here, $\mathrm{ReLU}$ is the rectified linear unit activation function~\cite{DBLP:conf/icml/NairH10}.
The spatial attention shown in Eq.~\ref{eq:spatial_attention_output} can be stabilized by using the multi-head mechanism~\cite{DBLP:conf/iclr/VelickovicCCRLB18}, where multiple spatial attentions work together in parallel. 

\vspace{-0.5em}
\begin{equation}
    \mathcal{H}_{2} = \Big{\Vert}^{U}_{u=1} \mathrm{ReLU}(\mathbf{E}'^{u}_{s} \mathbf{W}^{u}_{h})
    \label{eq:spatial_attention_output_multi}
\end{equation}
\vspace{-0.5em}

\noindent Here, $\Big{\Vert}$ denotes the concatenation operator, $U$ is the number of spatial attention heads, $\mathbf{E}'^{u}_{s} \in \mathbb{R}^{(N^{\pi} || _{\Delta} N^{\pi}) \times (N^{\pi} || _{\Delta} N^{\pi})}$ is the $u$-th normalized spatial attention matrix, and $\mathbf{W}^{u}_{h} \in \mathbb{R}^{(N^{\pi} || _{\Delta} N^{\pi}) \times \frac{F_2}{U} \times P}$, where $\mathbf{W}^{u}_{h} = \mathcal{H}_{1}\mathbf{W}^{u}$ is the weight matrix of the $u$-th spatial attention heads.

The output of the spatial attention is also the output of the spatial component, which is fed into the temporal component.
In the temporal component, the input is first fed into a \texttt{TCN} with $l$ layers. 
We use dilated causal convolution for \texttt{TCN}. 
Dilated causal convolution works by convoluting $V$ attribute vectors from $B$ different timestamps, where the attribute vectors to be convoluted are $d$ timestamps apart. 
At different layers, $B$ is often the same, while $d$ may be different. 

\begin{figure}[H]
    \vspace{-1.0em}
    \centering
    \includegraphics[width=0.6\linewidth]{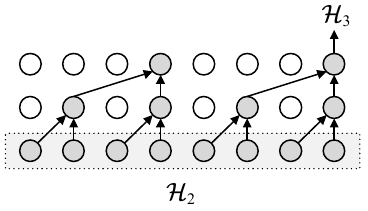}
    \caption{Temporal convolution.}
    \label{fig:tcn}
    \vspace{-1.0em}
\end{figure}

For example, Figure~\ref{fig:tcn} shows a \texttt{TCN} with $B=2$, meaning that the dilated causal convolution always convolutes two attribute vectors from two timestamps. 
The dilation factors $d$ in the 1\textsuperscript{st}, 2\textsuperscript{nd}, and 3\textsuperscript{rd} layers are 1, 2, and 4, respectively. 
Thus, in the 1\textsuperscript{st} layer, the two attributes vectors to be convoluted are 1 timestamp apart; in the 2\textsuperscript{nd} layer, 2 timestamps apart; in the 3\textsuperscript{rd} layer, the 4 timestamps apart.

The temporal convolution is defined as follows.

\vspace{-0.5em}
\begin{equation}
    \mathcal{H}_{3} = \mathcal{H}_{2} \star_{t} \mathbf{W}_{t_{1}} = \sum_{F_{3}=1}^{F_{3}} \sum_{b=1}^{B} \mathbf{W}[b, F_{3}] \odot \mathcal{H}_{2_{t-d \times (b-1)}}[F_{3}],
    \label{eq:temporal_convolution}
\end{equation}
\vspace{-0.5em}

\noindent where, $\star_{t}$ represents a convolutional operator at timestamp $t$, $\odot$ is Hadamard product (i.e., element-wise multiplication), and $\mathbf{W}_{t_{1}} \in \mathbb{R}^{B \times F_{3}}$ is the convolution filter.
We compute Eq.~\ref{eq:temporal_convolution} with a number of filters in the filter bank $\mathcal{W}_{t_{1}}$.

The output of temporal convolution, $\mathcal{H}_{3} \in \mathbb{R}^{(N^\pi || _\Delta N^\pi) \times F_{3} \times P}$, is fed into temporal attention.
The temporal attention can be viewed as the importance score between all timestamps.
First, the temporal attention matrix $\mathbf{E}_{t} \in \mathbb{R}^{P \times P}$ is computed as follows.

\vspace{-0.5em}
{
    \begin{align}
        \mathbf{E}_{t} = \mathbf{W}_{t_{2}}  \sigma( (\mathcal{H}_{3}^\top \mathbf{W}_{t_\text{key}}) \mathbf{W}_{t_{3}} (\mathbf{W}_{t_\text{query}} \mathcal{H}_3 ) + \mathbf{b}_{t})
    \end{align}
}

\noindent Here, $\mathbf{W}_{t_{2}}, \mathbf{b}_t \in \mathbb{R}^{P \times P}$, $\mathbf{W}_{t_\text{key}} \in \mathbb{R}^{F_{3} \times F_{4}}$, $\mathbf{W}_{t_{3}} \in \mathbb{R}^{F_{4}}$, and $\mathbf{W}_{t_\text{query}} \in \mathbb{R}^{F_{3}}$ are learnable weight matrices, and $\mathbf{b}_{t}$ is the bias.
A normalized temporal attention matrix $\mathbf{E}'_{t} \in \mathbb{R}^{P \times P}$ is computed from each element $\mathbf{E}_{t_{i,j}}$.

\vspace{-0.5em}
\begin{equation}
    \mathbf{E}'_{t} = \frac{\mathrm{exp}(\mathbf{E}_{t_{i,j}})}{\sum_{j}^{P}\mathrm{exp}(\mathbf{E}_{t_{i,j}})}
    \label{eq:normalized_temporal_attention_matrix}
\end{equation}
\vspace{-0.5em}

Then, the output of temporal attention $\mathcal{H}_{4} \in \mathbb{R}^{(N^\pi || _\Delta N^\pi) \times F_{4} \times P}$ is computed as follows.

\vspace{-0.5em}
\begin{equation}
    \mathcal{H}_{4} = \mathcal{H}_{3} \mathbf{E}'_{t}
\end{equation}
\vspace{-0.5em}

The output of temporal attention $\mathcal{H}_{4}$ is then conducted as a residual connection as shown in Eq.~\ref{eq:residual_block}.

\vspace{-0.5em}
\begin{equation}
    \mathcal{H}_{4} = \mathcal{H}_{4} + \mathcal{X}^{\pi}_{l}
    \label{eq:residual_block}
\end{equation}
\vspace{-0.5em}
    
Then the output of the residual connection is synchronously fed into the forecast and the backcast convolution to produce $\mathcal{Y}_{l} \in \mathbb{R}^{(N^\pi _\Delta N^\pi) \times p \times H}$ and $\mathcal{V}_{l} \in \mathbb{R}^{(N^\pi || _\Delta N^\pi) \times F_{in} \times P}$, which are the forecast and the backcast of block $l$, respectively.

\vspace{-0.5em}
{
    \begin{align}
        \mathcal{Y}_{l} &= \mathcal{H}_{4} \star \mathbf{W}_\text{forecast} + \mathbf{b}_\text{forecast}
        \label{eq:forecast}\\
        \mathcal{V}_{l} &= \mathcal{H}_{4} \star \mathbf{W}_\text{backcast} + \mathbf{b}_\text{backcast}
        \label{eq:backcast}
    \end{align}
}
\vspace{-0.5em}

\noindent Here, $\star$ indicates 2D convolution operator; $\mathbf{W}_\text{forecast} \in \mathbb{R}^{F_4 \times p}$ and $\mathbf{b}_\text{forecast} \in \mathbb{R}^{p}$ are the learnable weight matrix and bias vector of the forecast convolution component, where $p$ is the number of output feature; and $\mathbf{W}_\text{backcast} \in \mathbb{R}^{F_4 \times F_{in}}$ and $\mathbf{b}_\text{backcast} \in \mathbb{R}^{F_{in}}$ are the learnable weight matrix and bias vector of the backcast convolution component.

\subsubsection{Spatio-temporal Stack}
An \texttt{ST} stack is the upper level of an \texttt{ST} block.
More specifically, an \texttt{ST} stack consists of a sequence of $L$ \texttt{ST} blocks that connect sequentially in a novel doubly residual structure (see Fig.~\ref{subfig:cast_overview}). 
This structure has two residual branches, one for the forecast of all the \texttt{ST} blocks in an \texttt{ST} stack and one for the backcast output of the previous \texttt{ST} block. 
\textcolor{black}{
The backcast output can be interpreted as the portion of the information that is not needed for the forecast job of the following \texttt{ST} blocks~\cite{DBLP:conf/iclr/OreshkinCCB20}, making it easier for the following blocks to process the signal.
The forecast output summarizes the final prediction of \texttt{ST} blocks, providing an implicit ensemble architecture, which demonstrates better performance compared to single models~\cite{DBLP:journals/pvldb/CamposKGHZYJ21, DBLP:conf/ijcai/KieuYGJ19}.}
Next, we describe the operation of the $m$-th \texttt{ST} stack in detail.
As described before, the output of the $l$-th block in a stack is $\mathcal{Y}_{l}$ and $\mathcal{V}_{l}$, as shown in Eqs.~\ref{eq:forecast} and \ref{eq:backcast}.
Then, $\mathcal{V}_{l}$ is used for computing the input of the next block, $\mathcal{X}^{\pi}_{l+1} \in \mathbb{R}^{(N^{\pi} || _\Delta N^{\pi} ) \times F_{\text{in}} \times P}$ as shown in Eq.~\ref{eq:residual_backcast}.

\vspace{-0.5em}
\begin{equation}
    \mathcal{X}^{\pi}_{l+1} = \mathcal{X}^{\pi}_{l} - \mathcal{V}_{l}
    \label{eq:residual_backcast}
\end{equation}
\vspace{-0.5em}

The output of the last \texttt{ST} block in an \texttt{ST} stack (i.e., $\mathcal{X}_{L+1}$) is also the residual for the next \texttt{ST} stack $\tilde{\mathcal{X}}_{m+1}$, i.e., $\tilde{\mathcal{X}}_{m+1} \equiv \mathcal{X}_{L+1}$.
The forecast residual $\tilde{\mathcal{Y}}_{m} \in \mathbb{R}^{(N^\pi || _\Delta N^\pi) \times p \times H}$ is computed by summing the forecast output of all \texttt{ST} blocks in the \texttt{ST} stack as shown in Eq.~\ref{eq:residual_forecast}.

\vspace{-0.5em}
\begin{equation}
    \tilde{\mathcal{Y}}_{m} = \sum_{l=1}^{L}\mathcal{Y}_{l}
    \label{eq:residual_forecast}
\end{equation}
\vspace{-0.5em}

\subsubsection{\texttt{CAST} Model}
The entire \texttt{CAST} model consist of $M$ \texttt{ST} stacks
As described before, the $m$-th \texttt{ST} stack produces two outputs, $\tilde{\mathcal{Y}}_{m}$ and $\tilde{\mathcal{X}}_{m}$. The first output is the stack forecasting $\mathcal{\tilde{Y}}_{i} \in \mathbb{R}^{(N^\pi || _\Delta N^\pi) \times p \times H}$, which contributes to the model forecast. The second output is the stack residual $\tilde{\mathcal{X}}_{m} \in \mathbb{R}^{(N^\pi || _\Delta N^\pi) \times F_{in} \times P}$, which is fed to the next stack. The global forecast $\mathcal{\hat{X}} \in \mathbb{R}^{(N^\pi || _\Delta N^\pi) \times p \times H}$ is the final output of the model, which is computed by aggregating the forecasts of all the stacks as follows.

\vspace{-0.5em}
\begin{equation}
    \mathcal{\hat{X}} = \frac{1}{M} \sum_{m=1}^{M} \mathcal{\tilde{Y}}_{m}
    \label{eq:stack_residual}
\end{equation}
\vspace{-0.5em}

\subsection{Continual Learning Module}
\label{subsec:continual_module}
The continual learning module is responsible for leveraging old knowledge, helping to reduce the training complexity caused by the evolution of RNs.
First, we note that the motivation for reducing the complexity is to exploit the knowledge of the model trained on the historical RNs. 
With this knowledge, the model does not need to be trained on the entire RN subsequently. 
Rather, the model only needs to be trained on evolved parts.
To achieve this, we propose two strategies. 
(i) The model does not need to be trained on the data of nodes, whose traffic patterns do not change much after an evolution. 
The model only uses a little data of these strongly unchanged nodes for consolidating the knowledge that is learned from previous periods.
(ii) The model has to be trained on the data of nodes whose traffic patterns are affected strongly by an evolution of the RN.
The learned spatio-temporal representations of such nodes are no longer useful, so the new data from these nodes is used for updating the model.
We proceed to introduce the continual learning module that adopts the proposed strategies.
The core step is to select the nodes that are used for training.
After the evolution of the RN, the continual learning module first picks all newly added and removed nodes. Assuming those old nodes close to added and removed nodes are strongly affected, the module also includes nodes adjacent to newly added and removed nodes. Simultaneously, the continual module addresses the impact of newly added and removed edges by selecting nodes adjacent to newly added and removed edges. 
The selected nodes and their edges are used for constructing the updated part $_\Delta G$.
Then, the module identifies the most stable old nodes, which are used to revise the historical knowledge as an alternative to training on the entire RN. The most stable nodes are stored in a consolidation memory buffer $\mathcal{B}_c$, and this buffer is used to revise the model.
The stability of a node is determined by the change in the data histogram on that node before and after the graph evolves.
Specifically, we use the data from the last $\tau$ timestamps of period $\pi-1$ and of period $\pi$ to produce the histogram at period $\pi-1$ (denoted as $\mathbf{H}^{\pi-1}$) and at period $\pi$ (denoted as $\mathbf{H}^{\pi}$), respectively. 
After obtaining $\mathbf{H}^{\pi-1}$ and $\mathbf{H}^{\pi}$, we use the Earth mover's distance (\textrm{EMD}) to measure the stability of a node as shown in Eq.~\ref{eq:earth_move_divergence}.

\vspace{-1.0em}
\begin{equation}
    \mathrm{EMD}(\mathbf{H}^{\pi-1}, \mathbf{H}^{\pi}) = \min \limits_{\mathbf{Q} \ge 0} \sum \limits_{i}^{\lvert \mathbf{H}^{\pi-1} \rvert} \sum \limits_{j}^{\lvert \mathbf{H}^{\pi} \rvert} \mathbf{Q}_{i,j}  \lVert \mathbf{H}^{\pi-1}_i - \mathbf{H}^{\pi}_j \rVert
    \label{eq:earth_move_divergence}
\end{equation}
\vspace{-1.0em}

\noindent Here, $\displaystyle\mathbf{Q} \in \mathbb{R}^{\lvert \mathbf{H}^{\pi-1} \rvert \times \lvert \mathbf{H}^{\pi} \rvert}$ is the optimal transport plan matrix where $\displaystyle\sum \limits_{i} \mathbf{Q}_{i,j} = \frac{1}{\lvert \mathbf{H}^{\pi} \rvert}$ and $\displaystyle\sum \limits_{j} \mathbf{Q}_{i,j} = \frac{1}{\lvert \mathbf{H}^{\pi - 1} \rvert}$. Intuitively, EMD is defined by a minimal transportation ``cost'' to convert histogram $\mathbf{H}^{\pi-1}$ into histogram $\mathbf{H}^{\pi}$.
In simple terms, $\mathbf{H}^{\pi-1}$ and $\mathbf{H}^{\pi}$ can be seen as representations of traffic flow distributions before and after an RN evolution, and the \textrm{EMD} is capable of quantifying the difference between these two distributions, even if there are significant shifts in the distributions (e.g., a highway is constructed nearby, which may lead to substantial traffic flow changes). The nodes with the lowest \textrm{EMD} are stable and can be used as rehearsal nodes for the traffic prediction model.
The continual learning module selects the nodes with the lowest \textrm{EMD} and saves these most stable nodes in a consolidation memory buffer $\mathcal{B}_c$.
Further, after evolving, there are unstable nodes (i.e., the patterns of these nodes have changed considerably).
If the model keeps the old and now inaccurate space-time correlations of these nodes from the historical RN to conduct forecasting, the model's performance will decrease significantly.
Therefore, it is necessary to select unstable nodes so that the model is forced to re-learn these nodes with new data. 
Similar to selecting the stable nodes with the lowest \textrm{EMD}, the continual learning module selects the nodes with the highest \textrm{EMD} and saves these unstable nodes in an update memory buffer $\mathcal{B}_u$, thereby treating these old nodes as newly added nodes.

\begin{algorithm}[t]
    \caption{Training procedure}
    \label{alg:rehearsal_algorithm}
        \SetKwInOut{Input}{Input}
        \SetKwInOut{Output}{Output}
        \SetCustomAlgoRuledWidth{1cm}
        \SetArgSty{textnormal}
        \Input{model $f_{\theta}^{\pi - 1}$ trained on $G^{\pi-1}$, data $\mathcal{X}^{\pi}$ of $G^{\pi}$, data $\mathcal{X}^{\pi-1}$ of $G^{\pi - 1}$, period $\tau$, \textcolor{black}{consolidation buffer size $|\mathcal{B}_c|$, update buffer size $|\mathcal{B}_u|$}}
        \Output{model $f_{\theta}^{\pi}$.}
            $\textit{EMD\_list} = [\:]; \mathcal{B}_u= [\:], \mathcal{B}_c= [\:]$\;
            $_\Delta G^{\pi} \leftarrow G^{\pi} - G^{\pi - 1}$;\\
            \For{each node i $\in V^{\pi - 1}$}{
                $\mathcal{E}^{\pi-1} \leftarrow$ Select last $\tau$ timestamps from $\mathcal{X}^{\pi-1, i}$;\\ 
                $\mathcal{E}^{\pi} \leftarrow$ Select last $\tau$ timestamps from $\mathcal{X}^{\pi, i}$;\\ 
                $\mathbf{H}^{\pi-1} \leftarrow$ Build histogram for $\mathcal{E}^{\pi-1}$;\\
                $\mathbf{H}^{\pi} \leftarrow$ Build histogram for $\mathcal{E}^{\pi}$;\\
                $\textit{div} \leftarrow \mathrm{EMD}(\mathbf{H}^{\pi-1}, \mathbf{H}^{\pi})$;\\ 
                \textit{EMD\_list} $\leftarrow$ \textit{EMD\_list} $\cup$ (div);\\
            }
            \textrm{sort}(\textit{EMD\_list});\\
            Select $|\mathcal{B}_c|$ nodes has lowest \textrm{EMD} and store to $\mathcal{B}_c$;\\
            Select $|\mathcal{B}_u|$ nodes has highest \textrm{EMD} and store to $\mathcal{B}_u$;\\
            \textcolor{black}{
                $f_{\theta}^{\pi} \leftarrow \text{ training } f_{\theta}^{\pi - 1} \text{ with } \mathcal{B}_c, \mathcal{B}_u \text{, and }_\Delta G^{\pi}$ 
            }
\end{algorithm}

We proceed to introduce Algorithm~\ref{alg:rehearsal_algorithm} that describes the training process that incorporates both the consolidation of historical knowledge and the update of significantly altered nodes.
Algorithm~\ref{alg:rehearsal_algorithm} takes the model $f_{\theta}^{\pi - 1}$ trained on $G^{\pi-1}$, the data $\mathcal{X}^{\pi}$ of $G^{\pi}$, the data $\mathcal{X}^{\pi-1}$ of $G^{\pi - 1}$, the number of bins $\alpha$, the number of timestamps $\tau$, the consolidation buffer size $|\mathcal{B}_c|$, and the update buffer size $|\mathcal{B}_u|$ as the input to produce the model $f_{\theta}^{\pi}$.
First, the \textit{EMD\_list} and the two buffers, $\mathcal{B}_c$ and $\mathcal{B}_u$, are initialized (line 1).
Next, the algorithm constructs the graph from the evolved part of RN, denoted as $_\Delta G^{\pi}$, to update the model (line 2).
Then, the historical and current data of each node in the historical RN $G^{\pi-1}$ are used to compute two data histograms (lines 3--7).
The \textrm{EMD} is then computed based on the two obtained histograms (line 8).
The divergence result is added to the \textit{EMD\_list} (line 9).
After that, the \textit{EMD\_list} is sorted (line 10).
Next, the consolidation buffer $\mathcal{B}_c$ is filled with the nodes that have the lowest \textrm{EMD}, and the update buffer $\mathcal{B}_u$ is filled with the nodes that have the highest \textrm{EMD} (lines 11--12).
Finally, the model for period $\pi$, i.e., $f_{\theta}^{\pi}$, is produced by training the model for period $\pi - 1$, i.e., $f_{\theta}^{\pi - 1}$, on the data of evolved nodes and the buffer (line 13). 

In summary, the continual learning module aims to capture updates and adapt to evolving RN data, hence avoiding complete re-training. The continual learning module only affects the training runtime and does not affect the real-time forecasting capability of traffic forecasting models. The forecasting runtime depends on the inference time of traffic forecasting models. After being updated and adapted, a model can forecast in real-time.

\subsection{Objective Function}
To train the proposed framework, we use the Huber loss~\cite{Huber1992} as the main objective function.
This loss has demonstrated better results~\cite{DBLP:conf/icde/CirsteaYGKP22} for regression tasks than the traditional mean square error, especially when the training data contains noise and outliers that the Huber loss is less sensitive to.
The main objective function is defined as follows.

\vspace{-0.5em}
\begin{equation}
    \mathcal{L}_{\text{main}} = \begin{cases}\displaystyle{\frac {1}{2}} \left ( \hat{\mathcal{X}}-\mathcal{X} \right ) ^{2} ,\text {if} \left |\hat{\mathcal{X}}-\mathcal{X} \right | \leq \delta \\
    \delta \left ( \left |\hat{\mathcal{X}}-\mathcal{X} \right |-\displaystyle{\frac {1}{2}}\delta \right),\text { otherwise}\end{cases}
\end{equation}
\vspace{-0.5em}

In addition, to better support the rehearsal algorithm that was described in the previous section, we apply the elastic weight consolidation method to the training procedure.
This method measures the importance of each parameter $\theta_{i}$ in the parameter set $\theta$.
This allows us to estimate which parameters are the important ones for forecasting in the historical RN.
Then, we can skip updating these parameters and focus on updating the less important parameters.
By doing so, we achieve two benefits: (i) the model can reduce the knowledge-forgetting problem and (ii) the model can increase the ability to learn new knowledge.
Elastic weight consolidation is a regularization method, which is defined as the objective function shown in Eq.~\ref{eq:elastic_weight_consolidation}.

\vspace{-0.5em}
    \begin{equation}
        \mathcal{L}_{\text{regularization}} = \sum_{i}\mathbf{F}_{i} (\theta^{\pi}[i] -\theta^{\pi-1}[i])^{2}
        \label{eq:elastic_weight_consolidation}
    \end{equation}
\vspace{-0.5em}

\noindent Here, $\theta^{\pi-1}[i]$ and $\theta^{\pi}[i]$ are the $i$-th parameter in the parameters set $\theta^{\pi-1}$ and $\theta^{\pi}$, respectively. 
Next, $\mathbf{F}_{i}$ represents the importance of the $i$-th parameter in parameter set $\theta^{\pi-1}$ and is computed using the Fisher information~\cite{DBLP:conf/nips/SoenS21} as follows.

\vspace{-0.5em}
    \begin{equation}
        \mathbf{F} = \frac{1}{|\mathcal{X}^{\pi-1}|}\sum_{\mathbf{X}^{\pi-1} \in \mathcal{X}^{\pi-1}} \frac{\partial \theta^{\pi-1}}{\partial \mathbf{X}^{\pi-1}} \frac{\partial \theta^{{\pi-1}^\top}}{\partial \mathbf{X}^{\pi-1}}
        \label{eq:fisher_information}
    \end{equation}
\vspace{-0.5em}

\noindent Here, $\displaystyle\frac{\partial \theta^{\pi-1}}{\partial \mathbf{X}^{\pi-1}}$ is the partial first-order derivative of $\theta^{\pi-1}$ with respect to $\mathbf{X}^{\pi-1}$.
Finally, the overall objective function of the framework is the sum of the main objective function and the regularization.

\vspace{-1.0em}
    \begin{equation}
        \mathcal{L}_{\text{overall}} = \mathcal{L}_{\text{main}} + \lambda\mathcal{L}_{\text{regularization}},
        \label{eq:overall_objective_function}
    \end{equation}
\vspace{-1.0em}

\noindent where $\lambda$ is the hyperparameter to control the magnitude of the regularization.

\subsection{Complexity Analysis}
\label{sec:comp_analysis}

If a traffic forecasting model does not involve any recursive computation, most of its runtime is spent by the \texttt{GNN}s.
To simplify the complexity term of the proposed framework, we consider the computational complexity mostly based on the number of nodes $N^{\pi}$ of an RN $G^{\pi}$. 
We conduct a comparison between ordinary spatial embedding components, dynamic graph embedding methods, and our framework.
When $\pi=1$, the computational complexity is $\mathcal{O}((N^{\pi})^2)$ for each approach. 
When an RN evolves, i.e., $\pi > 1$, the need for a fully observed topological structure to extract spatial information becomes evident. 
Ordinary spatial embedding components require a re-initialization and a full training process with the data of all nodes in $G^{\pi}$, which increases the computational cost to $\mathcal{O}((N^{\pi})^2)$. 
Likewise, the computational cost of dynamic graph embedding methods is also $\mathcal{O}((N^{\pi})^2)$ since these have to train with the entire RN. 
However, dynamic graph embedding approaches offer a distinct advantage at traffic prediction due to their ability to model the evolution of the RN.
In contrast, our proposed model's complexity is significantly lower, at $\mathcal{O}((_{\Delta} N^{\pi} + \lvert \mathcal{B}_c \cup \mathcal{B}_u\rvert)^2)$. Note that $_{\Delta} N^{\pi} \ll N^{\pi}$, so that $\mathcal{O}((_{\Delta} N^{\pi} + \lvert \mathcal{B}_c \cup \mathcal{B}_u\rvert)^{2}) \ll \mathcal{O}((N^{\pi})^{2})$, where the total size of the buffers $\lvert \mathcal{B}_c \cup \mathcal{B}_u\rvert$ can be predetermined according to the specifications of the training system.
As a result, the framework exhibits lower complexity than the other proposals when the network evolves (i.e. the framework only requires training on newly added nodes and selected old nodes), making it highly suitable for handling continuously evolving topologies.

%% file: sections_arXiv/experiments.tex
\section{Experiments}
\label{sec:experiments}
\subsection{Experimental Settings}
\subsubsection{Datasets}
We experiment with two datasets: \textbf{PEMS03-Evolve} and \textbf{PEMS04-Evolve} that are collected in seven \textit{periods}. For simplicity, we consider periods $\pi$ with a duration of a \textit{month}. Obviously, if we change the duration, e.g., to \textit{quarters}, the proposal is not affected. 
Every month, a new RN is created by adding and removing nodes and edges to and from the RN of the previous month.
The data is collected from metropolitan areas of California by California Transportation Agencies Performance Measurement System (PEMS)\footnote{http://pems.dot.ca.gov/}. Details of the datasets are provided in Tables~\ref{tab:datasets_PEMS03-Evolve} and~\ref{tab:datasets_PEMS04-Evolve}.
Every 5 minutes, a time series observation is generated, representing the average traffic flow without revealing any personally identifiable information such as vehicle identity.
Following existing studies, we split all datasets with a ratio 60\%:20\%:20\% into training sets, validation sets, and testing sets, respectively.

\begin{figure*}[t]
    \centering
        \begin{subfigure}[t]{0.13\linewidth}
        \centering
            \includegraphics[width=0.95\linewidth]{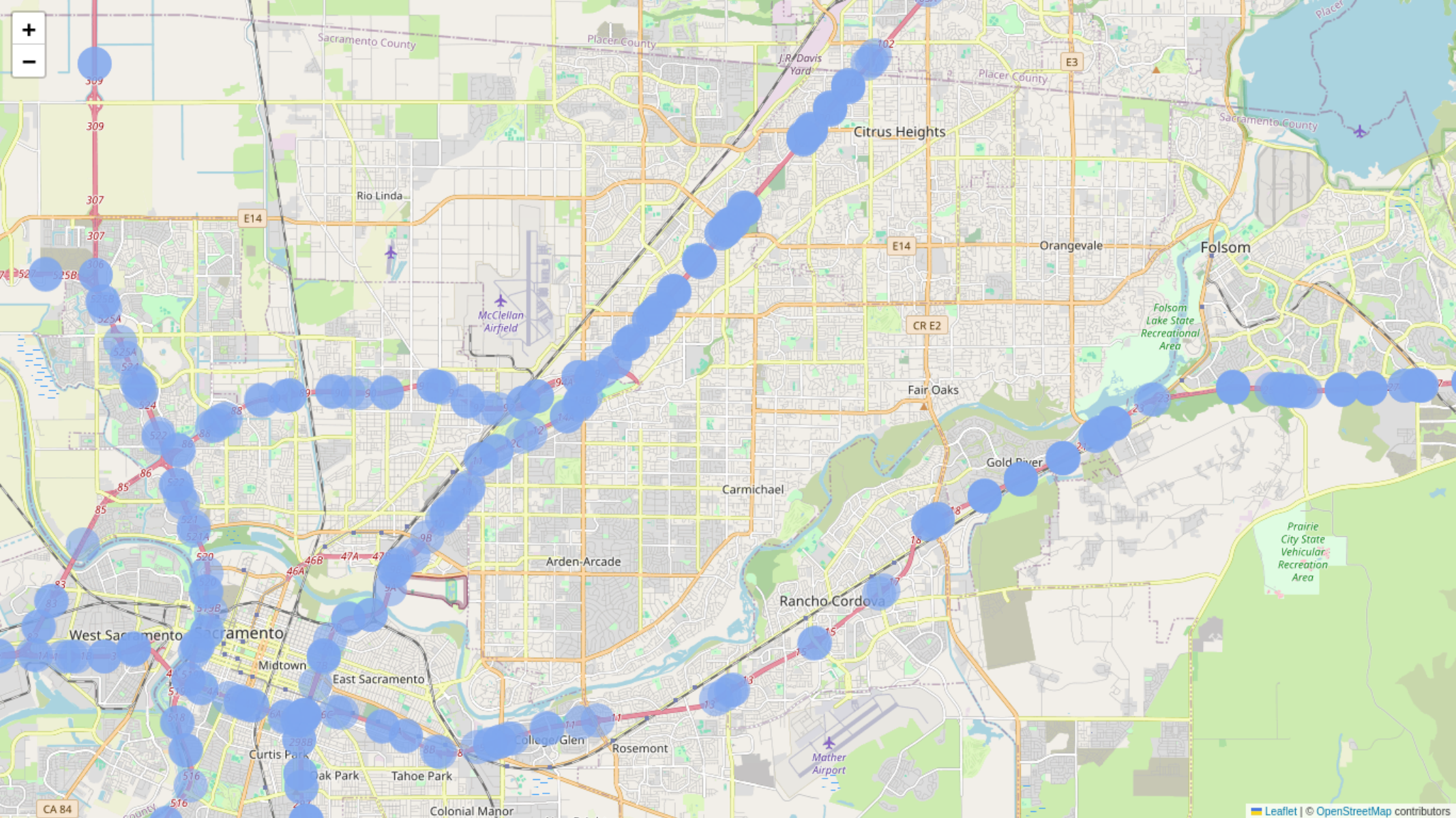}
            \caption{April}
        \end{subfigure}
        \begin{subfigure}[t]{0.13\linewidth}
        \centering
            \includegraphics[width=0.95\linewidth]{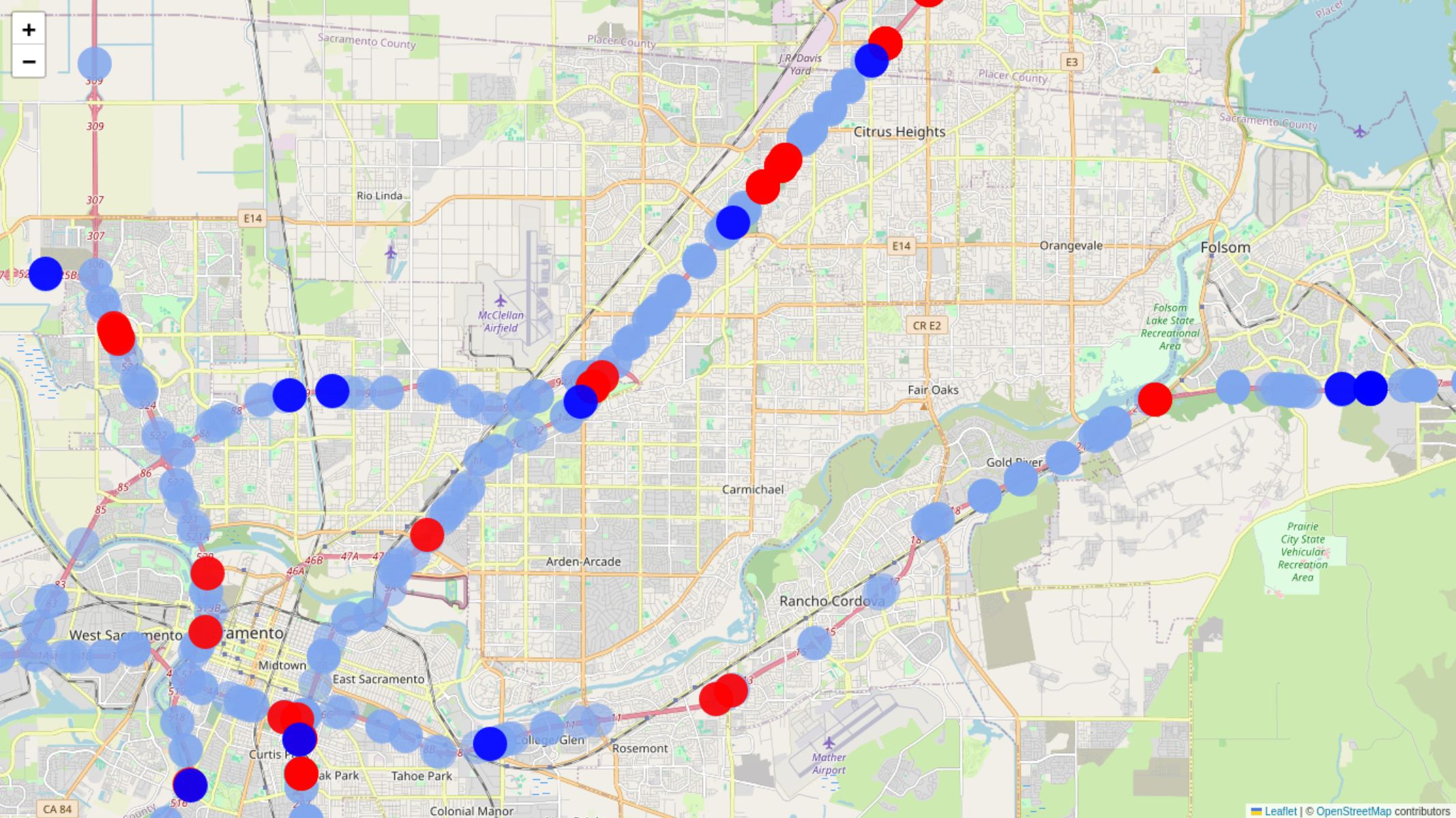}
            \caption{May}
        \end{subfigure}
        \begin{subfigure}[t]{0.13\linewidth}
        \centering
            \includegraphics[width=0.95\linewidth]{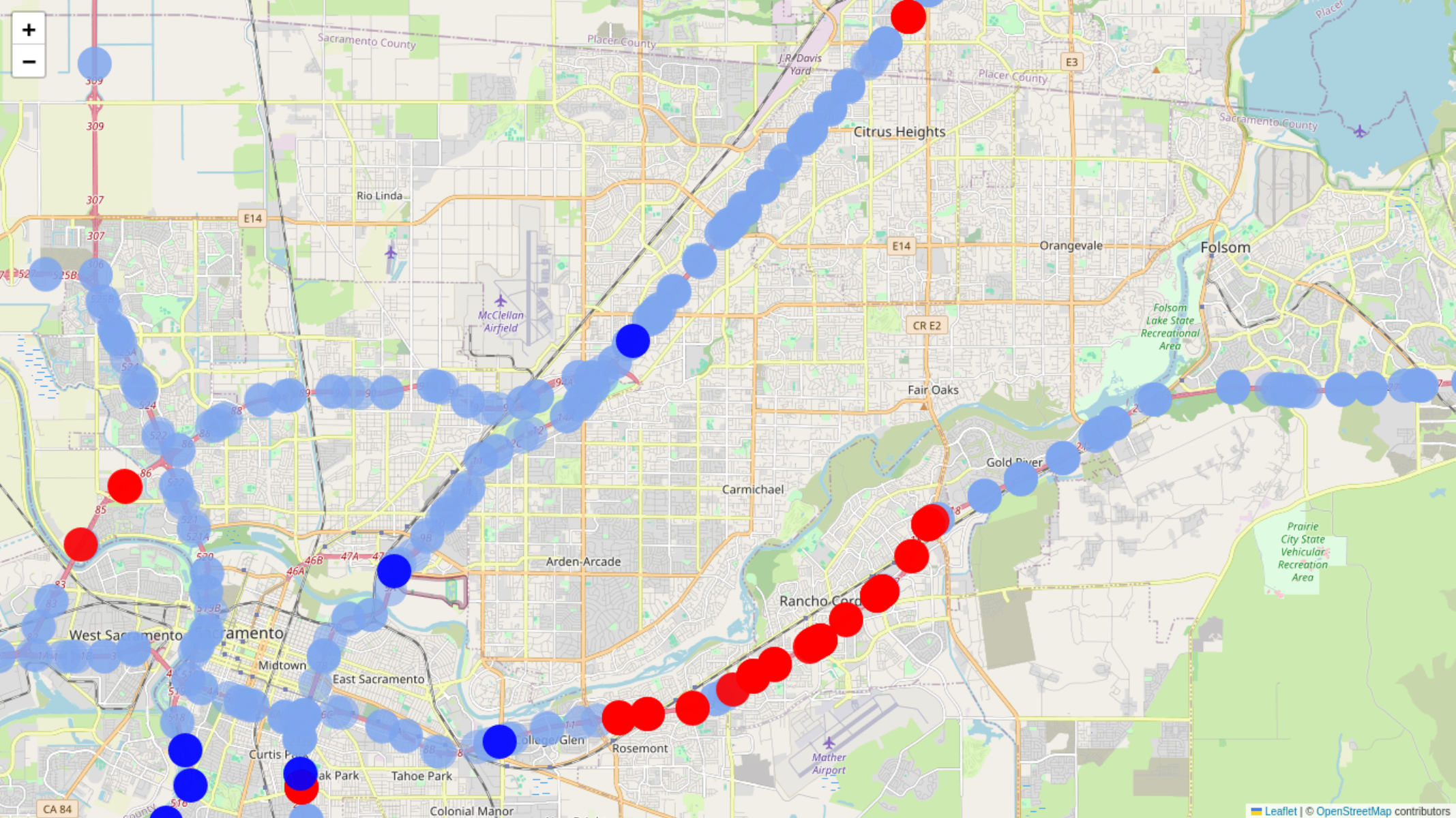}
            \caption{June}
        \end{subfigure}
        \begin{subfigure}[t]{0.13\linewidth}
        \centering
            \includegraphics[width=0.95\linewidth]{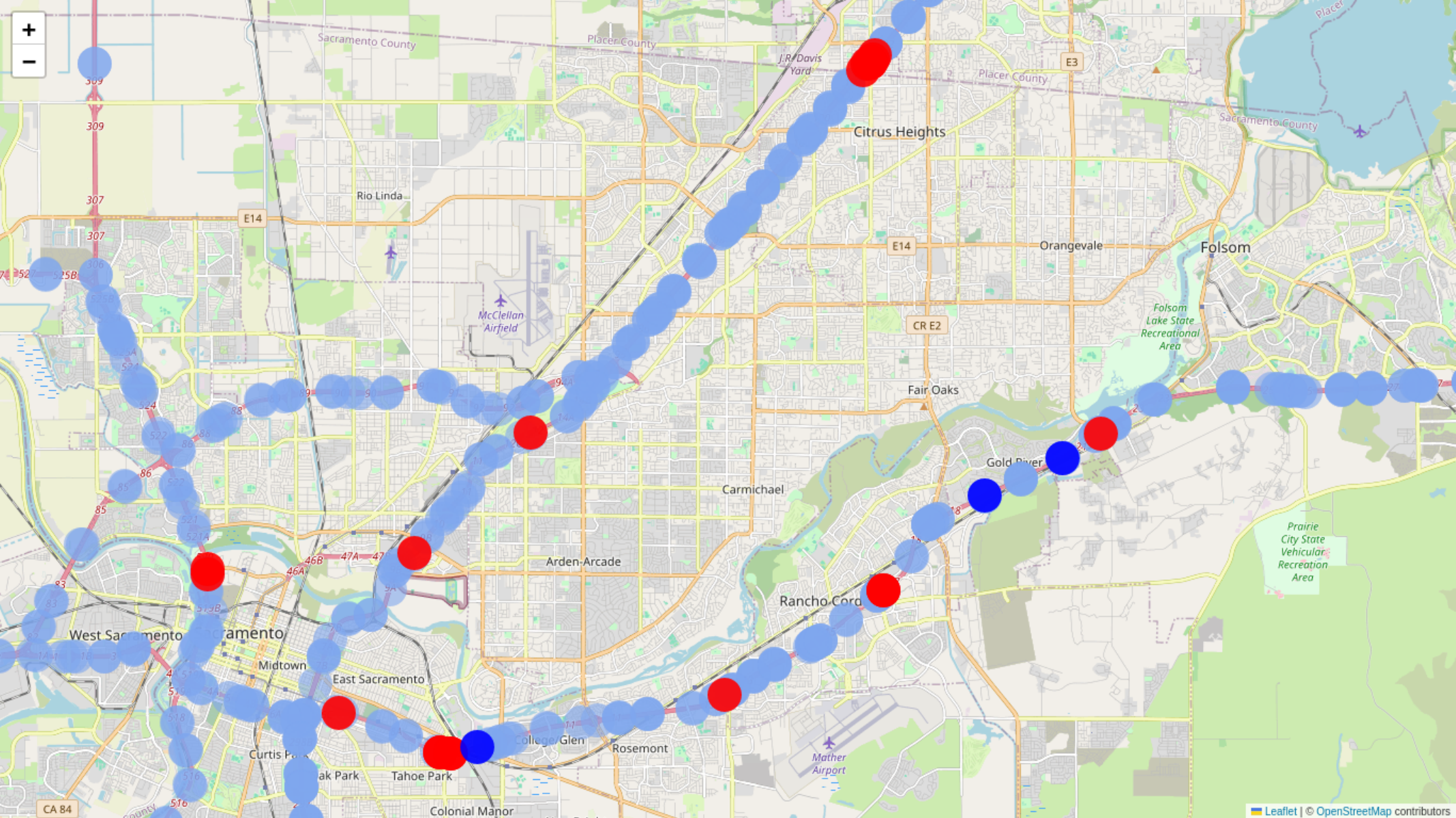}
            \caption{July}
        \end{subfigure}
        \begin{subfigure}[t]{0.13\linewidth}
        \centering
            \includegraphics[width=0.95\linewidth]{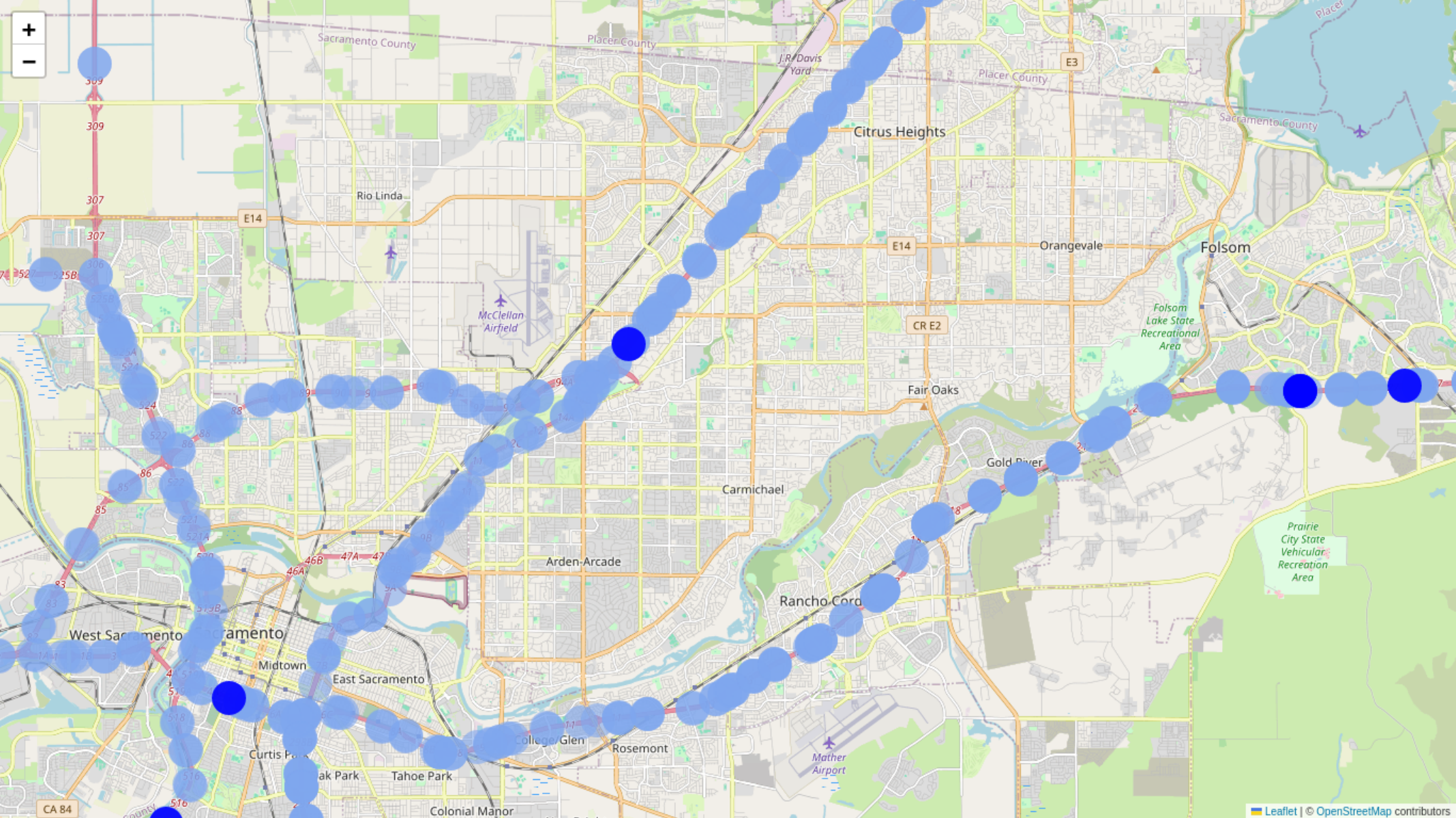}
            \caption{August}
        \end{subfigure}
        \begin{subfigure}[t]{0.13\linewidth}
        \centering
            \includegraphics[width=0.95\linewidth]{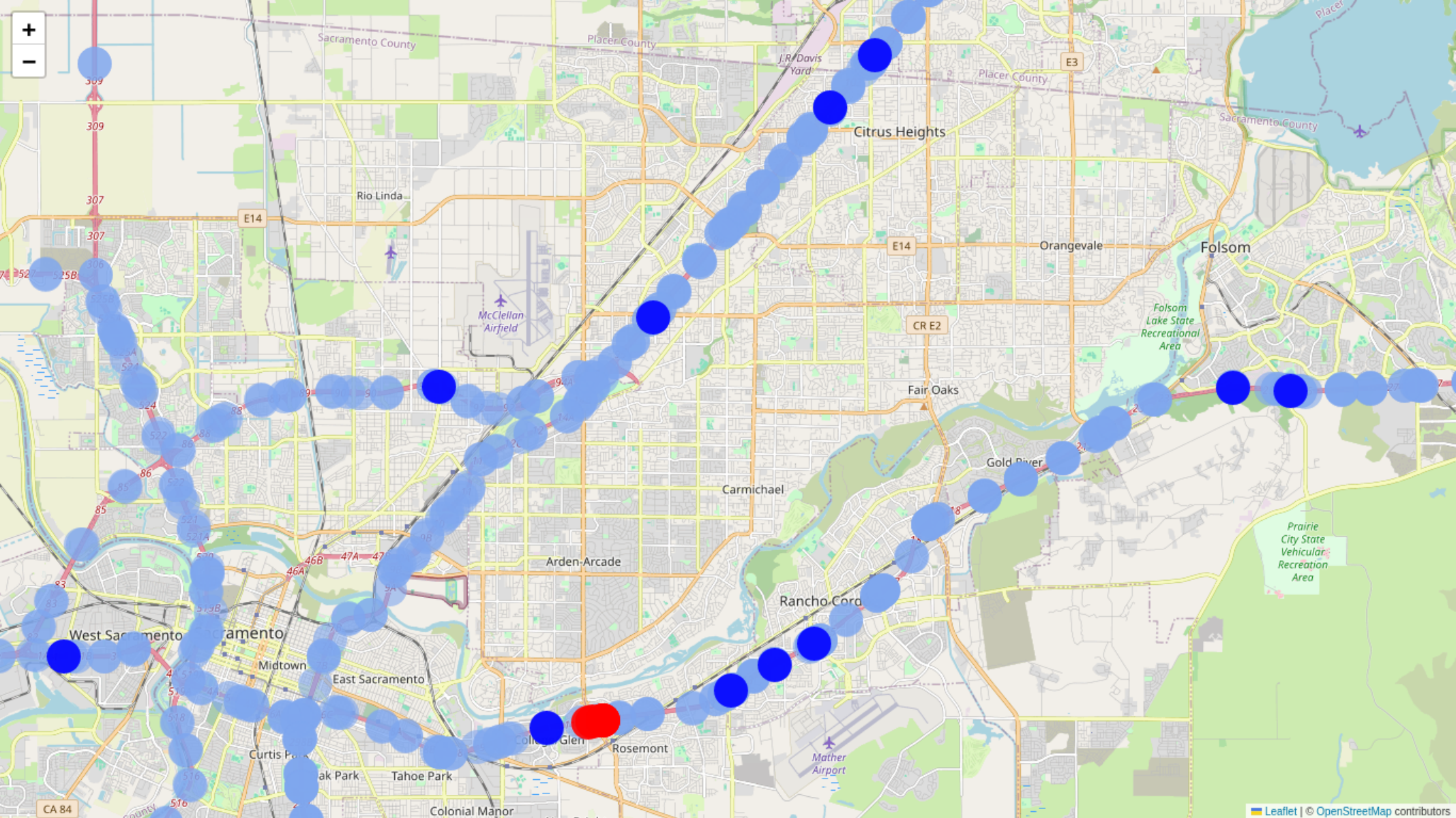}
            \caption{September}
        \end{subfigure}
        \begin{subfigure}[t]{0.13\linewidth}
        \centering
            \includegraphics[width=0.95\linewidth]{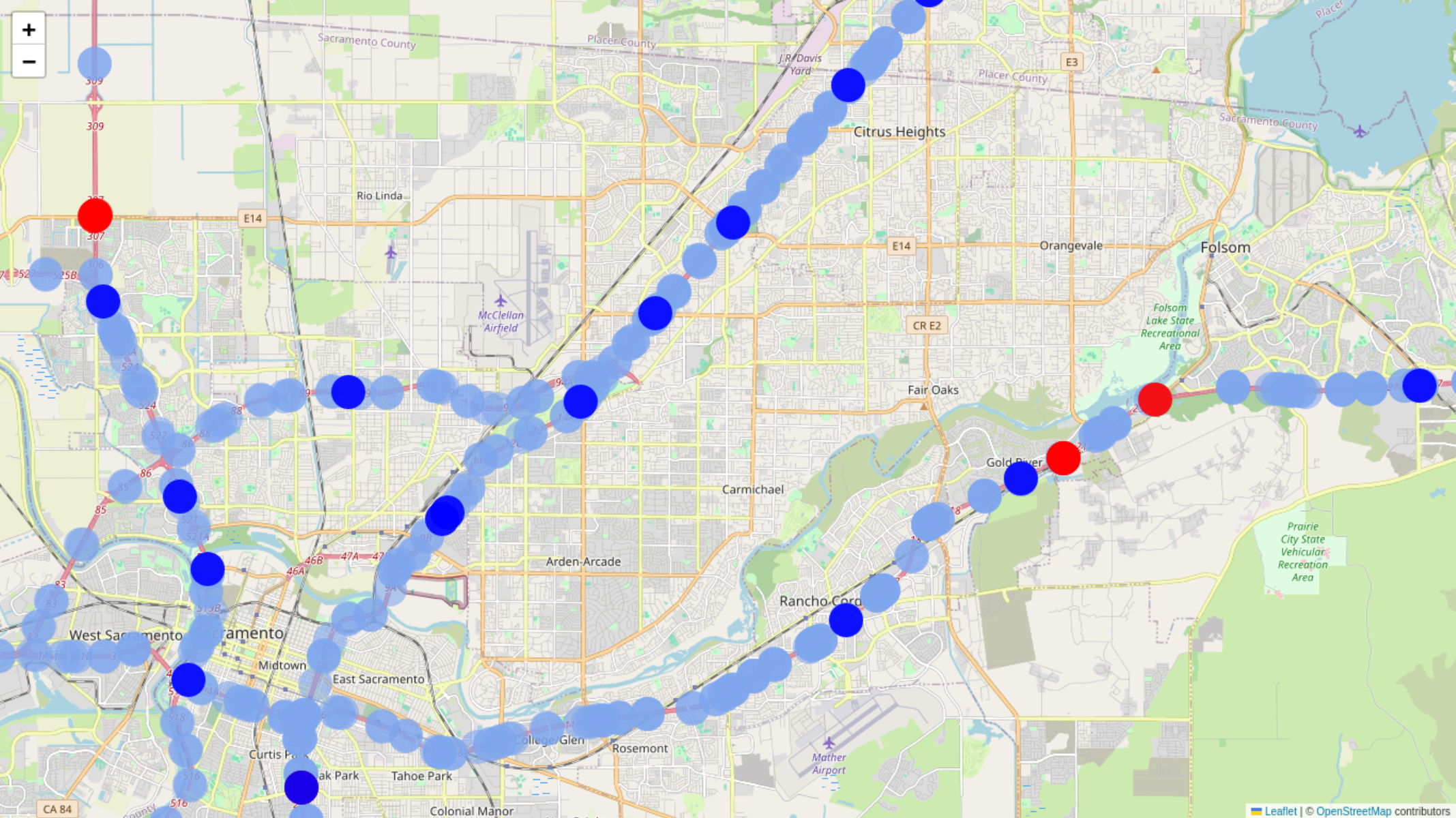}
            \caption{October}
        \end{subfigure}
        \caption{PEMS03-Evolve RN visualizations. Red nodes denote added nodes and blue nodes denote removed nodes, respectively.}
    \label{fig:pems03_dataset_visualization}
\end{figure*}

\begin{figure*}[t]
    \centering
        \begin{subfigure}[t]{0.13\linewidth}
        \centering
            \includegraphics[width=0.95\linewidth]{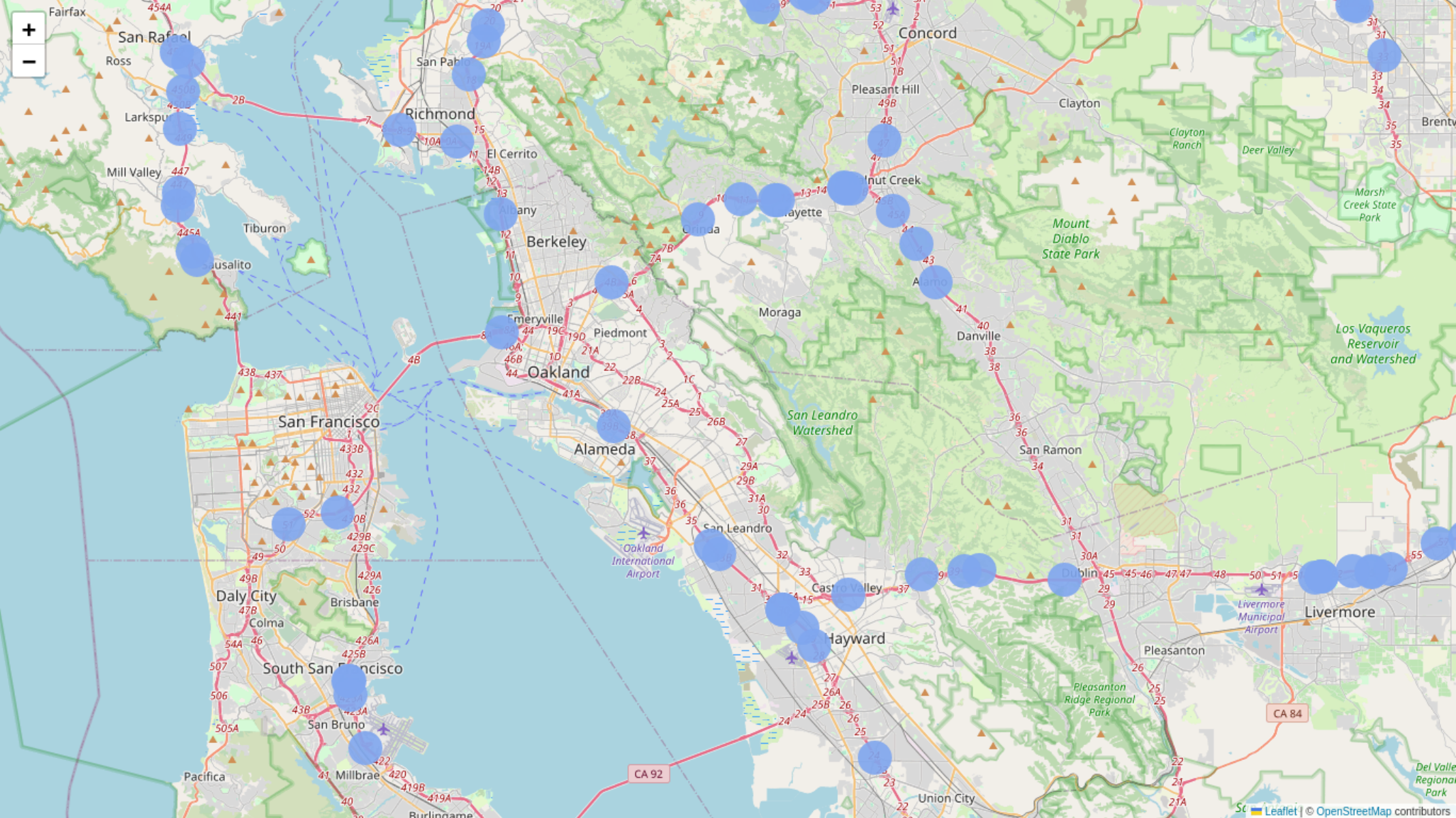}
            \caption{April}
        \end{subfigure}
        \begin{subfigure}[t]{0.13\linewidth}
        \centering
            \includegraphics[width=0.95\linewidth]{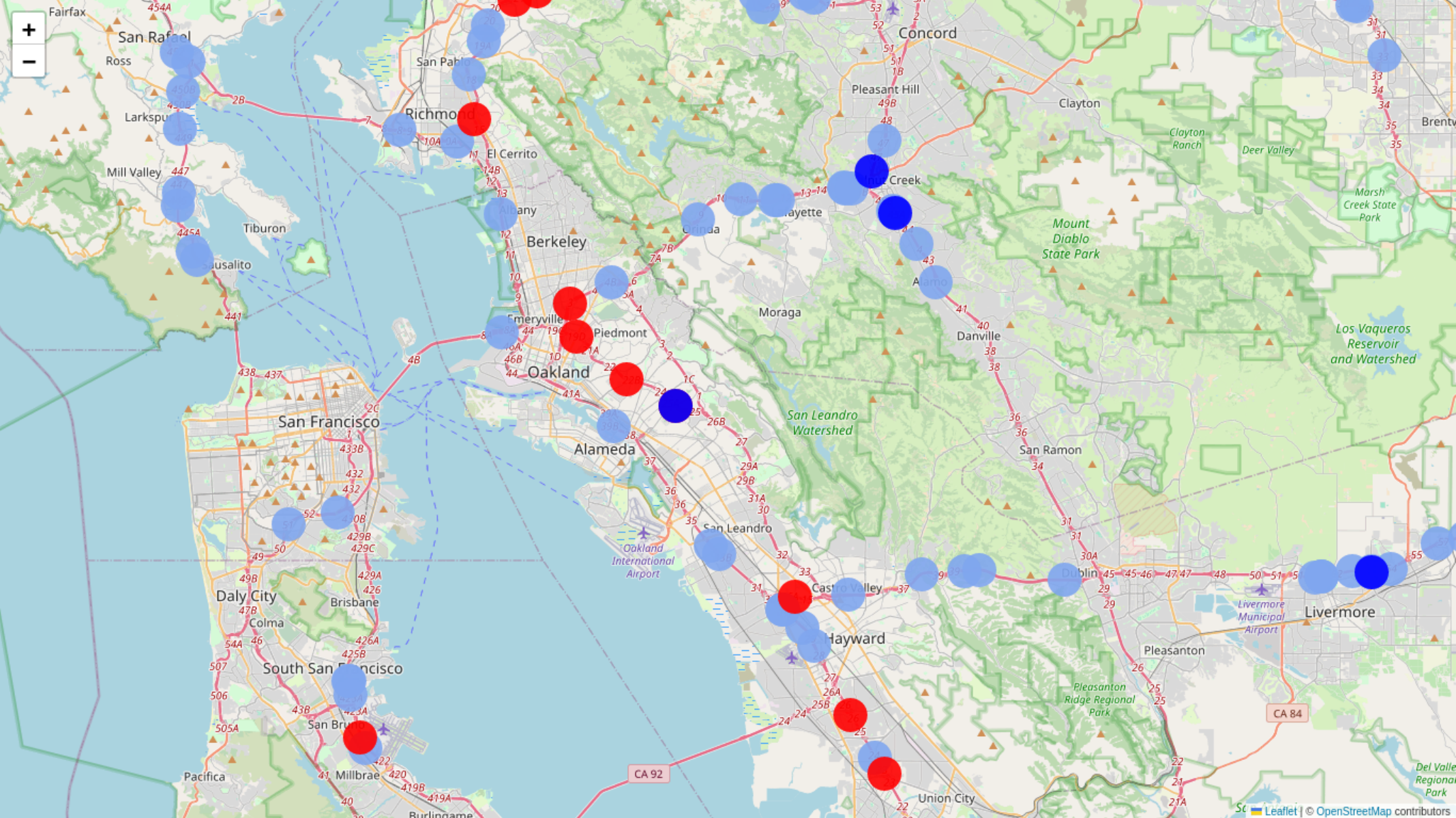}
            \caption{May}
        \end{subfigure}
        \begin{subfigure}[t]{0.13\linewidth}
        \centering
            \includegraphics[width=0.95\linewidth]{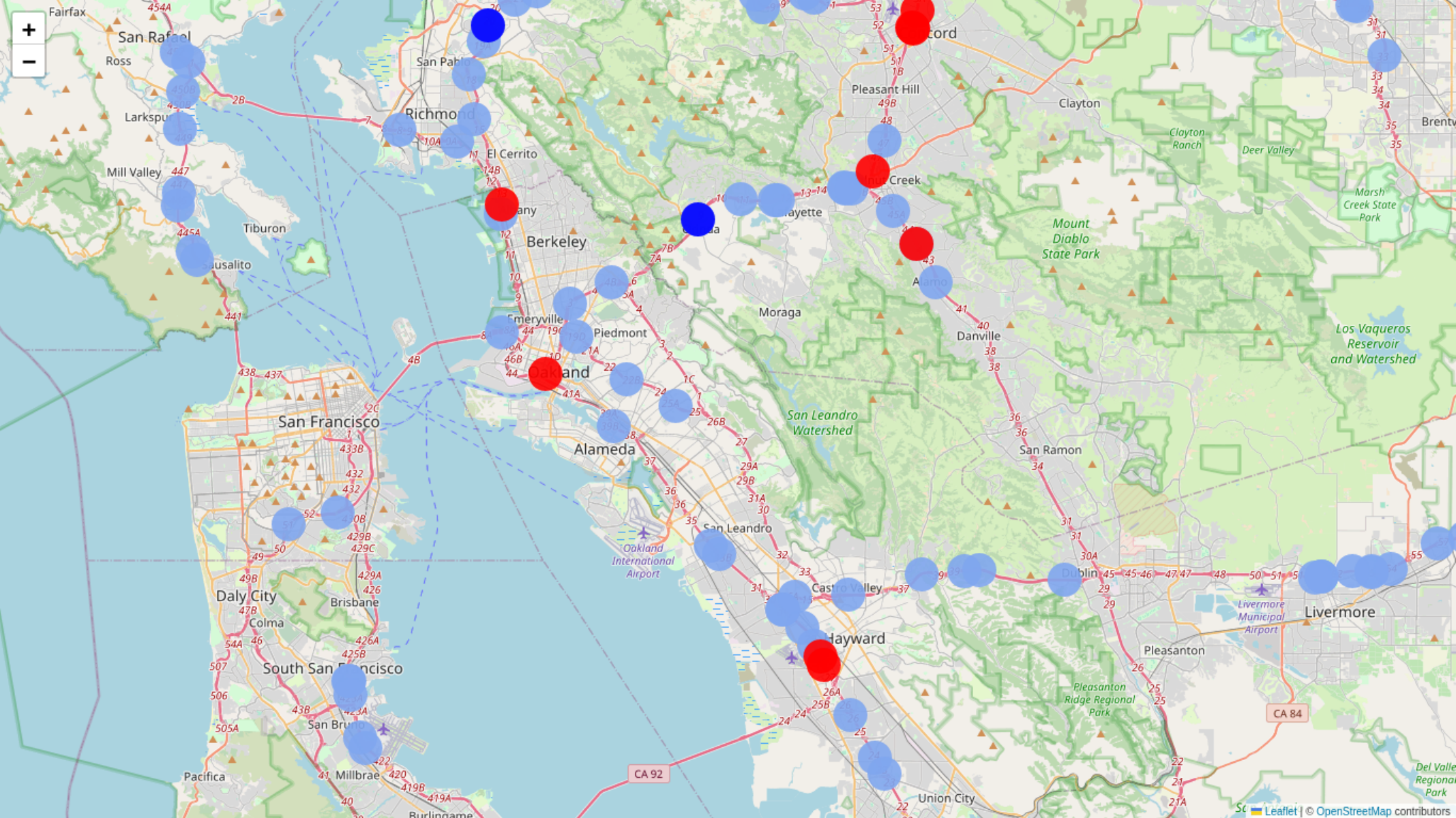}
            \caption{June}
        \end{subfigure}
        \begin{subfigure}[t]{0.13\linewidth}
        \centering
            \includegraphics[width=0.95\linewidth]{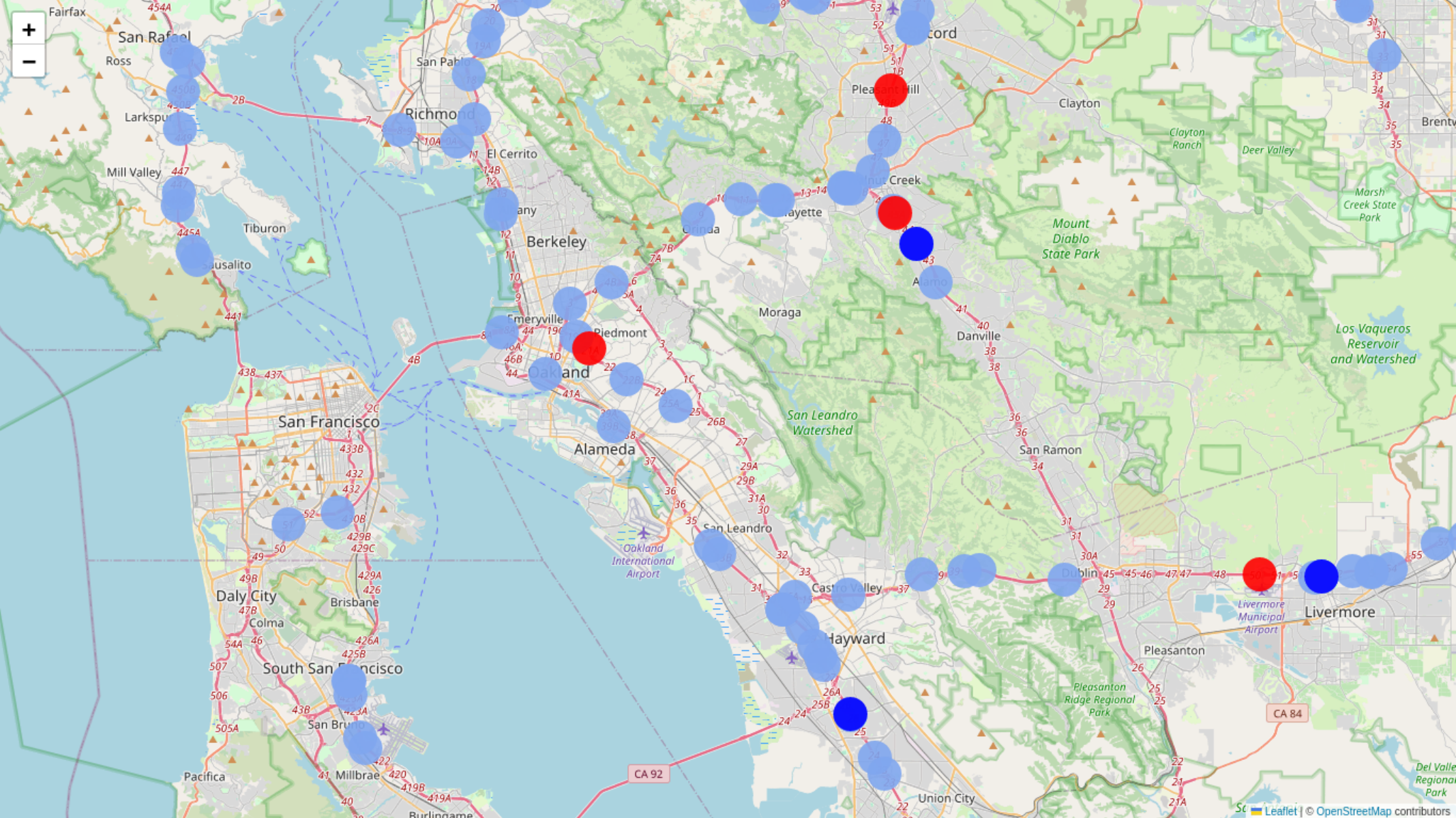}
            \caption{July}
        \end{subfigure}
        \begin{subfigure}[t]{0.13\linewidth}
        \centering
            \includegraphics[width=0.95\linewidth]{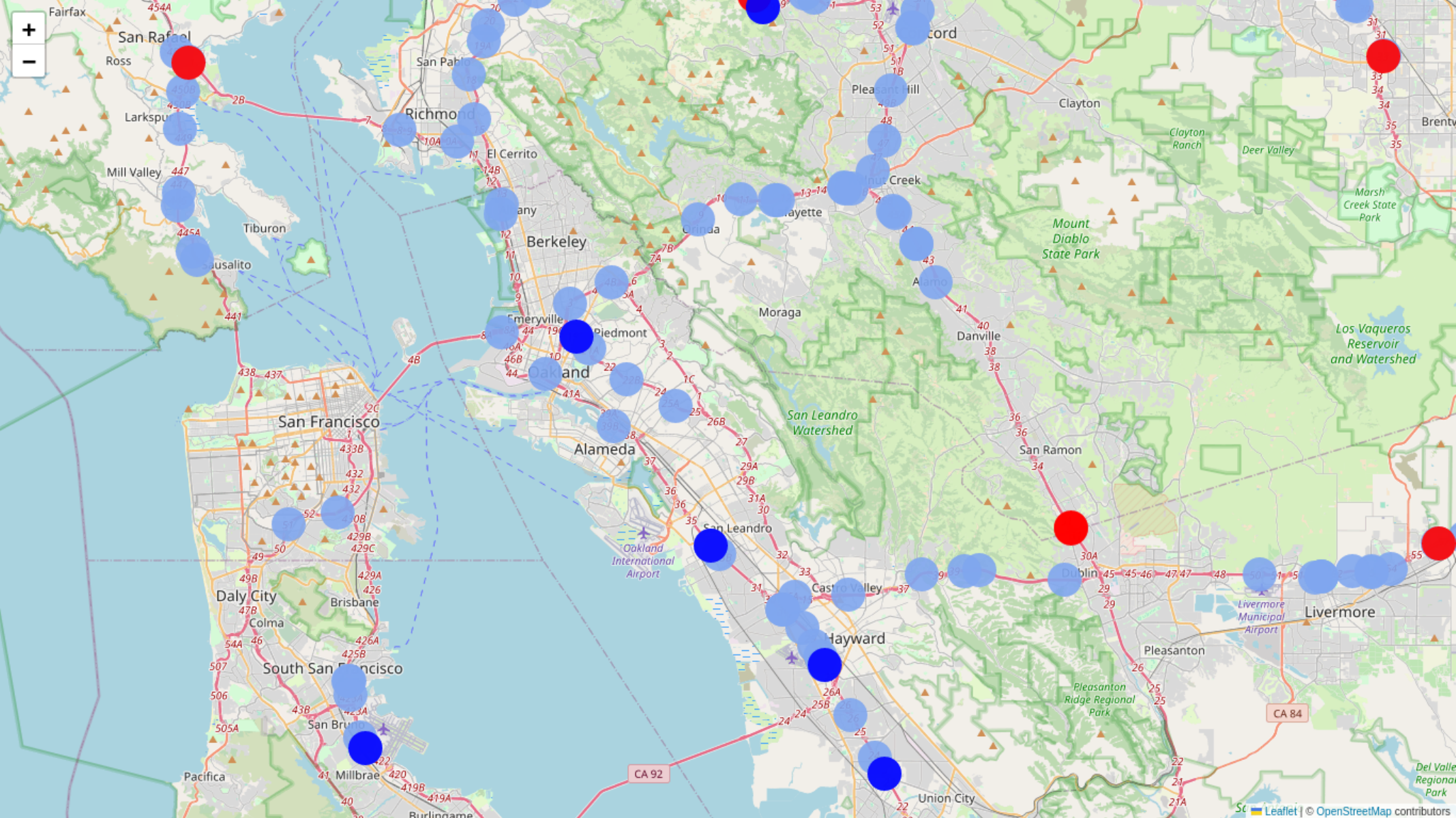}
            \caption{August}
        \end{subfigure}
        \begin{subfigure}[t]{0.13\linewidth}
        \centering
            \includegraphics[width=0.95\linewidth]{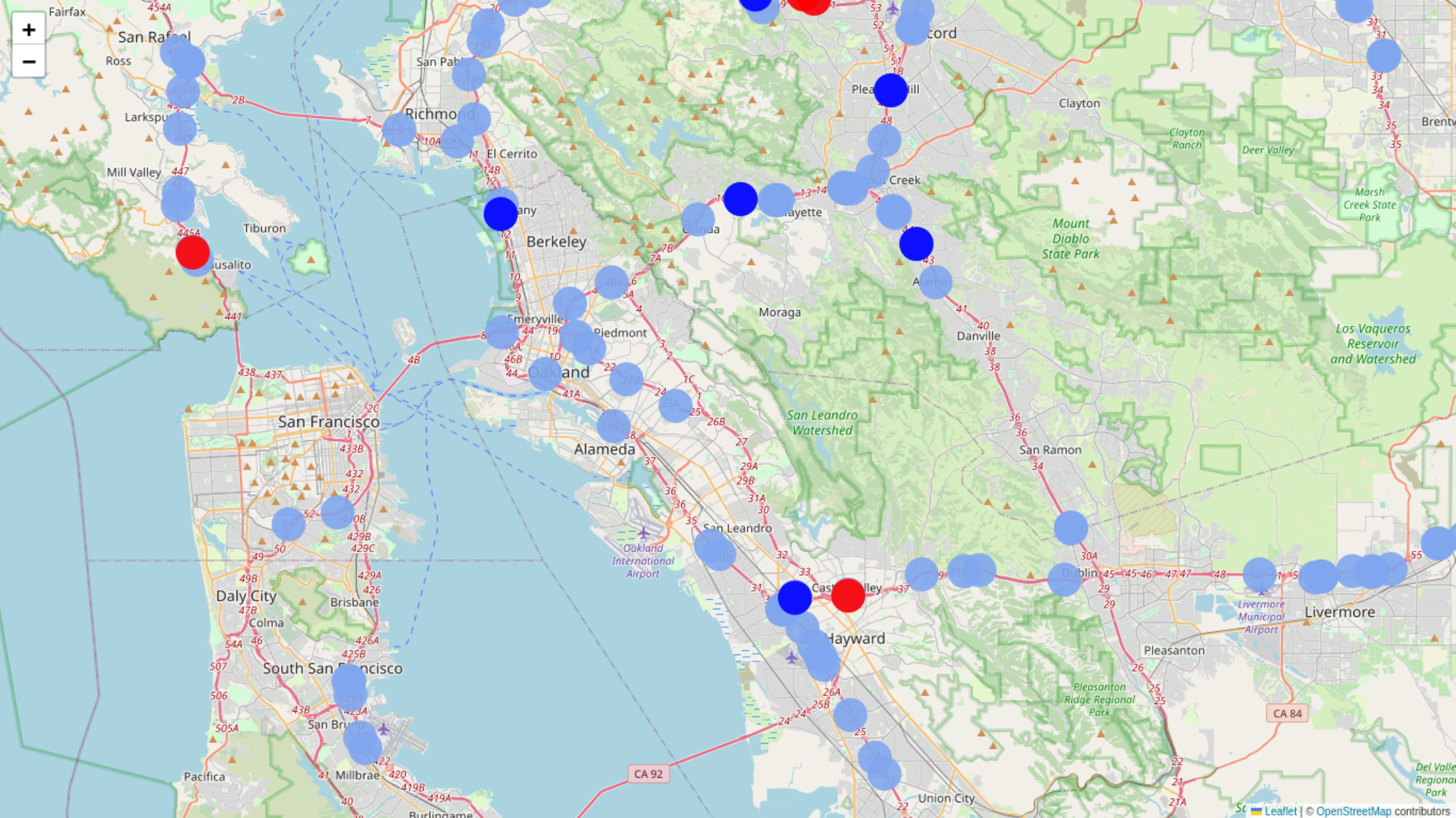}
            \caption{September}
        \end{subfigure}
        \begin{subfigure}[t]{0.13\linewidth}
        \centering
            \includegraphics[width=0.95\linewidth]{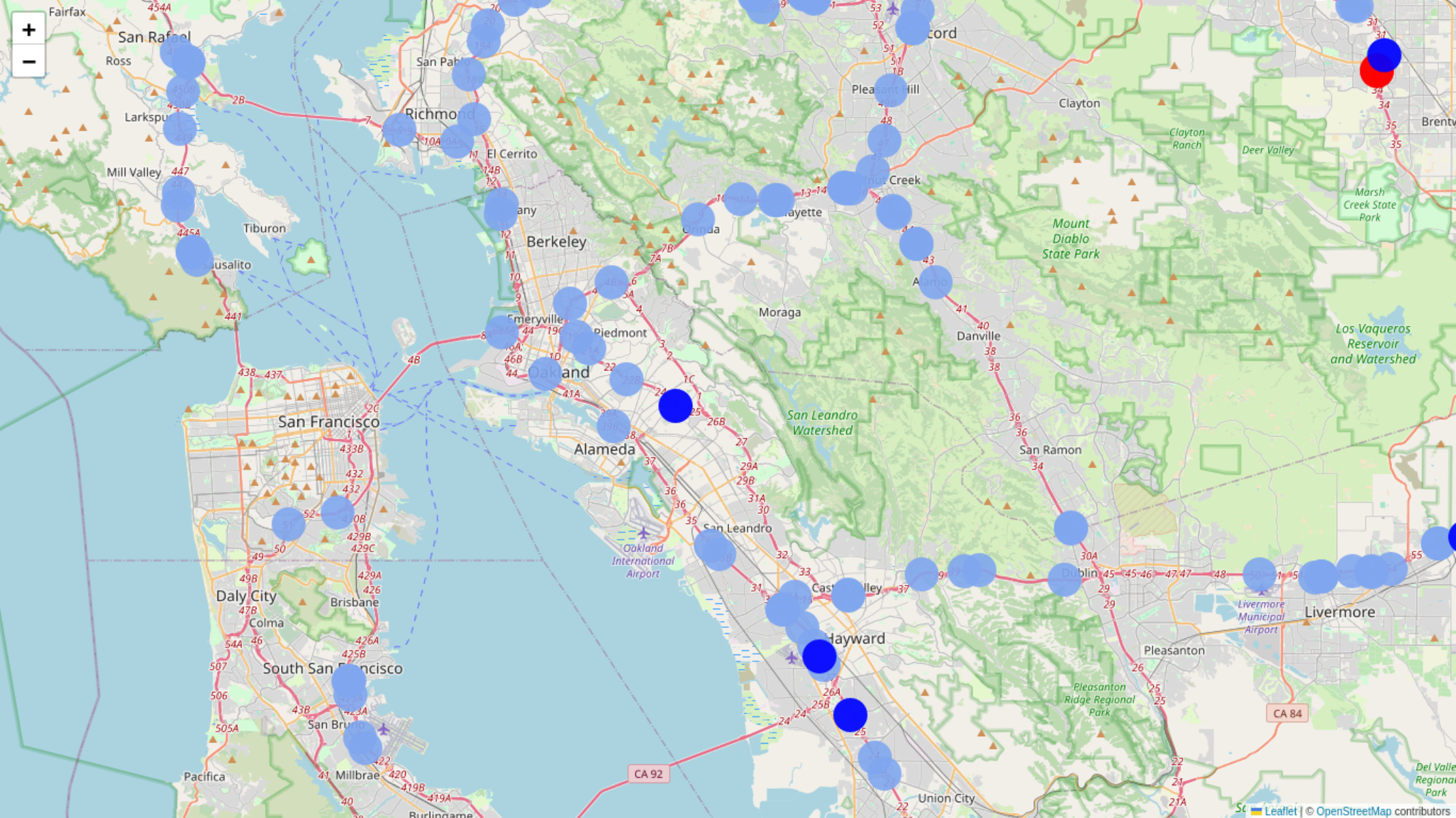}
            \caption{October}
        \end{subfigure}
        \caption{PEMS04-Evolve RN visualizations. Red nodes denote added nodes and blue nodes denote removed nodes, respectively.}
    \label{fig:pems04_dataset_visualization}
\end{figure*}

\begin{table}[t]
    \fontsize{7.5pt}{7.5pt}\selectfont
    \centering
    \caption{Details of \textbf{PEMS03-Evolve}. $+ \lvert \: \cdot \: \rvert$ and $- \lvert \: \cdot \: \rvert$ denote the number of added elements and removed elements, respectively.}
    \label{tab:datasets_PEMS03-Evolve}
    \begin{tabular}{c|p{0.5cm}p{0.5cm}p{0.5cm}p{0.5cm}p{0.5cm}p{0.5cm}p{0.5cm}}
        \toprule
        \textbf{Month}                    & \textbf{Apr} & \textbf{May} & \textbf{Jun} & \textbf{Jul} & \textbf{Aug} & \textbf{Sep} & \textbf{Oct} \\ \hline
        $\lvert V \rvert$        & 655  & 715  & 768  & 822  & 834  & 850  & 871  \\ 
        $\lvert E \rvert$        & 1,577 & 1,929 & 2,316 & 2,536 & 2,594 & 2,691 & 2,788 \\
        $+ \lvert \Delta V \rvert$ & N/A  & 60   & 53   & 54   & 12   & 16   & 21  \\ 
        $ + \lvert \Delta E \rvert$ & N/A  & 352  & 387  & 220  & 58   & 97   & 97 \\
        $- \lvert \Delta V \rvert$ & N/A  & 26   & 14   & 7   & 16   & 25   & 25  \\ 
        $- \lvert \Delta E \rvert$ & N/A  & 143  & 78  & 34  & 81   & 143   & 178 \\
        No. observations         & 8,856 & 8,856 & 8,856 & 8,856 & 8,856 & 8,856 & 8,856 \\
        \bottomrule
    \end{tabular}
\end{table}

\begin{table}[t]
    \fontsize{7.5pt}{7.5pt}\selectfont
    \centering
    \caption{Details of \textbf{PEMS04-Evolve}. $+ \lvert \: \cdot \: \rvert$ and $- \lvert \: \cdot \: \rvert$ denote the number of added elements and removed elements, respectively.}
    \label{tab:datasets_PEMS04-Evolve}
    \begin{tabular}{c|p{0.5cm}p{0.5cm}p{0.5cm}p{0.5cm}p{0.5cm}p{0.5cm}p{0.5cm}}
        \toprule
                \toprule
        \textbf{Month}                    & \textbf{Apr} & \textbf{May} & \textbf{Jun} & \textbf{Jul} & \textbf{Aug} & \textbf{Sep} & \textbf{Oct} \\ \hline
        $\lvert V \rvert$        &180  &198  &213  &225  &235  &243  &248  \\ 
        $\lvert E \rvert$        &308  &400  &469  &535  &583  &614  &623  \\
        $ + \lvert \Delta  V \rvert$ & N/A  & 18  & 15  & 12  & 10  & 8   & 5  \\
        $ + \lvert \Delta  E \rvert$ & N/A  & 92  & 69  & 66  & 48  & 31  & 9  \\
        $ - \lvert \Delta  V \rvert$ & N/A  & 10  & 9   & 4   & 15  & 14  & 12  \\ 
        $ - \lvert \Delta  E \rvert$ & N/A  & 36  & 30  & 24  & 72  & 66  & 60 \\
        No. observations         &8,905  &8,905  &8,905  &8,905  &8,905  &8,905  &8,905  \\
        \bottomrule
    \end{tabular}
\end{table}

\subsubsection{Forecasting Setting}
We design two scenarios to evaluate the accuracy and the runtime of the proposed framework.
In the first scenario, we evaluate \texttt{CAST}. \texttt{CAST} is first trained by using the data of the first period. 
Then, \texttt{CAST} is reinitialized and fully trained every period by using the data of all nodes for that period. 
Also, all the other baselines are evaluated in this setting.
We report the accuracy of the framework for the last period and the runtime for the whole training procedure.
The first scenario mimics the trivial solution of training a traffic forecasting model each time an RN evolves.

In the second scenario, we evaluate \texttt{TEAM}, which is first trained by using the data of the first period. 
Then, the framework is transferred and partially trained every following period by using only the data of newly updated nodes and the data in the buffer, i.e., employing the continual learning module.
We also report the accuracy of the framework for the last period and the runtime for the whole training procedure. Intuitively, the last period is the most difficult setting because the topology of the last period is the most different from that of the first period. By evaluating the accuracy of the model in the last period, we aim to evaluate the model in the most challenging setting. If the model performs well here, it can also perform well in intermediate periods.
We follow existing studies for the setting of forecasting horizon~\cite{DBLP:conf/ijcai/YuYZ18,DBLP:conf/iclr/LiYS018,DBLP:conf/aaai/ZhengFWQ20}. 
Given the previous $P = 12$ time steps (i.e., 1 hour), we aim to forecast the next $K = 12$ time steps (i.e., 1 hour).

\subsubsection{Baselines}

We compare our framework with the following baselines:
    (1) \texttt{HA}~\cite{HA_book}: a smoothing method, which forecasts the future values by averaging historical values; 
    (2) \texttt{VAR}~\cite{Ham1994}: an auto-regressive method, which assumes the data follow a predefined function;
    (3) \texttt{SVR}~\cite{DBLP:jour/eswa/Castro-NetoJJH09a}: a kernel-based method, which maps temporal data into a latent space and uses support vectors for forecasting;
    (4) \texttt{GRU}~\cite{DBLP:conf/corr/ChungGCB14}: a pure \texttt{RNN} model, which can capture long-term dependencies using gate mechanism;
    (5) \texttt{DCRNN}~\cite{DBLP:conf/iclr/LiYS018}: a sequence-to-sequence architecture, which uses \texttt{GCN}s to model spatial information and \texttt{RNN} to model temporal information; 
    (6) \texttt{STGCN}~\cite{DBLP:conf/ijcai/YuYZ18}: a sandwich architecture, which encloses \texttt{GCN}s with \texttt{1DCNN}s;
    (7) \texttt{GWN}~\cite{DBLP:conf/ijcai/WuPLJZ19}: a causal \texttt{CNN} based method, which uses \texttt{GCN}s to model spatial information and dilated causal \texttt{CNN}s to model temporal information;
    (8) \texttt{MSTGCN}~\cite{DBLP:conf/aaai/GuoLFSW19}: a state-of-the-art method, which incorporates multi-view mechanism into \texttt{GCN}s.   
    (9) \texttt{ASTGCN}~\cite{DBLP:conf/aaai/GuoLFSW19}: a state-of-the-art method, which uses attentions and \texttt{GCN}s for model temporal and spatial information, respectively;       
    (10) \texttt{STSGCN} \cite{DBLP:conf/aaai/SongLGW20}: an advanced approach that utilizes localized a spatial-temporal subgraph module to capture the spatial-temporal correlation simultaneously.
    (11) \texttt{EvolveGCN}~\cite{DBLP:conf/aaai/ParejaDCMSKKSL20}: a state-of-the-art dynamic graph embedding method that employs \texttt{RNN}s and \texttt{GCN}s to capture change between graph snapshots; 
    (12) \texttt{DyRep}~\cite{DBLP:conf/iclr/TrivediFBZ19}: a state-of-the-art dynamic graph embedding method that models the local and global topological evolution;
    (13) \texttt{GMAN}~\cite{DBLP:conf/aaai/ZhengFWQ20}: a traffic forecasting framework using multiple graph attention networks to model spatio-temporal dynamics;
    (14) \texttt{EnhanceNet}~\cite{DBLP:conf/icde/CirsteaKG0P21}: a plugin that integrates to \texttt{RNN}s and \texttt{GCN}s to capture correlation among different entities;
    (15) \texttt{ST-WA}~\cite{DBLP:conf/icde/CirsteaYGKP22}: a framework that considers location-specific and time-varying model parameters to capture complex spatio-temporal dynamics;
    (16) \texttt{D2STGNN}~\cite{DBLP:journals/pvldb/ShaoZWWXCJ22}: a framework that captures the diffusion and inherent traffic information separately;
    (17) \texttt{PDFormer}~\cite{DBLP:conf/aaai/JiangHZW23}: a model that captures both short-range and long-range dynamic spatial dependencies;
    (18) \texttt{TrafficStream}~\cite{DBLP:conf/ijcai/ChenWX21}: a method to efficiently support traffic forecasting on expandable RNs. 
    To adapt \texttt{EvolveGCN} and \texttt{DyRep} to traffic forecasting, we add a \texttt{1DCNN} layer on top of both methods to produce the forecasting results.
    Most of the baselines are applicable only to the first scenario whereas \texttt{TrafficStream} is also applicable to the second scenario. 

\subsubsection{Evaluation Metrics}
We use three metrics including Mean Absolute Error (\textit{MAE}), Root Mean Squared Error (\textit{RMSE}), and Mean Absolute Percentage Error (\textit{MAPE})~\cite{DBLP:conf/iclr/LiYS018} between forecasted time series and ground truth to measure the accuracy. We also report the average runtime for each epoch and the total runtime. 
The accuracy metrics are defined as follows.

{
    \begin{align}
        \textit{MAE} &= \frac{1}{K} \sum_{t=1}^{K} \left| \mathbf{X}^{\pi}_{T+t} - \hat{\mathbf{X}}^{\pi}_{T+t} \right|
    \end{align}
}
{
    \begin{align}
        \textit{RMSE} &= \sqrt{\frac{1}{K} \sum_{t=1}^{K} \left (\mathbf{X}^{\pi}_{T+t} - \hat{\mathbf{X}}^{\pi}_{T+t}\right)^{2}} 
    \end{align}
}
{
    \begin{align}
        \textit{MAPE} &= \frac{1}{K} \sum_{t=1}^{K} \left| \frac{\hat{\textbf{X}}^{\pi}_{T+t} - \textbf{X}^{\pi}_{T+t}}{\textbf{X}_{T+t}} \right| \cdot 100\%
    \end{align}
}

\begin{table*}[t]
	\centering
	\fontsize{7.5pt}{7.5pt}\selectfont
        \setlength{\tabcolsep}{3.0pt}
	\caption{Overall accuracy and runtime, \textbf{PEMS03-Evolve}.}
	\label{tab:main_result_1}
	\begin{tabular}{l|l|lll|lll|lll|ll}
		\toprule \multirow{2}{*}{\textbf{Scenario}} & \multirow{2}{*}{\textbf{Model}}          & \multicolumn{3}{c|}{\textbf{15 mins}} & \multicolumn{3}{c|}{\textbf{30 mins}} & \multicolumn{3}{c|}{\textbf{60 mins}} & \multicolumn{2}{c}{\textbf{Runtime (seconds)}} \\
		\cline{3-13}                                &                                          & \textit{MAE}                          & \textit{RMSE}                         & \textit{MAPE}                         & \textit{MAE}                                  & \textit{RMSE}           & \textit{MAPE}           & \textit{MAE}            & \textit{RMSE}           & \textit{MAPE}           & Total                        & Average                  \\
		\hline
		\multirow{13}{*}{1st}              & \texttt{HA}            & 17.25   & 31.55 & 25.67 & 17.14   & 31.34 & 25.67 & 17.19   & 31.49 & 25.43 & 89.91               & -       \\
         & \texttt{VAR}           & 17.65   & 28.41 & 23.48 & 18.49   & 29.28 & 23.63 & 20.93   & 33.03 & 26.00 & 502.81              & -       \\
         & \texttt{SVR}           & 15.04   & 24.37 & 17.98 & 16.21   & 26.10 & 21.23 & 19.51   & 32.36 & 25.68 & N/A                 & N/A     \\
		\cline{2-13}  
         & \texttt{GRU}           & 13.49   & 23.04 & 18.83 & 14.58   & 25.07 & 20.58 & 17.34   & 29.02 & 24.92 & \textbf{6316.75}             & \textbf{16.21}   \\
         & \texttt{DCRNN}         & \textbf{11.97}   & \textbf{18.57} & 19.82 & 14.77   & 34.09 & 22.93 & 17.28   & 29.50 & 24.13 & 231162.3            & 426.45  \\
         & \texttt{STGCN}         & 12.41   & 20.34 & \textbf{16.89} & 14.43   & 23.63 & 20.69 & 17.58   & 28.63 & 25.71 & 22848.19            & 54.60   \\
         & \texttt{STSGCN}        & 13.15   & 21.09 & 17.35 & 13.48   & 22.05 & \textbf{17.84} & 14.62   & 24.02 & \uline{19.54} & 63687.41            & 163.05  \\
         & \texttt{GWN}           & 13.10   & 21.67 & \uline{17.17} & 14.02   & 23.45 & 18.86 & 16.14   & 26.50 & 21.14 & 42294.72            & 96.77   \\
         & \texttt{MSTGCN}        & 13.67   & 21.73 & 19.73 & 14.55   & 23.65 & 20.59 & 16.84   & 27.34 & 24.98 & 29075.06            & 61.87   \\
         & \texttt{ASTGCN}        & 13.28   & 21.66 & 20.64 & 14.67   & 23.60 & 21.73 & 16.78   & 27.29 & 24.82 & 37941.11            & 63.66   \\
         & \texttt{EvolveGCN}     & 14.41   & 23.56 & 20.01 & 15.73   & 25.94 & 21.98 & 18.48   & 30.98 & 25.83 & 34936.41            & 48.51   \\
         & \texttt{DyRep}         & 13.65   & 22.26 & 21.48 & 15.11   & 23.61 & 21.79 & 17.22   & 27.21 & 24.89 & 39486.63            & 110.59  \\
         & \texttt{GMAN}          & 18.13   & 29.40 & 24.42 & 20.73   & 30.82 & 24.79 & 23.16   & 31.99 & 26.28 & 158859.4            & 229.33  \\
         & \texttt{EnhanceNet}    & 12.47   & 21.54 & 18.98 & 13.42   & 22.19 & 20.60 & 15.04   & 24.72 & 22.33 & 67431.15            & 94.45   \\
         & \texttt{ST-WA}         & 12.57   & 20.89 & 20.05 & 13.70   & 23.21 & 23.82 & 14.45   & 23.85 & 24.75 & 95898.2             & 248.31  \\
         & \texttt{D2STGNN}       & 14.47   & 24.06 & 19.52 & 16.23   & 26.79 & 21.89 & 17.24   & 29.72 & 27.02 & 85794.14            & 205.08  \\
         & \texttt{PDFormer}      & 12.49   & 20.43 & 17.47 & \uline{13.26}   & \uline{21.90} & 18.75 & \uline{14.28}   & \uline{23.22} & 19.83 & 331641.3            & 736.85  \\
         & \texttt{CAST} (\textbf{ours})   & \uline{12.16}   & \uline{20.28} & 17.85 & \textbf{13.09}   & \textbf{21.63} & \uline{18.63} & \textbf{14.17}   & \textbf{22.91} & \textbf{19.14} & 27341.47            & 61.08   \\
		\hline
         2nd   & \texttt{TrafficStream} & 13.74   & 22.86 & 21.44 & 15.21   & 25.00 & 20.92 & 17.72   & 29.40 & 22.41 & 6974.98             & 36.12   \\
         & \texttt{TEAM} (\textbf{ours})   & 12.82   & 21.37 & 17.98 & 13.57   & 22.89 & 18.95 & 15.16   & 25.69 & 21.88 & \uline{6574.24}             & \uline{33.55}  \\
		\bottomrule
	\end{tabular}
\end{table*}

\begin{table*}[t]
	\centering
	\fontsize{7.5pt}{7.5pt}\selectfont
        \setlength{\tabcolsep}{3.0pt}
	\caption{Overall accuracy and runtime, \textbf{PEMS04-Evolve}.}
	\label{tab:main_result_2}
	\begin{tabular}{l|l|lll|lll|lll|ll}
		\toprule \multirow{2}{*}{\textbf{Scenario}} & \multirow{2}{*}{\textbf{Model}}          & \multicolumn{3}{c|}{\textbf{15 mins}} & \multicolumn{3}{c|}{\textbf{30 mins}} & \multicolumn{3}{c|}{\textbf{60 mins}} & \multicolumn{2}{c}{\textbf{Runtime (seconds)}} \\
		\cline{3-13}                                &                                          & \textit{MAE}                          & \textit{RMSE}                         & \textit{MAPE}                         & \textit{MAE}                                  & \textit{RMSE}          & \textit{MAPE}          & \textit{MAE}           & \textit{RMSE}          & \textit{MAPE}          & Total                        & Average                   \\
		\hline
		\multirow{13}{*}{1st}              & \texttt{HA}            & 2.21    & 4.75 & 4.79 & 2.21    & 4.75 & 4.79 & 2.21    & 4.75 & 4.79 & 57.45               & -       \\
         & \texttt{VAR}           & 1.96    & 3.79 & 3.82 & 2.62    & 4.58 & 4.98 & 2.90    & 5.07 & 5.66 & 206.72              & -       \\
         & \texttt{SVR}           & 1.79    & 3.45 & 3.48 & 2.08    & 4.35 & 4.48 & 2.57    & 5.67 & 5.45 & N/A                 & N/A     \\
		\cline{2-13}                                & \texttt{GRU}           & 4.33    & 6.47 & 9.51 & 5.60    & 6.61 & 9.89 & 6.45    & 7.36 & 9.49 & \textbf{1784.52}             & \textbf{8.61}    \\
         & \texttt{DCRNN}         & 1.51    & 2.92 & 2.87 & 1.78    & 3.76 & 3.41 & 2.42    & 4.47 & \uline{4.19} & 66014.6             & 119.30  \\
         & \texttt{STGCN}         & 2.79    & 5.06 & 5.89 & 2.86    & 5.17 & 4.96 & 3.15    & 5.54 & 6.75 & 6160.2              & 14.76   \\
         & \texttt{STSGCN}        & 2.86    & 5.87 & 6.55 & 2.99    & 6.16 & 6.95 & 3.25    & 6.69 & 7.68 & 29596.6             & 70.44   \\
         & \texttt{GWN}           & 1.37    & 2.90 & 2.74 & 1.71    & 3.89 & 3.56 & 2.07    & 4.58 & 4.43 & 12025.85            & 26.43   \\
         & \texttt{MSTGCN}        & 1.54    & 3.06 & 3.05 & 2.01    & 4.12 & 4.23 & 2.69    & 5.36 & 5.95 & 8436.25             & 18.64   \\
         & \texttt{ASTGCN}        & 1.64    & 3.49 & 3.53 & 1.98    & 4.20 & 4.27 & 2.37    & 4.98 & 5.45 & 10818.8             & 23.86   \\
         & \texttt{EvolveGCN}     & 1.50    & 2.88 & 3.17 & 1.75    & 3.56 & 3.75 & 2.12    & 4.52 & 4.67 & 9183.63             & 15.92   \\
         & \texttt{DyRep}         & 1.52    & 3.01 & 3.13 & 1.84    & 3.81 & 3.67 & 2.14    & 4.51 & 4.74 & 13239.52            & 32.23   \\
         & \texttt{GMAN}          & 2.28    & 4.40 & 4.80 & 2.96    & 5.83 & 6.52 & 2.97    & 5.85 & 6.60 & 47048.81            & 67.31   \\
         & \texttt{EnhanceNet}    & 1.40    & 2.87 & 2.80 & 1.75    & 3.90 & 3.75 & 2.12    & 4.74 & 4.70 & 16859.54            & 24.08   \\
         & \texttt{ST-WA}         & 1.61    & 3.45 & 3.44 & 1.83    & 4.11 & 4.02 & 2.11    & 4.66 & 4.63 & 26777.4             & 78.68   \\
         & \texttt{D2STGNN}       & 1.53    & 3.13 & 3.20 & 2.10    & 4.61 & 4.78 & 2.61    & 6.04 & 6.43 & 22540.51            & 64.43   \\
         & \texttt{PDFormer}      & \uline{1.30}    & \uline{2.62} & \uline{2.53} & \textbf{1.57}    & \uline{3.26} & 3.17 & \uline{1.97}    & 4.22 & 4.22 & 113412.2            & 233.33  \\
         & \texttt{CAST} (\textbf{ours})   & \textbf{1.28}    & \textbf{2.50} & \textbf{2.51} & \textbf{1.57}    & \textbf{3.20} & 3.19 & \textbf{1.96}    & \textbf{4.10} & \textbf{4.14} & 7962.96            & 17.82   \\
		\hline
		2nd & \texttt{TrafficStream} & 1.68    & 2.99 & 3.68 & 2.15    & 3.93 & 4.35 & 2.66    & 4.76 & 5.56 & 2250.41          & 10.24   \\
         & \texttt{TEAM} (\textbf{ours})   & 1.37    & 2.66 & 2.61 & \uline{1.60}    & 3.29 & 3.31 & 2.05    & \uline{4.21} & 4.25 & \uline{2007.21}             & \uline{9.11}    \\
		\bottomrule
	\end{tabular}
\end{table*}

\begin{table}[t]
    \centering
	\fontsize{7.5pt}{7.5pt}\selectfont
	\setlength{\tabcolsep}{3.0pt}
		\caption{Ablation study, \textbf{PEMS04-Evolve}.}
		\label{tab:ablation}
		\begin{tabular}{p{1.2cm}|p{0.52cm}p{0.56cm}p{0.58cm}|p{0.52cm}p{0.56cm}p{0.58cm}|p{0.52cm}p{0.56cm}p{0.58cm}}
			\toprule                                      & \multicolumn{3}{c|}{\textbf{15 mins}} & \multicolumn{3}{c|}{\textbf{30 mins}} & \multicolumn{3}{c}{\textbf{60 mins}} \\
			\cline{2-10} \multirow{-2}{*}{\textbf{Model}} & \textit{MAE}                          & \textit{RMSE}                         & \textit{MAPE}                       & \textit{MAE} & \textit{RMSE} & \textit{MAPE} & \textit{MAE} & \textit{RMSE} & \textit{MAPE} \\
			\hline
			\texttt{TEAM-CONV}                            & 1.49                                  & 2.74                                  & 2.92                                & 1.71         & 3.39          & 3.47          & 2.07         & 4.28          & 4.45          \\
			\texttt{TEAM-ATT}                             & 1.83                                  & 3.06                                  & 3.17                                & 1.93         & 3.74          & 3.98          & 2.28         & 4.92          & 5.25          \\
			\texttt{TEAM-SWAP}                            & 1.50                                  & 2.93                                  & 3.42                                & 1.79         & 3.64          & 4.16          & 2.17         & 4.59          & 5.29          \\
			\texttt{TEAM-BACK}                            & 1.53                                  & 2.98                                  & 3.40                                & 1.78         & 3.62          & 4.06          & 2.19         & 4.52          & 5.13          \\
			\texttt{TEAM-RES}                             & 1.56                                  & 2.73                                  & 2.92                                & 1.81         & 3.43          & 3.56          & 2.14         & 4.34          & 4.45          \\
			\texttt{TEAM-CONT}                            & 4.71                                  & 8.71                                  & 11.45                               & 4.74         & 8.79          & 11.57         & 4.94         & 8.87          & 11.64         \\\hline
			\texttt{TEAM}                                 & \textbf{1.37}    & \textbf{2.66} & \textbf{2.61} & \textbf{1.60}    & \textbf{3.29} & \textbf{3.31} & \textbf{2.05}    & \textbf{4.21} & \textbf{4.25} \\
			\bottomrule
		\end{tabular}%
\end{table}

\begin{table}[t]
    \centering
        \fontsize{7.5pt}{7.5pt}\selectfont
	\setlength{\tabcolsep}{3.0pt}
		\caption{Number of evolved nodes vs. runtime, \texttt{TEAM}, \textbf{PEMS04-Evolve}.}
		\label{tab:correlation} 
        \begin{tabular}{p{1.0cm}p{1.0cm}p{1.7cm}p{1.7cm}p{1.7cm}}\toprule \textbf{\#Evolved nodes} & \textbf{\#Training nodes} & \textbf{Prep. runtime} (\textit{seconds}) & \textbf{Total runtime} (\textit{seconds}) & \textbf{Avg. runtime} (\textit{seconds})\\ \midrule 13 & 24 & 1.7 & 134.8 & 6.74 \\ 16 & 36 & 1.7 & 144.2 & 7.21 \\ 19 & 45 & 1.8 & 157.4 & 7.87 \\ 22 & 51 & 1.9 & 159.6 & 7.98 \\ 25 & 63 & 1.9 & 163.4 & 8.17 \\ 28 & 70 & 2.1 & 166.8 & 8.34 \\ \bottomrule
        \end{tabular}
\end{table}

\begin{table}[t]
    \fontsize{7.5pt}{7.5pt}\selectfont
    \setlength{\tabcolsep}{3.0pt}
    \caption{Accuracy during immediate periods, \textbf{PEMS04-Evolve}.}
		\label{tab:immedate_period}
	\begin{tabular}{l|l|p{0.7cm}|p{0.4cm}p{0.4cm}p{0.4cm}p{0.4cm}p{0.4cm}p{0.4cm}p{0.4cm}}
\toprule
\textbf{Scenario} & \textbf{Model}      &   \textbf{Metric}   & \textbf{Apr} & \textbf{May} & \textbf{Jun} & \textbf{Jul} & \textbf{Aug} & \textbf{Sep} & \textbf{Oct} \\
\midrule
\multirow{6}{*}{1st} & \multirow{3}{*}{\texttt{EnhanceNet}}    & 
\textit{MAE}  & 1.33 & \uline{1.38} & \uline{1.48} & \uline{1.51} & \uline{1.64} & 1.68 & 1.76 \\
         &               & \textit{RMSE} & 2.72 & \uline{2.79} & \uline{2.83} & \uline{2.92} & \uline{2.95} & 3.18 & 3.84 \\
         &               & \textit{MAPE} & 2.58 & \uline{2.67} & \uline{2.61} & \uline{2.64} & \uline{2.88} & 3.26 & 3.75 \\
         \cline{2-10}
& \multirow{3}{*}{\texttt{CAST}}          & \textit{MAE}  & \textbf{1.31} & \textbf{1.34} & \textbf{1.46} & \textbf{1.47} & \textbf{1.56} & \textbf{1.59} & \textbf{1.60} \\
         &               & \textit{RMSE} & \textbf{2.68} & \textbf{2.66} & \textbf{2.81} & \textbf{3.01} & \textbf{3.09} & \textbf{3.27} & \textbf{3.27} \\
         &               & \textit{MAPE} & \textbf{2.49} & \textbf{2.51} & \textbf{2.97} & \textbf{2.91} & \textbf{3.22} & \textbf{3.29} & \textbf{3.28} \\\hline
\multirow{6}{*}{2nd} & \multirow{3}{*}{\texttt{TrafficStream}} & \textit{MAE}  & 1.56 & 1.66 & 1.67 & 1.75 & 1.76 & 1.92 & 2.16 \\
         &               & \textit{RMSE} & 3.24 & 3.31 & 3.39 & 3.44 & 3.74 & 3.75 & 3.89 \\
         &               & \textit{MAPE} & 3.19 & 3.27 & 3.46 & 3.51 & 3.63 & 3.94 & 4.53 \\
         \cline{2-10}
& \multirow{3}{*}{\texttt{TEAM}}          & \textit{MAE}  & \textbf{1.31} & 1.45 & 1.52 & 1.54 & 1.60 & \uline{1.67} & \uline{1.67} \\
         &               & \textit{RMSE} & \textbf{2.68} & 2.81 & 3.04 & 3.28 & 3.27 & \uline{3.36} & \uline{3.38} \\
         &               & \textit{MAPE} & \textbf{2.49} & 2.87 & 2.98 & 3.33 & 3.57 & \uline{3.36} & \uline{3.39} \\
\bottomrule
\end{tabular}
\end{table}

\subsubsection{Hyperparameter Settings}
We train the framework using the Adam optimizer with a learning rate of 0.001 and a batch size of 64.
The total number of epochs is set to 200, and we use early stopping with a patience of 15. 
The regularization factor $\lambda$ (see Eq.~\ref{eq:overall_objective_function}) is set to 0.0001.
We tune the other hyperparameters by random search on the validation data as follows.
We consider different hyperparameter settings and report the best result on the validation data for all methods. 
Specifically, we define a range for each hyperparameter. 
We then use a random search with 100 random combinations to explore the hyperparameter space and identify a hyperparameter setting that gives the best result on the validation data among all the explored hyperparameter settings. 
We then report this best result and use this hyperparameter setting as the default setting. 
Next, we study the sensitivity of different hyperparameters. 
To do so, we vary a chosen hyperparameter in its range while fixing the other hyperparameters to their default settings. 
We proceed to provide the ranges for the hyperparameters.

For the proposed framework, we vary the number of \texttt{ST} blocks $L$, the number of \texttt{ST} stacks $M$, and the number of head $U$ in the multi-head attention among 1, 2, 3, 4, and 5.
We vary the number of convolution filters among 16, 32, 64, 128, and 256.
\textcolor{black}{We vary the consolidation buffer size $|\mathcal{B}_c|$ and update buffer size $|\mathcal{B}_u|$} among 5\%, 10\%, 15\%, 20\%, 25\%, and 30\% of the number of nodes $N^{\pi}$.
For all the methods that involve \texttt{RNN} (i.e., \texttt{GRU}, \texttt{DCRNN}, and \texttt{EvolveGCN}), we vary the number of the hidden units among 16, 32, 64, 128, and 256, and the number of hidden layers among 1, 2, 3, 4, and 5.
For all the methods that involve \texttt{CNN} (i.e., \texttt{STGCN}, \texttt{GWN}, \texttt{MSTGCN}, and \texttt{ASTGCN}), we vary the number of convolution filters among 16, 32, 64, 128, and 256, and the number of hidden layers among 1, 2, 3, 4, and 5.
For all methods, we vary the number of \texttt{GCN} filters among 16, 32, 64, 128, and 256. 

\subsubsection{Implementation Details}
We implement the proposed framework and other baselines on Python 3.7, PyTorch 1.10, Geometric 2.0.4, and Sklearn 0.24. 
All experiments are conducted on a cluster server, which runs Linux Ubuntu 18.04.6 LTS.
The server is equipped with a ten-core Intel(R) Xeon(R) W-2155 CPU, 128 GBs RAM, and two GPUs Titan RTX each with 24 GBs VRAM.

\subsection{Experimental Results}
\subsubsection{Main Results}

Tables~\ref{tab:main_result_1} and \ref{tab:main_result_2} show the overall accuracy and runtime of the proposed framework and the baselines.
In the first scenario, \texttt{CAST} achieves the best accuracy on both datasets.
The only exception is that \texttt{CAST} is behind \texttt{DCRNN} in the short-term setting (i.e., 15 mins) on \textbf{PEMS03-Evolve}.
In terms of runtime, \texttt{CAST} is efficient and only is behind \texttt{GRU} and \texttt{STGCN}.
The accuracy of \texttt{PDFormer} is also high, and it is the runner-up on both datasets. However, \texttt{PDFormer} is very inefficient: it is the slowest baseline and is 12x slower than \texttt{CAST}.
In the second scenario, \texttt{TEAM} can maintain good accuracy on \textbf{PEMS03-Evolve}.
Further, on \textbf{PEMS04-Evolve}, \texttt{TEAM} is strongly competitive in terms of accuracy.
Considering runtime, \texttt{TEAM} exhibits very good efficiency, especially in the continual setting.
\texttt{TrafficStream} also exhibits very good efficiency. However, its accuracy is well below that of \texttt{TEAM}. 
On both datasets, \texttt{TEAM} is only slower than \texttt{GRU}, which does not perform any spatial computation.
In summary, the result indicates that \texttt{CAST} can outperform the baselines w.r.t. accuracy in the first scenario and that \texttt{TEAM} outperform the baselines w.r.t. runtime while maintaining competitive accuracy in the second scenario.

\subsubsection{Performance Improvement Significance}
To evaluate the hypothesis that whether the performance improvements of the proposed framework over the baselines are statistically significant, we conduct $t$-tests to assess the significance of the proposed framework against the baselines on the average results of all datasets. 
The $p$-values for all the metrics are below 0.005. 
This indicates that the performance improvements over the state-of-the-art methods are statistically significant.

\begin{table}[t]
    \vspace{-0.5em}
    \fontsize{7.5pt}{7.5pt}\selectfont
    \setlength{\tabcolsep}{3.0pt}
    \caption{Accuracy for stable and unstable nodes, \textbf{PEMS04-Evolve}.}
		\label{tab:stable_unstable}
	\begin{tabular}{l|p{1.3cm}|p{0.9cm}|p{0.7cm}|p{0.4cm}p{0.4cm}p{0.4cm}p{0.4cm}p{0.4cm}p{0.4cm}}
\toprule
\textbf{Scenario} & \textbf{Model}  & \textbf{Nodes} &   \textbf{Metric} & \textbf{May} & \textbf{Jun} & \textbf{Jul} & \textbf{Aug} & \textbf{Sep} & \textbf{Oct} \\
\midrule
\multirow{12}{*}{1st} & \multirow{6}{*}{\texttt{EnhanceNet}} &    & 
\textit{MAE}  &1.65	&1.84	&2.16	&2.16	&1.67	&1.90
 \\
         & &     Stable          & \textit{RMSE} &3.78	&4.20	&4.13	&4.92	&3.92	&3.95
 \\
         & &               & \textit{MAPE} &3.51	&4.08	&4.42	&5.57	&3.93	&3.83
 \\
         \cline{3-10}
&  &    &  \textit{MAE}  &1.37	&2.14	&1.71	&1.17	&1.34	&1.49
  \\
         & &     Unstable          & \textit{RMSE} &2.95	&4.82	&3.53	&3.34	&2.62	&2.85
  \\
         & &               & \textit{MAPE} &2.29	&2.34	&2.94	&2.25	&2.95	&3.62
  \\
         \cline{2-10}
& \multirow{6}{*}{\texttt{CAST}}    &  & \textit{MAE}  &\textbf{0.34}	&\textbf{0.28}	&\textbf{1.08}	&\textbf{0.19}	&\uline{1.01}	&\uline{1.36}
  \\
         & &     Stable          & \textit{RMSE} &\textbf{2.25}	&\textbf{1.18}	&\textbf{2.89}	&\textbf{1.03}	&\uline{1.94}	&\uline{2.85}
  \\
         & &               & \textit{MAPE} &\textbf{1.13}	&\textbf{0.51}	&\textbf{2.34}	&\textbf{0.31}	&\uline{1.63}	&\uline{2.39}
  \\
         \cline{3-10}
&  &  &  \textit{MAE}  &\textbf{0.83}	&\textbf{1.41}	&\textbf{1.12}	&\textbf{1.16}	&\uline{1.21}	&\uline{1.64}
  \\
         & &     Unstable          & \textit{RMSE} &\textbf{1.61}	&\textbf{3.00}	&\textbf{2.32}	&\textbf{2.09}	&\uline{3.46}	&\uline{3.31}
\\
         & &               & \textit{MAPE} &\textbf{1.31}	&\textbf{2.89}	&\textbf{1.92}	&\textbf{1.88}	&\uline{3.14}	&\uline{2.94}
  \\
         \midrule
\multirow{12}{*}{2nd} & \multirow{6}{*}{\pbox{15cm}{\texttt{Traffic}\\\texttt{Stream}}} &  & \textit{MAE}  &1.41	&1.31	&1.66	&1.37	&1.79	&2.01
  \\
         & &     Stable          & \textit{RMSE} &2.44	&3.05	&3.63	&2.16	&3.11	&4.17
  \\
         & &               & \textit{MAPE} &2.29	&2.34	&2.94	&2.25	&2.95	&3.63
  \\
         \cline{3-10}
&  &    &  \textit{MAE}  &4.33	&2.31	&2.39	&1.66	&3.41	&2.59
  \\
         & &     Unstable          & \textit{RMSE} &9.79	&4.73	&5.34	&2.98	&6.72	&4.98
  \\
         & &               & \textit{MAPE} &13.19	&4.69	&5.01	&2.75	&9.27	&4.78
  \\
         \cline{2-10}
& \multirow{6}{*}{\texttt{TEAM}}    &  & \textit{MAE}  &\uline{0.82}	&\uline{0.84}	&\uline{1.29}	&\uline{0.99}	&\textbf{0.91}	&\textbf{0.85}
  \\
         & &       Stable        & \textit{RMSE} &\uline{1.61}	&\uline{1.84}	&\uline{2.44}	&\uline{2.11}	&\textbf{2.94}	&\textbf{1.14}
  \\
         & &               & \textit{MAPE} &\uline{1.31}	&\uline{1.41}	&\uline{2.18}	&\uline{1.68}	&\textbf{3.27}	&\textbf{1.31}
  \\
         \cline{3-10}
&  &    &  \textit{MAE} &\uline{1.88}	&\uline{1.55}	&\uline{1.81}	&\uline{1.20} &\textbf{0.99}	&\textbf{1.40}
  \\
         & &       Unstable        & \textit{RMSE} &\uline{4.32}	&\uline{3.21}	&\uline{3.44}	&\uline{2.13}	&\textbf{1.98}	&\textbf{2.84}
  \\
         & &               & \textit{MAPE} &\uline{4.26}	&\uline{3.29}	&\uline{3.29}	&\uline{1.96}	&\textbf{1.61}	&\textbf{2.48}
  \\
\bottomrule
\end{tabular}
\end{table}

\begin{table}[t]
\fontsize{7.5pt}{7.5pt}\selectfont
\setlength{\tabcolsep}{3.0pt}
\caption{Effect of sampling strategy, \textbf{PEMS04-Evolve}.}
\label{tab:sampling}
\begin{tabular}{l|p{0.50cm}p{0.52cm}p{0.58cm}|p{0.50cm}p{0.52cm}p{0.58cm}|p{0.50cm}p{0.52cm}p{0.58cm}}
\toprule
\multirow{2}{*}{\textbf{Sampling}}           & \multicolumn{3}{c|}{\textbf{15 min}}          & \multicolumn{3}{c|}{\textbf{30 min}}          & \multicolumn{3}{c}{\textbf{60 min}}          \\
\cline{2-10}
           & \textit{MAE} & \textit{RMSE} & \textit{MAPE} & \textit{MAE} & \textit{RMSE} & \textit{MAPE} & \textit{MAE} & \textit{RMSE} & \textit{MAPE} \\
\midrule
Random     & 2.66         & 3.86          & 4.74          & 2.88         & 4.31          & 5.23          & 3.20         & 4.98          & 6.01          \\
Random Walk     & 2.61         & 3.74          & 4.56          & 2.77         & 4.20          & 5.14          & 3.16         & 4.89          & 5.88          \\
Degree     & \uline{1.59}         & \uline{2.81}          & \uline{3.07}          & \uline{1.83}         & \uline{3.40}          & \uline{3.62}          & \uline{2.24}         & \uline{4.24}          & \uline{4.57}          \\
Closeness  & 1.89         & 3.10          & 3.59          & 2.25         & 3.66          & 4.06          & 2.85         & 4.65          & 5.15          \\
Betweenness & 1.98         & 3.11          & 3.65          & 2.30         & 3.74          & 4.13          & 2.92         & 4.69          & 5.22          \\
\texttt{PageRank} & 1.78         & 2.89          & 3.15          & 1.96         & 3.55          & 3.71          & 2.31         & 4.29          & 4.75          \\
\midrule
\texttt{TEAM}    & \textbf{1.37}    & \textbf{2.66} & \textbf{2.61} & \textbf{1.60}    & \textbf{3.29} & \textbf{3.31} & \textbf{2.05}    & \textbf{4.21} & \textbf{4.25} \\
\bottomrule
\end{tabular}
\end{table}

\begin{table}[t]
    \fontsize{7.5pt}{7.5pt}\selectfont
    \setlength{\tabcolsep}{3.0pt}
    \caption{Accuracy and avg. runtime (s) vs. update frequencies, \textbf{PEMS04-Evolve}.}
		\label{tab:update_frequency}
	\begin{tabular}{p{0.95cm}|p{1.3cm}|p{0.9cm}|p{0.6cm}p{0.7cm}p{0.7cm}p{0.85cm}p{0.75cm}}
\toprule
\textbf{Scenario} & \textbf{Model}      &   \textbf{Metric}   & \textbf{Day} & \textbf{Week} & \textbf{Month} & \textbf{Quarter} & \textbf{Year} \\
\midrule
\multirow{8}{*}{1st} & \multirow{4}{*}{\texttt{EnhanceNet}}    & 
\textit{MAE}   & 11.51  & 3.44   & 1.76    & 1.92      & 2.05   \\
         &               & \textit{RMSE}  & 15.28  & 6.35   & 3.84    & 4.07      & 4.17   \\
         &               & \textit{MAPE}  & 51.83  & 8.63   & 3.75    & 3.77      & 3.79   \\
         &               & \textit{Avg. RT} & 4.23 & 10.66 & 24.08   & 75.47     & 305.07 \\
         \cline{2-8}
& \multirow{4}{*}{\texttt{CAST}}          & \textit{MAE}  & \uline{11.41}   & \uline{2.86}   & \textbf{1.60}    & \textbf{1.79}      & \textbf{1.95}   \\
         &               & \textit{RMSE}  & \uline{15.14}   & \uline{4.97}   & \textbf{3.27}    & \textbf{3.30}      & \textbf{3.41}   \\
         &               & \textit{MAPE}  & \uline{51.56}   & \uline{7.17}   & \textbf{3.28}    & \textbf{3.46}      & \textbf{3.60}   \\
         &               & \textit{Avg. RT} & 3.07 & 6.92  & 17.82   & 56.41     & 227.26 \\ \hline
\multirow{8}{*}{2nd} & \multirow{4}{*}{\pbox{15cm}{\texttt{Traffic}\\\texttt{Stream}}} & \textit{MAE}  & 11.72  & 3.16   & 2.16    & 2.21      & 2.38   \\
         &               & \textit{RMSE}  & 15.78  & 5.34   & 3.89    & 4.05      & 4.11   \\
         &               & \textit{MAPE}  & 55.13  & 8.61   & 4.53    & 4.66      & 4.79   \\
         &               & \textit{Avg. RT} & \uline{2.28}  & \uline{4.21}  & \uline{10.24}   & \uline{23.43}     & \uline{73.18}  \\
         \cline{2-8}
& \multirow{4}{*}{\texttt{TEAM}}         & \textit{MAE}   & \textbf{10.19}
   & \textbf{2.82}   & \uline{1.67}    & \uline{1.85}      & \uline{1.99}   \\
         &               & \textit{RMSE}  & \textbf{13.99}
   & \textbf{4.85}   & \uline{3.38}    & \uline{3.59}      & \uline{3.64}   \\
         &               & \textit{MAPE}  & \textbf{47.45}
   & \textbf{6.72}   & \uline{3.39}    & \uline{3.44}      & \uline{3.69}   \\
         &               & \textit{Avg. RT} & \textbf{1.99} & \textbf{3.51}  & \textbf{9.11}    & \textbf{18.90}     & \textbf{57.49}  \\
\bottomrule
\end{tabular}
\end{table}

\begin{figure*}[t]
    \fontsize{7.5pt}{7.5pt}\selectfont
    \begin{minipage}[t]{0.33\textwidth} %
        \input{charts/number_of_blocks_nonfig}
    \end{minipage}
    \begin{minipage}[t]{0.33\textwidth} %
        \input{charts/number_of_stacks_nonfig}
    \end{minipage}
    \begin{minipage}[t]{0.33\textwidth} %
        \input{charts/number_of_heads_nonfig}
    \end{minipage}
\end{figure*}    
\begin{figure*}[t]
    \fontsize{7.5pt}{7.5pt}\selectfont
    \begin{minipage}[t]{0.33\textwidth} %
        \input{charts/number_of_filters_nonfig}
    \end{minipage}
    \begin{minipage}[t]{0.33\textwidth} %
        \input{charts/number_of_orders_nonfig}
    \end{minipage}
    \begin{minipage}[t]{0.33\textwidth} %
        \input{charts/number_of_conso_nodes_nonfig}
    \end{minipage}
\end{figure*}
\begin{figure*}[t]
    \fontsize{7.5pt}{7.5pt}\selectfont
    \begin{minipage}[t]{0.33\textwidth} %
        \input{charts/number_of_update_nodes_nonfig}
    \end{minipage}
    \begin{minipage}[t]{0.33\textwidth} %
        \input{charts/number_of_tau_nonfig}
    \end{minipage}
\end{figure*}

\subsubsection{Ablation Study}
We study the effect of each component in the proposed framework by removing the convolution (denoted as \texttt{TEAM-CONV}), removing the attention (denoted as \texttt{TEAM-ATT}), swapping the convolution and the attention position, i.e., the input is fed into the convolution first, then the output of the convolution is fed to the attention (denoted as \texttt{TEAM-SWAP}), removing the backcast convolution (denoted as \texttt{TEAM-BACK}), removing the stack residual (see Eq.~\ref{eq:stack_residual}), i.e., the output of the entire model only is the output of the last \texttt{ST} stack (denoted as \texttt{TEAM-RES}), and remove the continual module (denoted as \texttt{TEAM-CONT}).  
All the variants perform in the second scenario.
We report findings on dataset \textbf{PEMS04-Evolve} only. 
The results on \textbf{PEMS03-Evolve} show similar trends. 
Table~\ref{tab:ablation} shows the results w.r.t. \textit{MAE}, \textit{RMSE}, and \textit{MAPE}. 
We do not include the runtime in Table~\ref{tab:ablation} because the runtime of \texttt{TEAM} variants are only slightly different from the entire framework.
The results show that \texttt{TEAM-CONT} performs poorly and becomes the worst variant. 
This suggests that the continual learning module plays an important role in the proposed framework to enable the ability to train \texttt{TEAM} on evolved parts of RNs.
Both \texttt{TEAM-CONT} and \texttt{TEAM-ATT} perform worse than \texttt{TEAM}.
This suggests that the combination of convolution and attention improves the accuracy on incremental time series.
Next, \texttt{TEAM-SWAP}, \texttt{TEAM-BACK}, and \texttt{TEAM-RES} perform worse than \texttt{TEAM}.
This suggests that our doubly residual design and the proposed architecture of \texttt{TEAM} are efficient.

\subsubsection{Empirical Complexity Study}
Adding to the complexity analysis in Section~\ref{sec:comp_analysis}, we study the complexity empirically by considering the effect of the number of evolved nodes on the training time. We thus adjust the number of added and removed nodes and observe the total training time. We set the updating and the consolidation buffer size to 10. We also report the preprocessing time of the continual learning module (see lines 1--12 in Algorithm 1). The results on \textbf{PEMS04-Evolve} are presented in Table~\ref{tab:correlation}. Similar trends are observed on \textbf{PEMS03-Evolve}. The results show that the number of trained nodes is roughly two to three times the number of evolved nodes. Next, the training time is low compared to the baselines (see Table~\ref{tab:main_result_2}). This is evidence of the effectiveness of the continual learning module at reducing the training time by focusing only on essential nodes. The results also show that the preprocessing time of the continual learning module is low, indicating that the additional overhead of the continual learning module is negligible.

\subsubsection{Accuracy during Immediate Periods}
In addition to reporting on the accuracy for the last period as in Tables~\ref{tab:main_result_1} and~\ref{tab:main_result_2}, we report on the accuracy during immediate periods (i.e., accuracy across all periods). We adopt the same two scenarios as for the main result. For brevity, we only report the average result across three horizons of our proposal and \texttt{EnhanceNet} and \texttt{TrafficStream} on \textbf{PEMS04-Evolve}. The results for the other baselines on both \textbf{PEMS03-Evolve} and \textbf{PEMS04-Evolve} share the same characteristics. The results in Table~\ref{tab:immedate_period} show that \texttt{CAST} always outperforms the \texttt{EnhanceNet} in the first scenario and that \texttt{TEAM} always outperforms \texttt{TrafficStream} in the second scenario. \texttt{TEAM} is only insignificantly behind \texttt{EnhanceNet} in May, June, July, and August, and \texttt{TEAM} outperforms \texttt{EnhanceNet} in the first and the two last months. This is because \texttt{TEAM} can learn from historical periods and use the resulting knowledge to improve forecasting accuracy in future periods.  

\subsubsection{Accuracy for Stable and Unstable Nodes}
In addition to reporting on the accuracy across all nodes as in Tables~\ref{tab:main_result_1} and~\ref{tab:main_result_2}, we report on the accuracy for only stable and unstable nodes. Specifically, we report on the accuracy for nodes in the buffers $|\mathcal{B}_{c}|$ and $|\mathcal{B}_{u}|$. We again adopt the two scenarios from the main results. We only report average results across three horizons of our proposal and \texttt{EnhanceNet} and \texttt{TrafficStream} on \textbf{PEMS04-Evolve}. The results for the other baselines on both datasets share the same characteristics. Table~\ref{tab:stable_unstable} shows that \texttt{CAST} outperforms \texttt{EnhanceNet} w.r.t. accuracy for both stable and unstable nodes in the first scenario. Similarly, \texttt{TEAM} outperforms \texttt{TrafficStream} w.r.t. accuracy for both stable and unstable nodes in the second scenario. Moreover, \texttt{TEAM} outperforms \texttt{CAST} in the two last periods. This is because \texttt{TEAM} can learn from historical periods and use the resulting knowledge to improve forecasting accuracy in future periods. 

\vspace{-0.5em}
\subsubsection{Effect of Sampling Strategy}
In addition to reporting on the accuracy using the proposed sampling strategy for $\mathcal{B}_{c}$ and $\mathcal{B}_{u}$ (see Algorithm~1), we report on the accuracy for other sampling strategies. (1) We randomly select~\cite{DBLP:conf/kdd/LeskovecF06} $|\mathcal{B}_{c}|$ and $|\mathcal{B}_{u}|$ nodes for $\mathcal{B}_{c}$ and $\mathcal{B}_{u}$, respectively;  (2) We use random walks~\cite{DBLP:conf/kdd/LeskovecF06} to select $|\mathcal{B}_{c}|$ and $|\mathcal{B}_{u}|$ nodes for $\mathcal{B}_{c}$ and $\mathcal{B}_{u}$, respectively; (3) We use a degree matrix~\cite{DBLP:conf/dagstuhl/2004na} to select $|\mathcal{B}_{c}|$ nodes with highest degree and $|\mathcal{B}_{u}|$ nodes with lowest degree for $\mathcal{B}_{c}$ and $\mathcal{B}_{u}$, respectively; (4) We use the closeness centrality metric~\cite{DBLP:conf/dagstuhl/2004na} to select $|\mathcal{B}_{c}|$ nodes with highest centrality and $|\mathcal{B}_{u}|$ nodes with lowest centrality for $\mathcal{B}_{c}$ and $\mathcal{B}_{u}$, respectively; (5) We use the betweenness centrality metric~\cite{DBLP:conf/dagstuhl/2004na} to select $|\mathcal{B}_{c}|$ nodes with highest centrality and $|\mathcal{B}_{u}|$ nodes with lowest centrality for $\mathcal{B}_{c}$ and $\mathcal{B}_{u}$, respectively; (6) We use \texttt{PageRank}~\cite{DBLP:conf/dagstuhl/2004na} to select $|\mathcal{B}_{c}|$ nodes with highest rank and $|\mathcal{B}_{u}|$ nodes with lowest rank for $\mathcal{B}_{c}$ and $\mathcal{B}_{u}$, respectively. For brevity, we only report the result for the last month on \textbf{PEMS04-Evolve}. The result on \textbf{PEMS03-Evolve} shares the same characteristics. Table~\ref{tab:sampling} shows that the random sampling and random walk strategies perform the worst and that the sampling strategies based on degree, centrality, and \texttt{PageRank} perform better. This suggests that the latter three strategies can be used to select stable and unstable nodes. Our proposal achieves the best accuracy. This suggests that using temporal information (i.e., $\tau$), as does our strategy, is important.

\subsubsection{Effect of Evolving Frequency}
We study the effect of evolving frequency $\pi$. Thus, in addition to capturing the evolution of RNs every month (i.e., $\pi = 1$ month) as in the default setting, we capture the evolution of RNs every day, every week, every quarter, and every year. We synthesize evolved RNs by randomly adding and removing 10--15\% of the nodes and 20--25\% of the edges every period. We only report the average result across three horizons for the last period (i.e., the 7-th day, 7-th week, the 7-th quarter, and the 7-th year) on \textbf{PEMS04-Evolve}. The results on \textbf{PEMS03-Evolve} share similar characteristics. Table~\ref{tab:update_frequency} shows that \texttt{TEAM} achieves the best accuracy compared to the baselines when the RNs evolve rapidly ($\pi \le 1$ week). This is because \texttt{TEAM} can leverage knowledge extracted from previous periods, mitigating the impact of limited training samples (e.g., 288 samples for a day and 1440 samples for a week) that reduce the performance of the other methods that reinitialize and retrain for every period. When RNs evolve more slowly, e.g, $\pi \ge 1$ month, the methods that are restricted to the first scenario perform better than \texttt{TEAM} due to having sufficient training samples, and \texttt{CAST} outperforms the other baselines w.r.t accuracy. The results exhibit the efficiency and accuracy of \texttt{TEAM} when learning with limited training samples and adapting to rapidly evolving RNs.

\subsubsection{Effect of the Number of \texttt{ST} Blocks $L$}
We study the effect of the number of \texttt{ST} blocks $L$ in the proposed framework.
We vary $L$ among 1, 2, 3, 4, and 5.
Due to the space limitation, we report on the effect of the number of \texttt{ST} block $L$ on dataset \textbf{PEMS04-Evolve} only.
The results on \textbf{PEMS03-Evolve} show similar trends. 
Fig.~\ref{fig:effect_of_L} shows the results w.r.t. \textit{MAE}, \textit{RMSE}, \textit{MAPE}, and \textit{runtime}.
The proposed framework achieves the best performance when $L=3$, which is the default value.
When $L < 3$, the framework has not learned a good representation due to the underfitting.
When $L > 3$, the framework consists of more parameters, which may be easier to be overfitting or to be trapped in the local optimum.
When $L = 3$, the best trade-off is achieved.

\subsubsection{Effect of the Number of \texttt{ST} Stacks $M$}
We study the effect of the number of \texttt{ST} stacks $M$ in the proposed framework.
In particular, we vary $M$ among 1, 2, 3, 4, and 5.
Due to the space limitation, we report on the effect of the number of \texttt{ST} stacks $M$ on dataset \textbf{PEMS04-Evolve} only.
The results on \textbf{PEMS03-Evolve} show similar trends. 
Fig.~\ref{fig:effect_of_M} shows the results w.r.t. \textit{MAE}, \textit{RMSE}, \textit{MAPE}, and \textit{runtime}.
The proposed framework achieves the best performance when $M=3$, which is the default value.
When $M < 3$, the framework has not learned a good representation due to the underfitting.
When $M > 3$, the framework consists of more parameters, which may be easier to be overfitting or to be trapped in the local optimum.
When $M = 3$, the best trade-off is achieved.

\subsubsection{Effect of the Number of Attention Heads $U$}
We study the effect of the number of head $U$ in the multi-head attention (see Eq.~\ref{eq:spatial_attention_output_multi}).
In particular, we vary $U$ among 1, 2, 3, 4, and 5.
Due to the space limitation, we report on the effect of the number of \texttt{ST} stacks $M$ on dataset \textbf{PEMS04-Evolve} only.
The results on \textbf{PEMS03-Evolve} show similar trends. 
Fig.~\ref{fig:effect_of_U} shows the results w.r.t. \textit{MAE}, \textit{RMSE}, \textit{MAPE}, and \textit{runtime}.
The proposed framework achieves the best performance when $U=3$, which is the default value.
When $U < 3$, the framework has not learned a good representation due to the underfitting.
When $U > 3$, the framework consists of more parameters, which may be easier to be overfitting or to be trapped in the local optimum.
When $U = 3$, the best trade-off is achieved.

\subsubsection{Effect of the Number of Temporal Convolution Filters}
We study the effect of the number of temporal convolution filters $|\mathcal{W}_{t_{1}}|$ in the filter bank $\mathcal{W}_{t_{1}}$.
In particular, we vary $|\mathcal{W}_{t_{1}}|$ among 16, 32, 64, 128, and 256.
Due to the space limitation, we report on the effect of the number of convolution filters $|\mathcal{W}_{t_{1}}|$ on dataset \textbf{PEMS04-Evolve} only.
The results on \textbf{PEMS03-Evolve} show similar trends. 
Fig.~\ref{fig:effect_of_W} shows the results w.r.t. \textit{MAE}, \textit{RMSE}, \textit{MAPE}, and \textit{runtime}.
The proposed framework achieves the best performance when $|\mathcal{W}_{t_{1}}| = 64$, which is the default value.
When $|\mathcal{W}_{t_{1}}| < 64$, the framework has not learned a good representation due to the underfitting.
When $|\mathcal{W}_{t_{1}}| > 64$, the framework consists of more parameters, which may be easier to be overfitting or to be trapped in the local optimum.
When $|\mathcal{W}_{t_{1}}| = 64$, the best trade-off is achieved.

\subsubsection{Effect of the Order $o$ in Chebyshev Function}
We study the effect of the order $o$ in Chebyshev function (see Eq.~\ref{eq:chebyshev}).
In particular, we vary $o$ among 2, 3, 4, 5, and 6.
Due to the space limitation, we report on the effect of the order $o$ on dataset \textbf{PEMS04-Evolve} only.
The results on \textbf{PEMS03-Evolve} show similar trends. 
Fig.~\ref{fig:effect_of_o} shows the results w.r.t. \textit{MAE}, \textit{RMSE}, \textit{MAPE}, and \textit{runtime}.
The proposed framework achieves the best performance when $o=3$, which is the default value.
When $o < 3$, the framework has not learned a good representation due to the underfitting.
When $o > 3$, the framework consists of more computations, which may be easier to be overfitting or to be trapped in the local optimum.
When $o = 3$, the best trade-off is achieved.

\subsubsection{Effect of the Buffer Size $|\mathcal{B}_{c}|$ and $|\mathcal{B}_{u}|$}
\textcolor{black}{We study the effect of the consolidation buffer size $|\mathcal{B}_{c}|$ and update buffer size $|\mathcal{B}_{u}|$.
In particular, we vary one of $|\mathcal{B}_{c}|$ and $|\mathcal{B}_{u}|$ among 5\%, 10\%, 15\%, 20\%, 25\%, and 30\%, while fixing the value at 15\% for the other.
Due to the space limitation, we report on the effect of the consolidation buffer size $|\mathcal{B}_{c}|$ and update buffer size $|\mathcal{B}_{u}|$ on dataset \textbf{PEMS04-Evolve} only.
The results on \textbf{PEMS03-Evolve} show similar trends. 
Fig.~\ref{fig:effect_of_B_c} and Fig.~\ref{fig:effect_of_B_u} show the results w.r.t. \textit{MAE}, \textit{RMSE}, \textit{MAPE}, and \textit{runtime} when varying $|\mathcal{B}_c|$ and $|\mathcal{B}_{u}|$, respectively.
The proposed framework achieves the best performance when $|\mathcal{B}_{c}|=|\mathcal{B}_{u}|=15\%$, which is the default value.
When $|\mathcal{B}_{c}| < 15\%$, the framework lacks enough historical data for rehearsal and almost trains only on the data of newly added nodes and thus cannot exploit the continual module for revisiting the learned knowledge, which yields sub-optimal performance.
When $|\mathcal{B}_c| > 15\%$, the framework exploits too much historical knowledge from historical nodes and fails to learn the information from newly added nodes, which also yields sub-optimal performance.
When $|\mathcal{B}_{u}| < 15\%$, the model may not update nodes that have different patterns, thus preserving inaccurate knowledge, causing substandard performance.
When $|\mathcal{B}_{u}| > 15\%$, the model requires frequent updates to its knowledge, which may conflict with previously learned historical knowledge, leading to deficient outcomes.
When $|\mathcal{B}_c| = |\mathcal{B}_{u}| = 15\%$, the best trade-off is achieved.}

\subsubsection{Effect of the Length of Sampling Period $\tau$}
We study the effect of the length of sampling period $\tau$ in data histogram construction. Specifically,  we vary $\tau$  among 1, 3, 5, 7, 9, 11, and 13 days.
Due to the space limitation, we report on the effect of the length of sampling period $\tau$ on dataset \textbf{PEMS04-Evolve} only. 
The results on \textbf{PEMS03-Evolve} show similar trends.
Fig.~\ref{fig:effect_of_tau} shows the results w.r.t. \textit{MAE}, \textit{RMSE}, \textit{MAPE}, and \textit{runtime}.
The proposed framework achieves the best performance when $\tau = 7$ days, which is the default value.
When $\tau < 7$ days, the constructed histogram does not have complete observations of traffic flow patterns. Many temporal patterns, such as the correlation between weekdays and weekends, are excluded from the observations. In this case, \textrm{EMD} might not accurately measure the consistency of existing nodes.
When $\tau > 7$ days, the results are insignificantly improved while the buffer size $|\mathcal{B}_{c}|$ has to store a larger amount of historical data, which leads to increased computational and storage costs.

%% file: charts/number_of_blocks_nonfig.tex
    \begin{subfigure}[b]{0.49\linewidth}
        \begin{tikzpicture}
            \begin{axis}[
            xtick=data,
            ytick={1.2, 1.6, 2.0, 2.4},
            y tick label style={
                /pgf/number format/.cd,
                fixed,
                fixed zerofill,
                precision=1,
                /tikz/.cd
            },
            xlabel=$L$,
            ylabel=\textit{MAE},
            ymin=1.2,
            ymax=2.4,
            width=1.21\linewidth,
            height=0.48*\axisdefaultheight,
            legend style={at={(1.2,1.4)},anchor=north,legend columns=-1}]
            \addplot[red, line width=0.6pt, mark=triangle*] table[x=L, y=MAE_15] {data/number_of_blocks.dat};
            \addlegendentry{15 mins}
            \addplot[black, line width=0.6pt, mark=square*] table[x=L, y=MAE_30] {data/number_of_blocks.dat};
            \addlegendentry{30 mins}
            \addplot[blue, line width=0.6pt, mark=*] table[x=L, y=MAE_60] {data/number_of_blocks.dat};
            \addlegendentry{60 mins}
        \end{axis}
        \end{tikzpicture}
        \caption{\textit{MAE}} 
        \label{subfig:effect_of_L_MAE}  
    \end{subfigure}
    \begin{subfigure}[b]{0.49\linewidth}
      \begin{tikzpicture}
        \begin{axis}[
            xtick=data,
            ytick={2.0, 3.0, 4.0, 5.0},
            y tick label style={
                /pgf/number format/.cd,
                fixed,
                fixed zerofill,
                precision=1,
                /tikz/.cd
            },
            xlabel=$L$,
            ylabel=\textit{RMSE},
            ymin=2.0,
            ymax=5.0,
            width=1.21\linewidth,
            height=0.48*\axisdefaultheight,
            legend style={at={(1.7,1.35)},anchor=north,legend columns=-1}]
            \addplot[red, line width=0.6pt, mark=triangle*] table[x=L, y=RMSE_15] {data/number_of_blocks.dat};
            \addplot[black, line width=0.6pt, mark=square*] table[x=L, y=RMSE_30] {data/number_of_blocks.dat};
            \addplot[blue, line width=0.6pt, mark=*] table[x=L, y=RMSE_60] {data/number_of_blocks.dat};
            \end{axis}
        \end{tikzpicture}
        \caption{\textit{RMSE}} 
        \label{subfig:effect_of_L_RMSE}  
      \end{subfigure}
    \begin{subfigure}[b]{0.49\linewidth}
      \begin{tikzpicture}
        \begin{axis}[
            xtick=data,
            ytick={2.0, 3.0, 4.0, 5.0},
            y tick label style={
                /pgf/number format/.cd,
                fixed,
                fixed zerofill,
                precision=1,
                /tikz/.cd
            },
            xlabel=$L$,
            ylabel=\textit{MAPE},
            ymin=2.0,
            ymax=5.0,
            width=1.21\linewidth,
            height=0.48*\axisdefaultheight,
            legend style={at={(1.7,1.35)},anchor=north,legend columns=-1}]
            \addplot[red, line width=0.6pt, mark=triangle*] table[x=L, y=MAPE_15] {data/number_of_blocks.dat};
            \addplot[black, line width=0.6pt, mark=square*] table[x=L, y=MAPE_30] {data/number_of_blocks.dat};
            \addplot[blue, line width=0.6pt, mark=*] table[x=L, y=MAPE_60] {data/number_of_blocks.dat};
            \end{axis}
        \end{tikzpicture}
        \caption{\textit{MAPE}} 
        \label{subfig:effect_of_L_MAPE}  
      \end{subfigure}
      \begin{subfigure}[b]{0.49\linewidth}
      \begin{tikzpicture}
        \begin{axis}[
            xtick=data,
            ytick={1.0, 2.0, 3.0, 4.0},
            y tick label style={
                /pgf/number format/.cd,
                fixed,
                fixed zerofill,
                precision=1,
                /tikz/.cd
            },
            xlabel=$L$,
            ylabel=\textit{Runtime ($10^3$s)},
            ymin=1.0,
            ymax=4.0,
            width=1.21\linewidth,
            height=0.48*\axisdefaultheight,
            legend style={at={(1.7,1.35)},anchor=north,legend columns=-1}]
            \addplot[orange, line width=0.6pt, mark=pentagon*] table[x=L, y=runtime] {data/number_of_blocks.dat};
            \end{axis}
        \end{tikzpicture}
        \caption{\textit{Runtime} ($10^{3}s$)} 
        \label{subfig:effect_of_L_runtime}  
      \end{subfigure}
\caption{Effect of $L$, \textbf{PEMS04-Evolve}.}
\label{fig:effect_of_L}

%% file: charts/number_of_stacks_nonfig.tex
    \begin{subfigure}[b]{0.49\linewidth}
        \begin{tikzpicture}
            \begin{axis}[
            xtick=data,
            ytick={1.2, 1.6, 2.0, 2.4},
            y tick label style={
                /pgf/number format/.cd,
                fixed,
                fixed zerofill,
                precision=1,
                /tikz/.cd
            },
            xlabel=$M$,
            ylabel=\textit{MAE},
            ymin=1.2,
            ymax=2.4,
            width=1.21\linewidth,
            height=0.48*\axisdefaultheight,
            legend style={at={(1.2,1.4)},anchor=north,legend columns=-1}]
            \addplot[red, line width=0.6pt, mark=triangle*] table[x=M, y=MAE_15] {data/number_of_stacks.dat};
            \addlegendentry{15 mins}
            \addplot[black, line width=0.6pt, mark=square*] table[x=M, y=MAE_30] {data/number_of_stacks.dat};
            \addlegendentry{30 mins}
            \addplot[blue, line width=0.6pt, mark=*] table[x=M, y=MAE_60] {data/number_of_stacks.dat};
            \addlegendentry{60 mins}
        \end{axis}
        \end{tikzpicture}
        \caption{\textit{MAE}} 
        \label{subfig:effect_of_M_MAE}  
    \end{subfigure}
    \begin{subfigure}[b]{0.49\linewidth}
      \begin{tikzpicture}
        \begin{axis}[
            xtick=data,
            ytick={2.0, 3.0, 4.0, 5.0},
            y tick label style={
                /pgf/number format/.cd,
                fixed,
                fixed zerofill,
                precision=1,
                /tikz/.cd
            },
            xlabel=$M$,
            ylabel=\textit{RMSE},
            ymin=2.0,
            ymax=5.0,
            width=1.21\linewidth,
            height=0.48*\axisdefaultheight,
            legend style={at={(1.7,1.35)},anchor=north,legend columns=-1}]
            \addplot[red, line width=0.6pt, mark=triangle*] table[x=M, y=RMSE_15] {data/number_of_stacks.dat};
            \addplot[black, line width=0.6pt, mark=square*] table[x=M, y=RMSE_30] {data/number_of_stacks.dat};
            \addplot[blue, line width=0.6pt, mark=*] table[x=M, y=RMSE_60] {data/number_of_stacks.dat};
            \end{axis}
        \end{tikzpicture}
        \caption{\textit{RMSE}} 
        \label{subfig:effect_of_M_RMSE}  
      \end{subfigure}
    \begin{subfigure}[b]{0.49\linewidth}
      \begin{tikzpicture}
        \begin{axis}[
            xtick=data,
            ytick={2.0, 3.0, 4.0, 5.0},
            y tick label style={
                /pgf/number format/.cd,
                fixed,
                fixed zerofill,
                precision=1,
                /tikz/.cd
            },
            xlabel=$M$,
            ylabel=\textit{MAPE},
            ymin=2.0,
            ymax=5.0,
            width=1.21\linewidth,
            height=0.48*\axisdefaultheight,
            legend style={at={(1.7,1.35)},anchor=north,legend columns=-1}]
            \addplot[red, line width=0.6pt, mark=triangle*] table[x=M, y=MAPE_15] {data/number_of_stacks.dat};
            \addplot[black, line width=0.6pt, mark=square*] table[x=M, y=MAPE_30] {data/number_of_stacks.dat};
            \addplot[blue, line width=0.6pt, mark=*] table[x=M, y=MAPE_60] {data/number_of_stacks.dat};
            \end{axis}
        \end{tikzpicture}
        \caption{\textit{MAPE}} 
        \label{subfig:effect_of_M_MAPE}  
      \end{subfigure}
      \begin{subfigure}[b]{0.49\linewidth}
      \begin{tikzpicture}
        \begin{axis}[
            xtick=data,
            ytick={1.0, 2.0, 3.0, 4.0},
            y tick label style={
                /pgf/number format/.cd,
                fixed,
                fixed zerofill,
                precision=1,
                /tikz/.cd
            },
            xlabel=$M$,
            ylabel=\textit{Runtime ($10^3$s)},
            ymin=1.0,
            ymax=4.0,
            width=1.21\linewidth,
            height=0.48*\axisdefaultheight,
            legend style={at={(1.7,1.35)},anchor=north,legend columns=-1}]
            \addplot[orange, line width=0.6pt, mark=pentagon*] table[x=M, y=runtime] {data/number_of_stacks.dat};
            \end{axis}
        \end{tikzpicture}
        \caption{\textit{Runtime} ($10^{3}s$)} 
        \label{subfig:effect_of_M_runtime}  
      \end{subfigure}
\caption{Effect of $M$, \textbf{PEMS04-Evolve}.}
\label{fig:effect_of_M}

%% file: charts/number_of_heads_nonfig.tex
    \begin{subfigure}[b]{0.49\linewidth}
        \begin{tikzpicture}
            \begin{axis}[
            xtick=data,
            ytick={1.2, 1.6, 2.0, 2.4},
            y tick label style={
                /pgf/number format/.cd,
                fixed,
                fixed zerofill,
                precision=1,
                /tikz/.cd
            },
            xlabel=$U$,
            ylabel=\textit{MAE},
            ymin=1.2,
            ymax=2.4,
            width=1.21\linewidth,
            height=0.48*\axisdefaultheight,
            legend style={at={(1.2,1.4)},anchor=north,legend columns=-1}]
            \addplot[red, line width=0.6pt, mark=triangle*] table[x=U, y=MAE_15] {data/number_of_heads.dat};
            \addlegendentry{15 mins}
            \addplot[black, line width=0.6pt, mark=square*] table[x=U, y=MAE_30] {data/number_of_heads.dat};
            \addlegendentry{30 mins}
            \addplot[blue, line width=0.6pt, mark=*] table[x=U, y=MAE_60] {data/number_of_heads.dat};
            \addlegendentry{60 mins}
        \end{axis}
        \end{tikzpicture}
        \caption{\textit{MAE}} 
        \label{subfig:effect_of_U_MAE}  
    \end{subfigure}
    \begin{subfigure}[b]{0.49\linewidth}
      \begin{tikzpicture}
        \begin{axis}[
            xtick=data,
            ytick={2.0, 3.0, 4.0, 5.0},
            y tick label style={
                /pgf/number format/.cd,
                fixed,
                fixed zerofill,
                precision=1,
                /tikz/.cd
            },
            xlabel=$U$,
            ylabel=\textit{RMSE},
            ymin=2.0,
            ymax=5.0,
            width=1.21\linewidth,
            height=0.48*\axisdefaultheight,
            legend style={at={(1.7,1.35)},anchor=north,legend columns=-1}]
            \addplot[red, line width=0.6pt, mark=triangle*] table[x=U, y=RMSE_15] {data/number_of_heads.dat};
            \addplot[black, line width=0.6pt, mark=square*] table[x=U, y=RMSE_30] {data/number_of_heads.dat};
            \addplot[blue, line width=0.6pt, mark=*] table[x=U, y=RMSE_60] {data/number_of_heads.dat};
            \end{axis}
        \end{tikzpicture}
        \caption{\textit{RMSE}} 
        \label{subfig:effect_of_U_RMSE}  
      \end{subfigure}
    \begin{subfigure}[b]{0.49\linewidth}
      \begin{tikzpicture}
        \begin{axis}[
            xtick=data,
            ytick={2.0, 3.0, 4.0, 5.0},
            y tick label style={
                /pgf/number format/.cd,
                fixed,
                fixed zerofill,
                precision=1,
                /tikz/.cd
            },
            xlabel=$U$,
            ylabel=\textit{MAPE},
            ymin=2.0,
            ymax=5.0,
            width=1.21\linewidth,
            height=0.48*\axisdefaultheight,
            legend style={at={(1.7,1.35)},anchor=north,legend columns=-1}]
            \addplot[red, line width=0.6pt, mark=triangle*] table[x=U, y=MAPE_15] {data/number_of_heads.dat};
            \addplot[black, line width=0.6pt, mark=square*] table[x=U, y=MAPE_30] {data/number_of_heads.dat};
            \addplot[blue, line width=0.6pt, mark=*] table[x=U, y=MAPE_60] {data/number_of_heads.dat};
            \end{axis}
        \end{tikzpicture}
        \caption{\textit{MAPE}} 
        \label{subfig:effect_of_U_MAPE}  
      \end{subfigure}
      \begin{subfigure}[b]{0.49\linewidth}
      \begin{tikzpicture}
        \begin{axis}[
            xtick=data,
            ytick={1.4, 1.8, 2.2, 2.6},
            y tick label style={
                /pgf/number format/.cd,
                fixed,
                fixed zerofill,
                precision=1,
                /tikz/.cd
            },
            xlabel=$U$,
            ylabel=\textit{Runtime ($10^3$s)},
            ymin=1.4,
            ymax=2.6,
            width=1.21\linewidth,
            height=0.48*\axisdefaultheight,
            legend style={at={(1.7,1.35)},anchor=north,legend columns=-1}]
            \addplot[orange, line width=0.6pt, mark=pentagon*] table[x=U, y=runtime] {data/number_of_heads.dat};
            \end{axis}
        \end{tikzpicture}
        \caption{\textit{Runtime} ($10^{3}s$)} 
        \label{subfig:effect_of_U_runtime}  
      \end{subfigure}
\caption{Effect of $U$, \textbf{PEMS04-Evolve}.}
\label{fig:effect_of_U}

%% file: charts/number_of_filters_nonfig.tex
    \begin{subfigure}[b]{0.49\linewidth}
        \begin{tikzpicture}
            \begin{axis}[
            xtick=data,
            xmode=log,
            xticklabels={16,32,64,128,256},
            ytick={1.2,1.6,2.0,2.4},
            y tick label style={
                /pgf/number format/.cd,
                fixed,
                fixed zerofill,
                precision=1,
                /tikz/.cd
            },
            xlabel=$|\mathcal{W}_{t_{1}}|$,
            ylabel=\textit{MAE},
            ymin=1.2,
            ymax=2.4,
            width=1.21\linewidth,
            height=0.48*\axisdefaultheight,
            legend style={at={(1.2,1.4)},anchor=north,legend columns=-1}]
            \addplot[red, line width=0.6pt, mark=triangle*] table[x=W, y=MAE_15] {data/number_of_filters.dat};
            \addlegendentry{15 mins}
            \addplot[black, line width=0.6pt, mark=square*] table[x=W, y=MAE_30] {data/number_of_filters.dat};
            \addlegendentry{30 mins}
            \addplot[blue, line width=0.6pt, mark=*] table[x=W, y=MAE_60] {data/number_of_filters.dat};
            \addlegendentry{60 mins}
        \end{axis}
        \end{tikzpicture}
        \caption{\textit{MAE}} 
        \label{subfig:effect_of_W_MAE}  
    \end{subfigure}
    \begin{subfigure}[b]{0.49\linewidth}
      \begin{tikzpicture}
        \begin{axis}[
            xtick=data,
            xmode=log,
            xticklabels={16,32,64,128,256},
            ytick={2.0, 3.0, 4.0, 5.0},
            y tick label style={
                /pgf/number format/.cd,
                fixed,
                fixed zerofill,
                precision=1,
                /tikz/.cd
            },
            xlabel=$|\mathcal{W}_{t_{1}}|$,
            ylabel=\textit{RMSE},
            ymin=2.0,
            ymax=5.0,
            width=1.21\linewidth,
            height=0.48*\axisdefaultheight,
            legend style={at={(1.7,1.35)},anchor=north,legend columns=-1}]
            \addplot[red, line width=0.6pt, mark=triangle*] table[x=W, y=RMSE_15] {data/number_of_filters.dat};
            \addplot[black, line width=0.6pt, mark=square*] table[x=W, y=RMSE_30] {data/number_of_filters.dat};
            \addplot[blue, line width=0.6pt, mark=*] table[x=W, y=RMSE_60] {data/number_of_filters.dat};
            \end{axis}
        \end{tikzpicture}
        \caption{\textit{RMSE}} 
        \label{subfig:effect_of_W_RMSE}  
      \end{subfigure}
    \begin{subfigure}[b]{0.49\linewidth}
      \begin{tikzpicture}
        \begin{axis}[
            xtick=data,
            xmode=log,
            xticklabels={16,32,64,128,256},
            ytick={2.0, 3.0, 4.0, 5.0},
            y tick label style={
                /pgf/number format/.cd,
                fixed,
                fixed zerofill,
                precision=1,
                /tikz/.cd
            },
            xlabel=$|\mathcal{W}_{t_{1}}|$,
            ylabel=\textit{MAPE},
            ymin=2.0,
            ymax=5.0,
            width=1.21\linewidth,
            height=0.48*\axisdefaultheight,
            legend style={at={(1.7,1.35)},anchor=north,legend columns=-1}]
            \addplot[red, line width=0.6pt, mark=triangle*] table[x=W, y=MAPE_15] {data/number_of_filters.dat};
            \addplot[black, line width=0.6pt, mark=square*] table[x=W, y=MAPE_30] {data/number_of_filters.dat};
            \addplot[blue, line width=0.6pt, mark=*] table[x=W, y=MAPE_60] {data/number_of_filters.dat};
            \end{axis}
        \end{tikzpicture}
        \caption{\textit{MAPE}} 
        \label{subfig:effect_of_W_MAPE}  
      \end{subfigure}
      \begin{subfigure}[b]{0.49\linewidth}
      \begin{tikzpicture}
        \begin{axis}[
            xtick=data,
            xmode=log,
            xticklabels={16,32,64,128,256},
            ytick={1.6,2.0,2.4,2.8},
            y tick label style={
                /pgf/number format/.cd,
                fixed,
                fixed zerofill,
                precision=1,
                /tikz/.cd
            },
            xlabel=$|\mathcal{W}_{t_{1}}|$,
            ylabel=\textit{Runtime ($10^3$s)},
            ymin=1.6,
            ymax=2.8,
            width=1.21\linewidth,
            height=0.48*\axisdefaultheight,
            legend style={at={(1.7,1.35)},anchor=north,legend columns=-1}]
            \addplot[orange, line width=0.6pt, mark=pentagon*] table[x=W, y=runtime] {data/number_of_filters.dat};
            \end{axis}
        \end{tikzpicture}
        \caption{\textit{Runtime} ($10^{3}s$)} 
        \label{subfig:effect_of_W_runtime}  
      \end{subfigure}
\caption{Effect of $|\mathcal{W}_{t_{1}}|$, \textbf{PEMS04-Evolve}.}
\label{fig:effect_of_W}

%% file: charts/number_of_orders_nonfig.tex
    \begin{subfigure}[b]{0.49\linewidth}
        \begin{tikzpicture}
            \begin{axis}[
            xtick=data,
            ytick={1.2,1.6,2.0,2.4},
            y tick label style={
                /pgf/number format/.cd,
                fixed,
                fixed zerofill,
                precision=1,
                /tikz/.cd
            },
            xlabel=$o$,
            ylabel=\textit{MAE},
            ymin=1.2,
            ymax=2.4,
            width=1.21\linewidth,
            height=0.48*\axisdefaultheight,
            legend style={at={(1.2,1.4)},anchor=north,legend columns=-1}]
            \addplot[red, line width=0.6pt, mark=triangle*] table[x=o, y=MAE_15] {data/number_of_orders.dat};
            \addlegendentry{15 mins}
            \addplot[black, line width=0.6pt, mark=square*] table[x=o, y=MAE_30] {data/number_of_orders.dat};
            \addlegendentry{30 mins}
            \addplot[blue, line width=0.6pt, mark=*] table[x=o, y=MAE_60] {data/number_of_orders.dat};
            \addlegendentry{60 mins}
        \end{axis}
        \end{tikzpicture}
        \caption{\textit{MAE}} 
        \label{subfig:effect_of_o_MAE}  
    \end{subfigure}
    \begin{subfigure}[b]{0.49\linewidth}
      \begin{tikzpicture}
        \begin{axis}[
            xtick=data,
            ytick={2.0,3.0,4.0,5.0},
            y tick label style={
                /pgf/number format/.cd,
                fixed,
                fixed zerofill,
                precision=1,
                /tikz/.cd
            },
            xlabel=$o$,
            ylabel=\textit{RMSE},
            ymin=2.0,
            ymax=5.0,
            width=1.21\linewidth,
            height=0.48*\axisdefaultheight,
            legend style={at={(1.7,1.35)},anchor=north,legend columns=-1}]
            \addplot[red, line width=0.6pt, mark=triangle*] table[x=o, y=RMSE_15] {data/number_of_orders.dat};
            \addplot[black, line width=0.6pt, mark=square*] table[x=o, y=RMSE_30] {data/number_of_orders.dat};
            \addplot[blue, line width=0.6pt, mark=*] table[x=o, y=RMSE_60] {data/number_of_orders.dat};
            \end{axis}
        \end{tikzpicture}
        \caption{\textit{RMSE}} 
        \label{subfig:effect_of_o_RMSE}  
      \end{subfigure}
    \begin{subfigure}[b]{0.49\linewidth}
      \begin{tikzpicture}
        \begin{axis}[
            xtick=data,
            ytick={2.0,3.0,4.0,5.0},
            y tick label style={
                /pgf/number format/.cd,
                fixed,
                fixed zerofill,
                precision=1,
                /tikz/.cd
            },
            xlabel=$o$,
            ylabel=\textit{MAPE},
            ymin=2.0,
            ymax=5.0,
            width=1.21\linewidth,
            height=0.48*\axisdefaultheight,
            legend style={at={(1.7,1.35)},anchor=north,legend columns=-1}]
            \addplot[red, line width=0.6pt, mark=triangle*] table[x=o, y=MAPE_15] {data/number_of_orders.dat};
            \addplot[black, line width=0.6pt, mark=square*] table[x=o, y=MAPE_30] {data/number_of_orders.dat};
            \addplot[blue, line width=0.6pt, mark=*] table[x=o, y=MAPE_60] {data/number_of_orders.dat};
            \end{axis}
        \end{tikzpicture}
        \caption{\textit{MAPE}} 
        \label{subfig:effect_of_o_MAPE}  
      \end{subfigure}
      \begin{subfigure}[b]{0.49\linewidth}
      \begin{tikzpicture}
        \begin{axis}[
            xtick=data,
            ytick={1.0,2.0,3.0,4.0},
            y tick label style={
                /pgf/number format/.cd,
                fixed,
                fixed zerofill,
                precision=1,
                /tikz/.cd
            },
            xlabel=$o$,
            ylabel=\textit{Runtime ($10^3$s)},
            ymin=1.0,
            ymax=4.0,
            width=1.21\linewidth,
            height=0.48*\axisdefaultheight,
            legend style={at={(1.7,1.35)},anchor=north,legend columns=-1}]
            \addplot[orange, line width=0.6pt, mark=pentagon*] table[x=o, y=runtime] {data/number_of_orders.dat};
            \end{axis}
        \end{tikzpicture}
        \caption{\textit{Runtime} ($10^{3}s$)} 
        \label{subfig:effect_of_o_runtime}  
      \end{subfigure}
\caption{Effect of $o$, \textbf{PEMS04-Evolve}.}
\label{fig:effect_of_o}

%% file: charts/number_of_conso_nodes_nonfig.tex
    \begin{subfigure}[b]{0.49\linewidth}
        \begin{tikzpicture}
            \begin{axis}[
            xtick=data,
            ytick={1.2,1.6,2.0,2.4},
            y tick label style={
                /pgf/number format/.cd,
                fixed,
                fixed zerofill,
                precision=1,
                /tikz/.cd
            },
            xlabel=$|\mathcal{B}_c|$ (\%),
            ylabel=\textit{MAE},
            ymin=1.2,
            ymax=2.4,
            width=1.21\linewidth,
            height=0.48*\axisdefaultheight,
            legend style={at={(1.2,1.4)},anchor=north,legend columns=-1}]
            \addplot[red, line width=0.6pt, mark=triangle*] table[x=B, y=MAE_15] {data/number_of_conso_nodes_v2.dat};
            \addlegendentry{15 mins}
            \addplot[black, line width=0.6pt, mark=square*] table[x=B, y=MAE_30] {data/number_of_conso_nodes_v2.dat};
            \addlegendentry{30 mins}
            \addplot[blue, line width=0.6pt, mark=*] table[x=B, y=MAE_60] {data/number_of_conso_nodes_v2.dat};
            \addlegendentry{60 mins}
        \end{axis}
        \end{tikzpicture}
        \caption{\textit{MAE}} 
        \label{subfig:effect_of_B_c_MAE}  
    \end{subfigure}
    \begin{subfigure}[b]{0.49\linewidth}
      \begin{tikzpicture}
        \begin{axis}[
            xtick=data,
            ytick={2.0,3.0,4.0,5.0},
            y tick label style={
                /pgf/number format/.cd,
                fixed,
                fixed zerofill,
                precision=1,
                /tikz/.cd
            },
            xlabel=$|\mathcal{B}_c|$ (\%),
            ylabel=\textit{RMSE},
            ymin=2.0,
            ymax=5.0,
            width=1.21\linewidth,
            height=0.48*\axisdefaultheight,
            legend style={at={(1.7,1.35)},anchor=north,legend columns=-1}]
            \addplot[red, line width=0.6pt, mark=triangle*] table[x=B, y=RMSE_15] {data/number_of_conso_nodes_v2.dat};
            \addplot[black, line width=0.6pt, mark=square*] table[x=B, y=RMSE_30] {data/number_of_conso_nodes_v2.dat};
            \addplot[blue, line width=0.6pt, mark=*] table[x=B, y=RMSE_60] {data/number_of_conso_nodes_v2.dat};
            \end{axis}
        \end{tikzpicture}
        \caption{\textit{RMSE}} 
        \label{subfig:effect_of_B_c_RMSE}  
      \end{subfigure}
    \begin{subfigure}[b]{0.49\linewidth}
      \begin{tikzpicture}
        \begin{axis}[
            xtick=data,
            ytick={2.5,3.5,4.5,5.5},
            y tick label style={
                /pgf/number format/.cd,
                fixed,
                fixed zerofill,
                precision=1,
                /tikz/.cd
            },
            xlabel=$|\mathcal{B}_c|$ (\%),
            ylabel=\textit{MAPE},
            ymin=2.5,
            ymax=5.5,
            width=1.21\linewidth,
            height=0.48*\axisdefaultheight,
            legend style={at={(1.7,1.35)},anchor=north,legend columns=-1}]
            \addplot[red, line width=0.6pt, mark=triangle*] table[x=B, y=MAPE_15] {data/number_of_conso_nodes_v2.dat};
            \addplot[black, line width=0.6pt, mark=square*] table[x=B, y=MAPE_30] {data/number_of_conso_nodes_v2.dat};
            \addplot[blue, line width=0.6pt, mark=*] table[x=B, y=MAPE_60] {data/number_of_conso_nodes_v2.dat};
            \end{axis}
        \end{tikzpicture}
        \caption{\textit{MAPE}} 
        \label{subfig:effect_of_B_c_MAPE}  
      \end{subfigure}
      \begin{subfigure}[b]{0.49\linewidth}
      \begin{tikzpicture}
        \begin{axis}[
            xtick=data,
            ytick={1.4,1.8,2.2,2.6},
            y tick label style={
                /pgf/number format/.cd,
                fixed,
                fixed zerofill,
                precision=1,
                /tikz/.cd
            },
            xlabel=$|\mathcal{B}_c|$ (\%),
            ylabel=\textit{Runtime ($10^3$s)},
            ymin=1.4,
            ymax=2.6,
            width=1.21\linewidth,
            height=0.48*\axisdefaultheight,
            legend style={at={(1.7,1.35)},anchor=north,legend columns=-1}]
            \addplot[orange, line width=0.6pt, mark=pentagon*] table[x=B, y=runtime] {data/number_of_conso_nodes_v2.dat};
            \end{axis}
        \end{tikzpicture}
        \caption{\textit{Runtime} ($10^{3}s$)} 
        \label{subfig:effect_of_B_c_runtime}  
      \end{subfigure}
\caption{Effect of $|\mathcal{B}_c|$, \textbf{PEMS04-Evolve}.}
\label{fig:effect_of_B_c}

%% file: charts/number_of_update_nodes_nonfig.tex
    \begin{subfigure}[b]{0.49\linewidth}
        \begin{tikzpicture}
            \begin{axis}[
            xtick=data,
            ytick={1.2,1.6,2.0,2.4},
            y tick label style={
                /pgf/number format/.cd,
                fixed,
                fixed zerofill,
                precision=1,
                /tikz/.cd
            },
            xlabel=$|\mathcal{B}_u|$ (\%),
            ylabel=\textit{MAE},
            ymin=1.2,
            ymax=2.4,
            width=1.21\linewidth,
            height=0.48*\axisdefaultheight,
            legend style={at={(1.2,1.4)},anchor=north,legend columns=-1}]
            \addplot[red, line width=0.6pt, mark=triangle*] table[x=B, y=MAE_15] {data/number_of_update_nodes.dat};
            \addlegendentry{15 mins}
            \addplot[black, line width=0.6pt, mark=square*] table[x=B, y=MAE_30] {data/number_of_update_nodes.dat};
            \addlegendentry{30 mins}
            \addplot[blue, line width=0.6pt, mark=*] table[x=B, y=MAE_60] {data/number_of_update_nodes.dat};
            \addlegendentry{60 mins}
        \end{axis}
        \end{tikzpicture}
        \caption{\textit{MAE}} 
        \label{subfig:effect_of_B_u_MAE}  
    \end{subfigure}
    \begin{subfigure}[b]{0.49\linewidth}
      \begin{tikzpicture}
        \begin{axis}[
            xtick=data,
            ytick={2.0,3.0,4.0,5.0},
            y tick label style={
                /pgf/number format/.cd,
                fixed,
                fixed zerofill,
                precision=1,
                /tikz/.cd
            },
            xlabel=$|\mathcal{B}_u|$ (\%),
            ylabel=\textit{RMSE},
            ymin=2.0,
            ymax=5.0,
            width=1.21\linewidth,
            height=0.48*\axisdefaultheight,
            legend style={at={(1.7,1.35)},anchor=north,legend columns=-1}]
            \addplot[red, line width=0.6pt, mark=triangle*] table[x=B, y=RMSE_15] {data/number_of_update_nodes.dat};
            \addplot[black, line width=0.6pt, mark=square*] table[x=B, y=RMSE_30] {data/number_of_update_nodes.dat};
            \addplot[blue, line width=0.6pt, mark=*] table[x=B, y=RMSE_60] {data/number_of_update_nodes.dat};
            \end{axis}
        \end{tikzpicture}
        \caption{\textit{RMSE}} 
        \label{subfig:effect_of_B_u_RMSE}  
      \end{subfigure}
    \begin{subfigure}[b]{0.49\linewidth}
      \begin{tikzpicture}
        \begin{axis}[
            xtick=data,
            ytick={2.5,3.5,4.5,5.5},
            y tick label style={
                /pgf/number format/.cd,
                fixed,
                fixed zerofill,
                precision=1,
                /tikz/.cd
            },
            xlabel=$|\mathcal{B}_u|$ (\%),
            ylabel=\textit{MAPE},
            ymin=2.5,
            ymax=5.5,
            width=1.21\linewidth,
            height=0.48*\axisdefaultheight,
            legend style={at={(1.7,1.35)},anchor=north,legend columns=-1}]
            \addplot[red, line width=0.6pt, mark=triangle*] table[x=B, y=MAPE_15] {data/number_of_update_nodes.dat};
            \addplot[black, line width=0.6pt, mark=square*] table[x=B, y=MAPE_30] {data/number_of_update_nodes.dat};
            \addplot[blue, line width=0.6pt, mark=*] table[x=B, y=MAPE_60] {data/number_of_update_nodes.dat};
            \end{axis}
        \end{tikzpicture}
        \caption{\textit{MAPE}} 
        \label{subfig:effect_of_B_MAPE}  
      \end{subfigure}
      \begin{subfigure}[b]{0.49\linewidth}
      \begin{tikzpicture}
        \begin{axis}[
            xtick=data,
            ytick={1.4,1.8,2.2,2.6},
            y tick label style={
                /pgf/number format/.cd,
                fixed,
                fixed zerofill,
                precision=1,
                /tikz/.cd
            },
            xlabel=$|\mathcal{B}_u|$ (\%),
            ylabel=\textit{Runtime ($10^3$s)},
            ymin=1.4,
            ymax=2.6,
            width=1.21\linewidth,
            height=0.48*\axisdefaultheight,
            legend style={at={(1.7,1.35)},anchor=north,legend columns=-1}]
            \addplot[orange, line width=0.6pt, mark=pentagon*] table[x=B, y=runtime] {data/number_of_update_nodes.dat};
            \end{axis}
        \end{tikzpicture}
        \caption{\textit{Runtime} ($10^{3}s$)} 
        \label{subfig:effect_of_B_u_runtime}  
      \end{subfigure}
    \caption{Effect of $|\mathcal{B}_u|$, \textbf{PEMS04-Evolve}.}
\label{fig:effect_of_B_u}

%% file: charts/number_of_tau_nonfig.tex
    \begin{subfigure}[b]{0.49\linewidth}
        \begin{tikzpicture}
            \begin{axis}[
            xtick=data,
            ytick={1.0,1.8,2.6,3.4},
            y tick label style={
                /pgf/number format/.cd,
                fixed,
                fixed zerofill,
                precision=1,
                /tikz/.cd
            },
            xlabel=$\tau$ ($days$),
            ylabel=\textit{MAE},
            ymin=1.0,
            ymax=3.4,
            width=1.21\linewidth,
            height=0.48*\axisdefaultheight,
            legend style={at={(1.2,1.4)},anchor=north,legend columns=-1}]
            \addplot[red, line width=0.6pt, mark=triangle*] table[x=tau, y=MAE_15] {data/number_of_tau.dat};
            \addlegendentry{15 mins}
            \addplot[black, line width=0.6pt, mark=square*] table[x=tau, y=MAE_30] {data/number_of_tau.dat};
            \addlegendentry{30 mins}
            \addplot[blue, line width=0.6pt, mark=*] table[x=tau, y=MAE_60] {data/number_of_tau.dat};
            \addlegendentry{60 mins}
        \end{axis}
        \end{tikzpicture}
        \caption{\textit{MAE}} 
        \label{subfig:effect_of_tau_MAE}  
    \end{subfigure}
    \begin{subfigure}[b]{0.49\linewidth}
      \begin{tikzpicture}
        \begin{axis}[
            xtick=data,
            ytick={2.4,3.6,4.8,6.0},
            y tick label style={
                /pgf/number format/.cd,
                fixed,
                fixed zerofill,
                precision=1,
                /tikz/.cd
            },
            xlabel=$\tau$ ($days$),
            ylabel=\textit{RMSE},
            ymin=2.4,
            ymax=6.0,
            width=1.21\linewidth,
            height=0.48*\axisdefaultheight,
            legend style={at={(1.7,1.35)},anchor=north,legend columns=-1}]
            \addplot[red, line width=0.6pt, mark=triangle*] table[x=tau, y=RMSE_15] {data/number_of_tau.dat};
            \addplot[black, line width=0.6pt, mark=square*] table[x=tau, y=RMSE_30] {data/number_of_tau.dat};
            \addplot[blue, line width=0.6pt, mark=*] table[x=tau, y=RMSE_60] {data/number_of_tau.dat};
            \end{axis}
        \end{tikzpicture}
        \caption{\textit{RMSE}} 
        \label{subfig:effect_of_tau_RMSE}  
      \end{subfigure}
    \begin{subfigure}[b]{0.49\linewidth}
      \begin{tikzpicture}
        \begin{axis}[
            xtick=data,
            ytick={2.4,3.8,5.2,6.6},
            y tick label style={
                /pgf/number format/.cd,
                fixed,
                fixed zerofill,
                precision=1,
                /tikz/.cd
            },
            xlabel=$\tau$ ($days$),
            ylabel=\textit{MAPE},
            ymin=2.4,
            ymax=6.6,
            width=1.21\linewidth,
            height=0.48*\axisdefaultheight,
            legend style={at={(1.7,1.35)},anchor=north,legend columns=-1}]
            \addplot[red, line width=0.6pt, mark=triangle*] table[x=tau, y=MAPE_15] {data/number_of_tau.dat};
            \addplot[black, line width=0.6pt, mark=square*] table[x=tau, y=MAPE_30] {data/number_of_tau.dat};
            \addplot[blue, line width=0.6pt, mark=*] table[x=tau, y=MAPE_60] {data/number_of_tau.dat};
            \end{axis}
        \end{tikzpicture}
        \caption{\textit{MAPE}} 
        \label{subfig:effect_of_tau_MAPE}  
      \end{subfigure}
      \begin{subfigure}[b]{0.49\linewidth}
      \begin{tikzpicture}
        \begin{axis}[
            xtick=data,
            ytick={1.98, 2.02, 2.06, 2.1},
            y tick label style={
                /pgf/number format/.cd,
                fixed,
                fixed zerofill,
                precision=2,
                /tikz/.cd
            },
            xlabel=$\tau$ ($days$),
            ylabel=\textit{Runtime ($10^3s$)},
            ymin=1.98,
            ymax=2.10,
            width=1.21\linewidth,
            height=0.48*\axisdefaultheight,
            legend style={at={(1.7,1.35)},anchor=north,legend columns=-1}]
            \addplot[orange, line width=0.6pt, mark=pentagon*] table[x=tau, y=runtime] {data/number_of_tau.dat};
            \end{axis}
        \end{tikzpicture}
        \caption{\textit{Runtime} ($10^3s$)} 
        \label{subfig:effect_of_tau_time}  
      \end{subfigure}
\caption{Effect of $\tau$, \textbf{PEMS04-Evolve}.}
\label{fig:effect_of_tau}

%% file: sections_arXiv/relatedwork.tex
\section{Related Work}
\label{sec:relatedwork}
\noindent\textbf{Traffic Forecasting.} 
Capturing spatio-temporal dynamics is one of the most essential aspects of traffic forecasting models. 
Thus, a variety of neural network based methods that build on \texttt{GNN}s~\cite{DBLP:conf/icde/CirsteaYGKP22} and \texttt{TCN}s~\cite{DBLP:conf/icde/CirsteaKG0P21} have been proposed that aim to capture complex spatio-temporal dynamics, thereby achieving competitive forecasting performance. 
Yu et al.~\cite{DBLP:conf/ijcai/YuYZ18} propose a traffic forecasting framework that combines \texttt{GCN}s and \texttt{1DCNN}s in a ``sandwich" architecture. 
Li et al.~\cite{DBLP:conf/iclr/LiYS018} propose a traffic forecasting model using a sequence-to-sequence architecture that combines \texttt{GCN}s and Recurrent Neural Networks (\texttt{RNN}s). 
Wu et al.~\cite{DBLP:conf/ijcai/WuPLJZ19} combines \texttt{GCN}s with \texttt{WaveNet}~\cite{DBLP:jour/corr/OordDZSVGKSK16}, which is an advanced \texttt{1DCNN}.
Attention mechanisms are also used for modeling spatial and temporal information. 
Zheng et al.~\cite{DBLP:conf/aaai/ZhengFWQ20} propose a framework that employs spatial attentions to model the correlations among multiple time series in an RN and employs temporal attentions to model the importance of each time step in a time series.
Razvan et al.~\cite{DBLP:conf/icde/CirsteaYGKP22} propose an attention-based framework that encompasses location-specific and time-varying model parameters to better capture complex spatio-temporadynamics. Jiang et al.~\cite{DBLP:conf/aaai/JiangHZW23} model the time delay in spatial information propagation and use a masking mechanism to model both short-range and long-range spatial dependencies.
Motivated by meta-learning, \texttt{MetaStore}~\cite{DBLP:journals/tist/LiuGZZCZZSY21} and \texttt{MetaST}~\cite{DBLP:conf/www/YaoLWTL19} enable transfer of knowledge from a data-abundant source city to a data-limited target city.
Lanza et al.~\cite{DBLP:conf/ciot/LanzaAMD23} propose a framework that combines federated learning and continual learning to sequentially train a global traffic forecasting model from the traffic signals of different local nodes.
However, existing studies only work on static time series and fixed-topology RNs.
The most relevant study to ours is \texttt{TrafficStream}~\cite{DBLP:conf/ijcai/ChenWX21} that also employs a subset of nodes of RNs to efficiently capture traffic signal. However, \texttt{TrafficStream} can contend with only the expansion of RNs and cannot work on evolved RNs, where RNs can expand, shrink, or undergo topological updates. Further, \texttt{TrafficStream} does not work well on small-scale data that is typically available for newly expanded regions. In contrast, TEAM contends with these aspects and works on evolved RNs. In particular, TEAM works well on small-scale data due to its hybrid architecture.
To the best of our knowledge, \texttt{TEAM} is the first proposal to contend with incremental time series and evolving RNs.

\vspace{0.5em}
\noindent\textbf{Dynamic Graph Embedding.} 
Approaches exist that model dynamic graphs by capturing the progression of topology changing over time~\cite{DBLP:journals/jmlr/KazemiGJKSFP20}.
\texttt{DynGEM}~\cite{DBLP:journals/ijcai/GoyalKHL17} exploits graph-autoencoders for incrementally updating node embedding by initializing them based on the previous step. 
\texttt{DyRep}~\cite{DBLP:conf/iclr/TrivediFBZ19} employs point processes to dynamically model edge occurrences between changing nodes.
\texttt{Dynamic\-Triad}~\cite{DBLP:conf/aaai/ZhouYR0Z18} focuses on the specific structure of triads to model how closed triads (three interconnected vertices) are formed from open triads (three vertices not interconnected).
\texttt{HTNE}~\cite{DBLP:conf/kdd/ZuoLLGHW18} captures dynamics by employing the Hawkes process and an attention mechanism to assess the impact of historical neighbors on the current neighbors of a node.
\texttt{EvolveGCN}~\cite{DBLP:conf/aaai/ParejaDCMSKKSL20} adopts \texttt{GCN}s to generate node embeddings for each snapshot and utilizes \texttt{RNN}s to train the \texttt{GCN}s.
These approaches face the notable scalability issue that they must fully train the models on the entire graph at every timestep. To the best of our knowledge, our proposed framework is the first study that considers efficiency in model training.

\vspace{0.5em}
\noindent\textbf{Rehearsal-based Continual Learning.} 
Rehearsal-based methods avoid catastrophic forgetting in continual learning by replaying a subset of old representative samples stored in a size-constrained memory buffer.
Numerous studies propose algorithms to choose the most representative samples for the buffer~\cite{DBLP:journals/eswa/NguyenKDKDNL24}.
Chaudhry et al.~\cite{DBLP:conf/iclr/ChaudhryRRE19} propose reservoir sampling that guarantees that each input sample has the same probability of entering the buffer.
Lopez et al.~\cite{DBLP:conf/nips/LopezR17} propose a ring buffer that allocates an equal-sized buffer to each class. 
Aljundi et al.~\cite{DBLP:conf/nips/AljundiLGB19} propose a gradient-based sampling algorithm to reduce overfitting of the reservoir and the ring algorithms by maximizing the diversity of samples in the buffer. 
However, rehearsal-based continual learning algorithms are only applied to perform classification tasks.
To the best of our knowledge, our study is the first to adapt rehearsal-based continual learning for traffic forecasting, a regression problem.

%% file: sections_arXiv/conclusion.tex
\section{Conclusion and Future Work}
\label{sec:conclusion}
We present Topological Evolution-aware Framework (\texttt{TEAM}), a framework for solving traffic forecasting in evolving RNs.
For the core of the framework, we propose a spatio-temporal model with a hybrid architecture, namely Convolution Attention for Spatio-Temporal (\texttt{CAST}), that combines convolution and attention to adapt better to incremental time series.
We propose a continual learning module based on the rehearsal method and integrate the module into the framework.
The continual module works as a buffer to store limited time series subsequences from the most stable and the most unstable nodes.
The model is then trained on the data of newly added nodes and the data in the buffer.
Experimental studies show that the framework is capable of outperforming strong baselines and state-of-the-art methods.

\vspace{0.5em}
In future research, it is of interest to study traffic forecasting with only a limited amount of training data~\cite{DBLP:journals/tip/DemirBB13,miao2024condensation}. 
It is also of interest to attempt to further improve the continual learning module, e.g., by identifying a better strategy for selecting representative time series subsequences~\cite{DBLP:conf/iclr/YoonMYH22}, by capturing temporal information better~\cite{DBLP:conf/nips/TangSLZZH23,DBLP:journals/pacmmod/Wu0ZG0J23,cheng2024memfromer}, or by modeling the continual learning using a generative model~\cite{DBLP:conf/iclr/AyubW21}. Further, it is of interest to consider distributed model training~\cite{DBLP:journals/pvldb/LiangW22}.